\documentclass{article} 
\let\PaperAddContentsLine\addcontentsline
\PassOptionsToPackage{table}{xcolor}
\usepackage{iclr2027_conference,times}
\let\addcontentsline\PaperAddContentsLine

\usepackage{amsmath,amsfonts,bm}

\def\eqref#1{equation~\ref{#1}}

\def\1{\bm{1}}

\def\vzero{{\bm{0}}}
\def\vone{{\bm{1}}}
\def\vmu{{\bm{\mu}}}

\def\vc{{\bm{c}}}

\def\vh{{\bm{h}}}

\def\vo{{\bm{o}}}
\def\vp{{\bm{p}}}

\def\vu{{\bm{u}}}

\def\vw{{\bm{w}}}
\def\vx{{\bm{x}}}
\def\vy{{\bm{y}}}
\def\vz{{\bm{z}}}

\def\mA{{\bm{A}}}
\def\mB{{\bm{B}}}
\def\mC{{\bm{C}}}
\def\mD{{\bm{D}}}

\def\mH{{\bm{H}}}
\def\mI{{\bm{I}}}

\def\mK{{\bm{K}}}

\def\mP{{\bm{P}}}
\def\mQ{{\bm{Q}}}
\def\mR{{\bm{R}}}

\def\mU{{\bm{U}}}
\def\mV{{\bm{V}}}
\def\mW{{\bm{W}}}
\def\mX{{\bm{X}}}
\def\mY{{\bm{Y}}}
\def\mZ{{\bm{Z}}}

\def\mSigma{{\bm{\Sigma}}}

\DeclareMathAlphabet{\mathsfit}{\encodingdefault}{\sfdefault}{m}{sl}
\SetMathAlphabet{\mathsfit}{bold}{\encodingdefault}{\sfdefault}{bx}{n}

\def\sL{{\mathbb{L}}}

\newcommand{\E}{\mathbb{E}}

\newcommand{\R}{\mathbb{R}}

\DeclareMathOperator*{\argmin}{arg\,min}

\DeclareMathOperator{\Tr}{Tr}

\usepackage{microtype}
\usepackage{xspace}

\newcommand{\methodname}{\textsc{EvolvingAvatar}}
\newcommand{\benchname}{\textsc{InterHead-Bench}}
\newcommand{\factoryname}{\textsc{dialog3d-factory}}

\usepackage{lipsum}

\usepackage{amsmath,amssymb,amsfonts}
\usepackage{mathtools}
\usepackage{bm}
\usepackage{bbm}
\providecommand{\vsigma}{\bm{\sigma}}

\usepackage{graphicx}
\usepackage{subcaption}
\usepackage{wrapfig}
\usepackage{adjustbox}

\usepackage{booktabs}
\usepackage{multirow}
\usepackage{array}
\usepackage{tabularx}
\usepackage{makecell}
\usepackage{siunitx}
\usepackage[table]{xcolor}
\definecolor{TableStateText}{HTML}{58636E}
\definecolor{TableSeparator}{HTML}{A7ADB3}
\definecolor{DatasetBandBG}{HTML}{EEF2F5}
\definecolor{DatasetBandText}{HTML}{263B50}
\definecolor{BestResultBG}{HTML}{B7F0E2}
\definecolor{SecondResultBG}{HTML}{CFE8FF}
\newcolumntype{Y}{l}
\newcolumntype{Z}{
  >{\centering\arraybackslash}m{0.95cm}
}
\newcommand{\DatasetBand}[1]{%
  \addlinespace[3pt]
  \rowcolor{DatasetBandBG}%
  \multicolumn{18}{c}{%
    \rule{0pt}{2.7ex}%
    \textbf{\color{DatasetBandText}#1}%
  }\\[-0.2ex]
}
\newcommand{\StateRow}[1]{%
  \addlinespace[2pt]
  \multicolumn{18}{c}{%
    \textit{\textbf{\color{TableStateText}#1}}%
  }\\[-0.05ex]
}
\newcommand{\MethodRef}[2]{%
  \mbox{%
    \textbf{#1}%
    ~{\normalfont\scriptsize\citep{#2}}%
  }%
}
\newcommand{\MethodOurs}{%
  \mbox{%
    \textbf{\methodname}%
    ~{\normalfont\scriptsize(Ours)}%
  }%
}
\newcommand{\DataRow}[8]{%
  #1
  & #2 
  & #3 
  & #4 
  & #5 
  & #6 
  & #7 
  & #8 
  \\
}
\newcommand{\PlainMetricHead}[1]{%
  \makecell[c]{%
    \raisebox{-0.8ex}[0pt][0pt]{\textbf{#1}}%
  }%
}

\newcommand{\best}[1]{\cellcolor{BestResultBG}\textbf{#1}}
\newcommand{\second}[1]{\cellcolor{SecondResultBG}\underline{#1}}

\usepackage{algorithm}
\usepackage{algpseudocode}

\usepackage{tikz}
\usetikzlibrary{
    arrows.meta,
    positioning,
    shapes,
    calc,
    fit
}

\usepackage{enumitem}

\usepackage{pifont}
\usepackage[normalem]{ulem}
\newcommand{\ContextYes}{\textcolor{green!55!black}{\ding{51}}}
\newcommand{\ContextNo}{\textcolor{red!70!black}{\ding{55}}}

\renewcommand{\paragraph}[1]{\vspace{.1em}\noindent\textbf{#1.}}

\usepackage[
    colorlinks=true,
    allcolors=black
]{hyperref}
\usepackage{url}

\definecolor{PublicInk}{HTML}{23364A}
\definecolor{PublicAccent}{HTML}{286D88}
\definecolor{PublicMuted}{HTML}{556473}
\definecolor{PublicSurface}{HTML}{F5F8FB}
\definecolor{PublicBorder}{HTML}{CCD8E2}

\newcommand{\AuthorAffiliation}[1]{%
  \textsuperscript{\normalfont\bfseries\scriptsize\color{PublicAccent}#1}%
}
\newcommand{\Institution}[2]{%
  {\normalfont\small\color{PublicMuted}%
    \tikz[baseline=(affiliationnumber.base)]{%
      \node[
        fill=PublicAccent!9,
        text=PublicAccent,
        rounded corners=1.5pt,
        inner sep=1pt,
        minimum width=9pt,
        minimum height=9pt,
        font=\normalfont\bfseries\scriptsize
      ] (affiliationnumber) {#1};%
    }\hspace{3pt}#2}%
}

\newsavebox{\PublicAbstractBox}
\AtBeginDocument{%
  \ificlrfinal
    \fancyhead[L]{\normalfont\small\scshape\color{PublicMuted}Preprint}
    \fancyhead[R]{\normalfont\small\color{PublicMuted}\methodname}
    \renewenvironment{abstract}{%
      \par\addvspace{8pt}%
      \begin{lrbox}{\PublicAbstractBox}%
      \begin{minipage}{\dimexpr\textwidth-2\leftmargini\relax}%
      \normalfont\normalsize
      \setlength{\parindent}{0pt}%
      \setlength{\parskip}{0pt}%
      \ignorespaces
    }{%
      \par\end{minipage}%
      \end{lrbox}%
      \noindent\makebox[\textwidth][c]{%
        \begin{tikzpicture}
          \node[
            draw=PublicBorder,
            fill=PublicSurface,
            line width=0.5pt,
            rounded corners=5pt,
            inner xsep=16pt,
            inner ysep=11pt
          ] (abstractpanel) {\usebox{\PublicAbstractBox}};
          \node[
            anchor=west,
            fill=PublicSurface,
            text=PublicAccent,
            inner xsep=6pt,
            inner ysep=1.5pt,
            font=\normalfont\scshape\small
          ] at ([xshift=10pt]abstractpanel.north west) {Abstract};
        \end{tikzpicture}%
      }\par\addvspace{9pt}%
    }
  \fi
}

\iclrfinalcopy
\authorcentercopy

\title{\methodname: Interactive 3D Head Generation That Adapts as Conversations Unfold}

\author{
{\normalfont\large\bfseries\color{PublicInk}%
Junjie Chen\AuthorAffiliation{1,2}\thanks{Work done during an internship at EPIC Lab.}%
\qquad
Fei Wang\AuthorAffiliation{3}%
\qquad
Kun Li\AuthorAffiliation{6}%
\qquad
Yiqi Nie\AuthorAffiliation{3,7}%
}
\\[3pt]
{\normalfont\large\bfseries\color{PublicInk}%
Xun Yang\AuthorAffiliation{5}%
\qquad
Yanbin Hao\AuthorAffiliation{1}\footnotemark[2]%
\qquad
Linfeng Zhang\AuthorAffiliation{2,4}\thanks{Corresponding authors.}%
\qquad
Meng Wang\AuthorAffiliation{1}\footnotemark[2]%
}
\\[7pt]
\Institution{1}{Hefei University of Technology}
\quad
\Institution{2}{EPIC Lab, Shanghai Jiao Tong University}
\\[2pt]
\Institution{3}{Institute of Artificial Intelligence, Hefei Comprehensive National Science Center}
\\[2pt]
\Institution{4}{SAI, Shanghai Jiao Tong University}
\quad
\Institution{5}{University of Science and Technology of China}
\\[2pt]
\Institution{6}{United Arab Emirates University}
\quad
\Institution{7}{Anhui University}
\\[5pt]
{\normalfont\small
\href{mailto:jorji.chen@gmail.com}{\textcolor{PublicMuted}{jorji.chen@gmail.com}}}
\\[5pt]
\href{https://blog.evolving-avatar.com}{%
\begin{tikzpicture}[baseline=(projectlink.base)]
\node[
  draw=PublicBorder,
  fill=PublicSurface,
  text=PublicInk,
  line width=0.4pt,
  rounded corners=3pt,
  inner xsep=10pt,
  inner ysep=2.5pt,
  font=\normalfont\small
] (projectlink) {%
  {\scriptsize\scshape\color{PublicAccent}Project Page}%
  \hspace{8pt}%
  \textcolor{PublicBorder}{\rule[-2pt]{0.4pt}{10pt}}%
  \hspace{8pt}%
  blog.evolving-avatar.com%
  \hspace{6pt}\ensuremath{\nearrow}%
};
\end{tikzpicture}%
}
}
\begin{document}
\addtocontents{toc}{\protect\setcounter{tocdepth}{-1}}

\maketitle

\suppressfloats[t]
\begin{figure}[t]
    \centering
    \newsavebox{\overviewcaptionbox}
    \sbox{\overviewcaptionbox}{%
        \begin{minipage}[t]{0.45\textwidth}
            \captionsetup{skip=0pt}
            \caption{\textbf{Learning from interaction.} (a) Our method learns from user face
            video and dyadic audio through test-time training, carrying context-driven
            updates across the conversation without target motion labels. Existing
            approaches use (b) avatar audio~\citep[e.g.,][]{sun2024diffposetalk}, (c)
            dyadic audio~\citep[e.g.,][]{chu2026unils}, or (d) dyadic audio and
            estimated user motion~\citep[e.g.,][]{peng2025dualtalk}. Their parameters
            remain fixed during inference. A dialogue model can supply avatar speech
            from user audiovisual input. Generated 3D motion supports avatar
            animation, robotic facial actuation, and talking-head synthesis.}
            \label{fig:overview}
        \end{minipage}%
    }
    \makebox[\linewidth][s]{%
        \raisebox{\dimexpr\ht\overviewcaptionbox-\height\relax}{%
            \adjustbox{min width=0.50\linewidth}{%
                \includegraphics[
                    height=\dimexpr\ht\overviewcaptionbox+\dp\overviewcaptionbox\relax
                ]{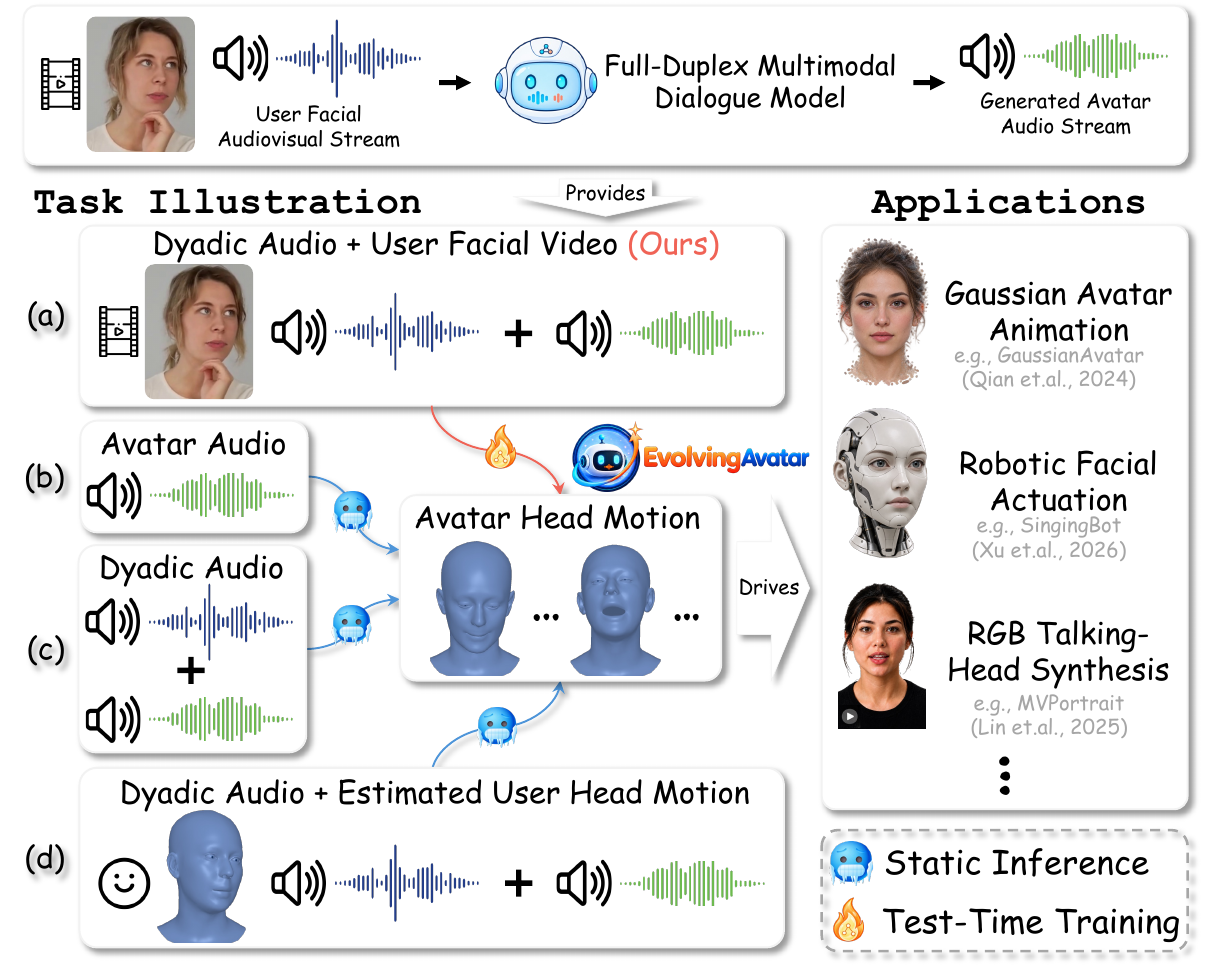}%
            }%
        }\hfill
        \usebox{\overviewcaptionbox}%
    }
\end{figure}

\begin{abstract}
Interactive 3D head generation requires coordinated speaking and listening
motion that responds to an evolving conversation. Existing generators use
incoming observations as context but keep their parameters fixed, leaving
conversational patterns unused as a learning signal. We introduce
\methodname{}, a causal generator that uses test-time training to adapt to
user face video and dyadic audio during interaction. Its \textit{dyadic
context prediction} objective provides a self-supervised learning signal
from audiovisual context without target motion labels at test time.
Persistent fast weights accumulate these updates within each conversation
to guide motion generation, while transient jaw adaptation responds to
current audiovisual context. Predicted speech activity controls how
persistent adaptation guides motion. We also introduce \benchname{}, a
unified 455.95-hour benchmark built from single-view and dual-view
conversation videos. Experiments show improved conversational motion
statistics over strong baselines. On the hardest out-of-distribution split,
generation improves as conversations unfold, reducing mismatch with recorded
user--avatar expression statistics by up to 11.1\% from the first interval.
\end{abstract}

\section{Introduction}

Interactive 3D head generation must coordinate speech articulation,
listening responses, and transitions between them. Generated motion should
reflect the partner's behavior and the ongoing conversation. Potential
applications include virtual tutors~\citep{graesser2005autotutor} and
avatars that offer mental health support by joining therapeutic
conversations alongside human clinicians~\citep{cao2026jodie}. Avatar speech
can come from dialogue systems that respond to user input in real time~(Figure~\ref{fig:overview}).
These include full-duplex models that support simultaneous listening and
speaking~\citep{defossez2024moshi,roy2026personaplex} and multimodal models
that combine audiovisual perception of the user with streaming
generation~\citep{xu2025qwen3omni,huang2026duplexomni,thinkingmachines2026interactionmodels}.

In human conversation, listeners signal understanding through nods, while
partners coordinate response
timing~\citep{clark1991grounding,stivers2009universals}. People may also
mirror their partner's nonverbal behavior, known as the \textit{chameleon
effect}~\citep{chartrand1999chameleon}. These findings motivate learning
from each conversation. Repeated observations of user behavior and both
participants' speech provide a basis for self-supervised learning that may
help guide motion.

As illustrated in Figure~\ref{fig:overview}, existing generators use (b)
avatar audio~\citep[e.g.,][]{sun2024diffposetalk}, (c) dyadic
audio~\citep[e.g.,][]{chu2026unils}, or (d) dyadic audio with estimated user
motion~\citep[e.g.,][]{peng2025dualtalk}. These inputs add context, but the
models do not update their parameters during interaction. Test-time training
(TTT)~\citep{sun2020test,zhang2025testtime} enables learning from this
context. The challenge is to design an objective that guides motion
generation without target motion labels.

We introduce \methodname{}, a causal generator that learns during
interaction without requiring precomputed user motion
(Figure~\ref{fig:overview}(a)). Its \emph{Dyadic Context Prediction}
objective uses user video and dyadic audio to provide a learning signal for
TTT. We train the generator on paired audiovisual context and motion to use
these updates for conversational motion generation. Persistent fast weights
retain information learned from the conversation across intervals, while
transient jaw adaptation responds to current speech. Predicted activity
controls how persistent adaptation affects speaking and listening motion. A
region-structured codec coordinates expression, neck pose, and jaw pose.

To evaluate this setting, \factoryname{} processes single-view and
dual-view conversation videos into aligned multimodal annotations. The
resulting 455.95-hour \benchname{} combines parameter-space and mesh-space
metrics with human ratings to evaluate generated motion. Experiments show
improvements in motion statistics over baselines and suggest that learning
from conversation context can benefit interactive motion generation. Our
three main contributions are:

\begin{itemize}[leftmargin=1em,nosep]
  \item \methodname{}, a causal generator that learns at test time without motion
  labels through Dyadic Context Prediction, with persistent conversational
  adaptation and transient jaw adaptation.
  \item \benchname{}, a unified benchmark of single-view and dual-view
  conversation videos with aligned multimodal annotations and evaluation
  of speaking and listening across distributions.
  \item Our results suggest that adaptation during interaction can benefit
  interactive 3D head generation and motivate further research on learning
  from video and audio as a conversation unfolds.
\end{itemize}

\section{Related Work}
\label{sec:related-work}

\paragraph{Interactive 3D Head Generation}
Prior work differs in the conversational context available to the avatar.
\ding{182}\,\textbf{\emph{Avatar audio.}}
Audio-driven talking-head models generate lip motion, expression, and head
pose in video~\citep{zhou2020makelttalk,prajwal2020lip,xu2024vasa}.
Geometry-aware methods predict motion coefficients~\citep{zhang2023sadtalker}
in 3DMM representations~\citep{blanz2023morphable,egger20203d}, while direct-3D
methods generate parameters or meshes using data-driven
models~\citep{karras2017audio,cudeiro2019capture,richard2021meshtalk},
transformers, discrete priors, autoregression, or
diffusion~\citep{fan2022faceformer,xing2023codetalker,sun2024diffposetalk}.
\ding{183}\,\textbf{\emph{User context.}}
Complementing speech animation, listening-head models use the interlocutor's
speech or facial motion to generate nonverbal
feedback~\citep{zhou2022responsive,ng2022learning}.  This line spans early
conversational agents~\citep{cassell1994animated} and recent 3DMM- or mesh-based
generation~\citep{tran2024dim,wang2025diffusion}.
\ding{184}\,\textbf{\emph{Dyadic context.}}
Jointly modeling both participants reflects the temporal interdependence of
dialogue~\citep{sacks1974simplest,skantze2021turn}.  Approaches include
image-space coordination~\citep{zhu2025infp,guo2025arig}, audiovisual dyadic
head generation~\citep{zhou2025interactive,chen2026towards}, and dual-audio 3D motion
synthesis~\citep{chu2026unils}.  DualTalk further incorporates user motion to
model speaking and listening~\citep{peng2025dualtalk}.
These developments expand the context used for generation, but existing generators retain fixed parameters at
deployment. This motivates adapting the mapping from context to motion as each conversation reveals distinct patterns of
user behavior, avatar speaking style, and turn-taking dynamics.
  
\paragraph{Test-Time Training}
Test-time training (TTT) provides a mechanism for such adaptation by updating
fast parameters from unlabeled deployment observations.  Recent work studies
TTT training strategies~\citep{zhang2025testtime} and next-token or in-place
updates~\citep{ouyang2026testtime,feng2026inplace}, with applications to vision
and spatial memory~\citep{han2026vit3,ma2026fast}, long-context 3D
reconstruction~\citep{wang2026tttlrm}, and robot policies~\citep{jiang2026robottt}.
Interactive 3D head generation offers a natural setting for TTT: each
conversation continuously supplies unlabeled audiovisual evidence about user
behavior, avatar speaking style, and their evolving coordination.  These
interaction-specific regularities provide an opportunity to learn during
generation, beyond conditioning a fixed model on incoming context.
\methodname{} bridges this gap by using the unfolding dialogue itself as a
self-supervised adaptation signal, allowing the generator to adjust its
context-to-motion mapping within each conversation.  This brings
conversation-specific adaptation to causal head generation without requiring
target motion at deployment.

\section{Preliminaries}
\label{sec:preliminaries}

\paragraph{3D head motion representation}
Following prior work~\citep{peng2025dualtalk,chu2026unils}, we represent 3D head motion using FLAME parameters~\citep{li2017learning}. At each frame $t$, the motion is encoded as a 106-D vector consisting of expression coefficients, neck pose, and jaw pose:
\begin{equation*}
\vz_t
= \left[
(\vz_t^e)^\top,
(\vz_t^n)^\top,
(\vz_t^j)^\top
\right]^\top,
\qquad
(d_e,d_n,d_j)=(100,3,3).
\end{equation*}
The components encode expression, neck rotation, and jaw rotation,
respectively.  Neck and jaw rotations use axis-angle coordinates, with
$\mR:\R^3\rightarrow\mathrm{SO}(3)$ mapping each vector to a rotation matrix.
Following the task definition of~\citet{peng2025dualtalk}, we exclude FLAME
shape, global root pose, and eye pose.
Let $\vx_t$ and $\vy_t$ denote the predicted and reference motion parameters
at frame $t$.  Their temporally ordered sequences define the corresponding
motion trajectories
$\mX=(\vx_1,\ldots,\vx_T)^\top$ and
$\mY=(\vy_1,\ldots,\vy_T)^\top$,
where $\mX,\mY\in\R^{T\times106}$ are temporally aligned.

\paragraph{Interactive 3D head generation task formulation}
We study interactive 3D head generation, where an avatar's head motion is
generated from the unfolding conversational context.
We call the observed interlocutor the \emph{user} and the participant being
animated the \emph{avatar}.  Let
$\mathcal{A}^{\mathrm{U}}_{\leq t}$,
$\mathcal{V}^{\mathrm{U}}_{\leq t}$, and
$\mathcal{A}^{\mathrm{A}}_{\leq t}$ denote the user audio, user visual evidence,
and avatar audio available through time $t$, respectively.  Generalizing
existing formulations~\citep{sun2024diffposetalk,peng2025dualtalk,chu2026unils},
we define the task as
\begin{equation*}
  \vx_t \sim p_\theta\!\left(\cdot \mid \mathcal{O}_{\leq t}\right),
  \qquad
  \varnothing \neq \mathcal{O}
  \subseteq
  \left\{
    \mathcal{A}^{\mathrm{U}},
    \mathcal{V}^{\mathrm{U}},
    \mathcal{A}^{\mathrm{A}}
  \right\},
  \qquad
  \mX \in \R^{T\times106}.
\end{equation*}
Our setting takes $\mathcal{O}=\{\mathcal{A}^{\mathrm{U}},
\mathcal{V}^{\mathrm{U}},\mathcal{A}^{\mathrm{A}}\}$ and observes
$\mathcal{V}^{\mathrm{U}}$ as causally arrived user face frames rather than a
precomputed user FLAME trajectory.

\paragraph{Test-time training}
We write a TTT layer with fixed, offline-learned \emph{slow weights} $\theta$
and \emph{fast weights} $W_t$ that constitute its adaptive
state~\citep{zhang2025testtime}.  Given an
arrived observation $\vo_t$, a self-supervised inner objective first updates
the fast weights, which then encode the same observation:
\begin{equation*}
  W_t
  = W_{t-1}
    - \eta_t \nabla_W \mathcal{L}_{\mathrm{inner}}(W_{t-1},\vo_t),
  \qquad
  \vh_t = f_{\theta,W_t}(\vo_t).
\end{equation*}
The arrived observation provides its own learning signal, requiring no task label.
Section~\ref{sec:evolving-avatar-generator} instantiates the
observation, inner objective, and fast-state lifetime used by \methodname.

\section{\methodname}
\label{sec:evolving-avatar}

\begin{figure*}[t]
  \centering
  \includegraphics[width=1\linewidth]{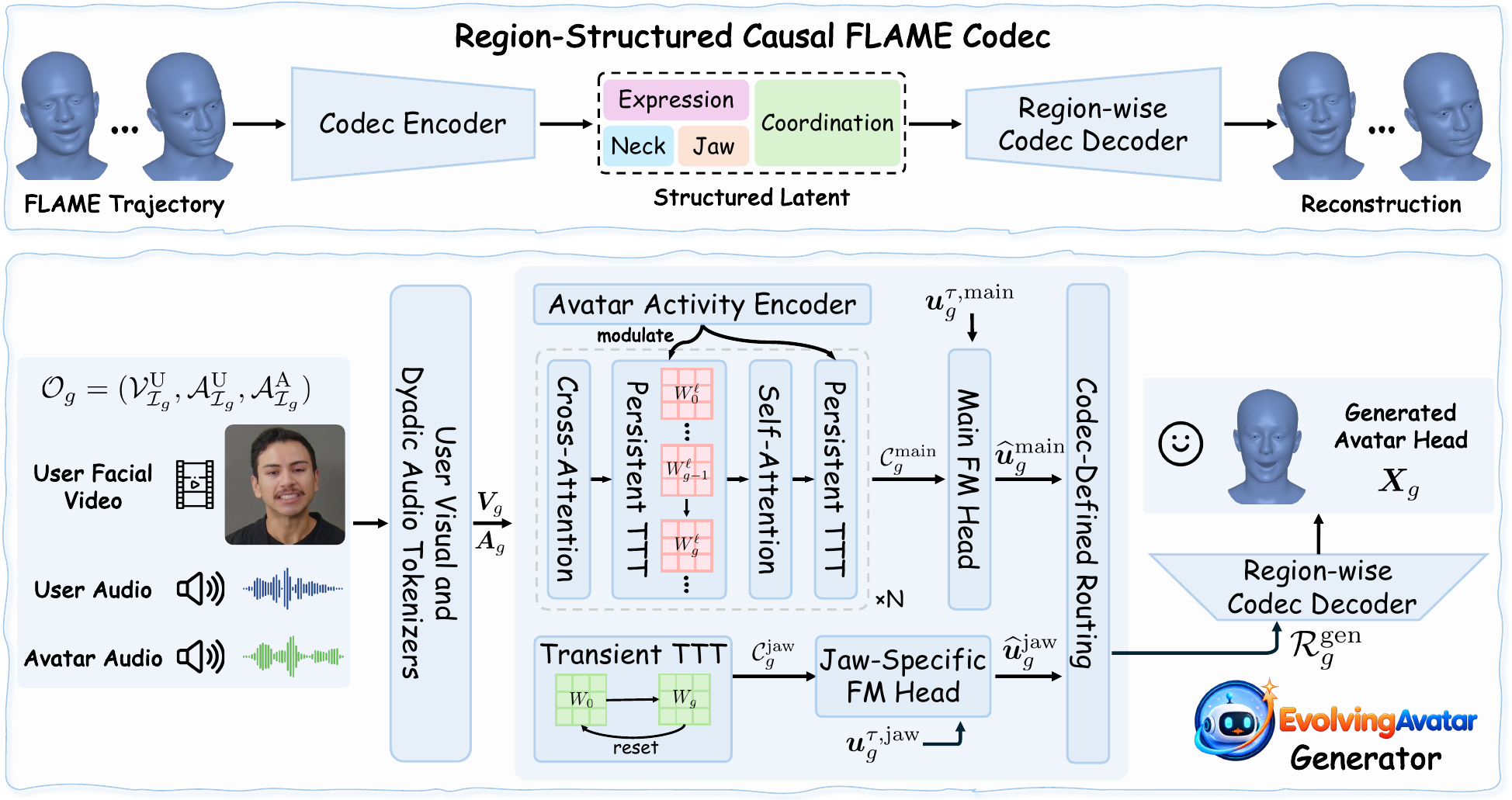}
  \caption{\methodname{} learns from the ongoing conversation at deployment without motion
  labels. Dyadic context prediction drives two complementary forms of adaptation:
  persistent fast weights capture conversational regularities, while transient jaw
  adaptation tracks current articulation. Regional latent routing combines their
  predictions through a shared causal decoder.}
  \label{fig:evolving-avatar-overview}
\end{figure*}

\methodname{} combines a region-structured causal FLAME \textit{codec} with an
adaptive \textit{generator}, shown in Figure~\ref{fig:evolving-avatar-overview}.
It predicts latent motion from user video and dyadic audio through
persistent conversational and transient articulation adaptation.
Time is partitioned into fixed-duration intervals $\mathcal{I}_g$, with observations
$\mathcal{O}_g=(\mathcal{V}^{\mathrm U}_{\mathcal{I}_g},
\mathcal{A}^{\mathrm U}_{\mathcal{I}_g},\mathcal{A}^{\mathrm A}_{\mathcal{I}_g})$.
Generation uses only $\mathcal{O}_{\leq g}$ and emits motion at the interval
boundary. Full implementation and optimization details appear in
Appendix~\ref{app:method-details}.

\subsection{Region-Structured Causal FLAME Codec}
\label{sec:flame-codec}

In Figure~\ref{fig:evolving-avatar-overview}, the \emph{Codec Encoder} maps
$\mY_g=[\vy_t]_{t\in\mathcal{I}_g}$ to a posterior over the \emph{Structured Latent}:
\begin{equation}
  q_\phi(\vu_g\mid\mY_{\leq g})
  =\mathcal{N}\!\left(\vmu_g,\operatorname{diag}(\vsigma_g^2)\right),
  \qquad
  \vu_g=[\vu_g^e,\vu_g^n,\vu_g^j,\vu_g^c].
  \label{eq:structured-flame-codec}
\end{equation}
The components represent expression, neck pose, jaw pose, and shared
coordination, respectively. Neck pose $n$ represents neck rotation,
corresponding to \emph{Neck} in the figure. The \emph{Region-wise Codec Decoder}
$D_\psi$ combines each private latent with shared coordination via
$\mathcal{R}(\vu_g)=([\vu_g^e,\vu_g^c],[\vu_g^n,\vu_g^c],[\vu_g^j,\vu_g^c])$.
Regional projections then feed a shared causal backbone that reconstructs
expression, neck pose, and jaw pose through separate region-specific output heads.

\paragraph{Codec training}
A region-balanced Gaussian loss $\mathcal{L}_{\mathrm{rec}}$ supervises motion
and its first and second temporal differences. We regularize the posterior
under a constraint on its information rate:
\begin{equation}
  \begin{aligned}
    \min_{\phi,\psi}\;&\mathcal{L}_{\mathrm{rec}}+
      \E_s\!\left[\sum_g\kappa_{s,g}\right]
      \quad\text{subject to}\quad
      \frac{\sum_{s,g}\kappa_{s,g}}{\sum_s G_s}\leq\rho,\\
    &\kappa_{s,g}=D_{\mathrm{KL}}\!\left(
      q_\phi(\vu_{s,g}\mid\mY_{s,\leq g})\|\mathcal{N}(\vzero,\mI)\right).
  \end{aligned}
  \label{eq:codec-training}
\end{equation}
Here $G_s$ counts valid intervals in sequence $s$ and $\rho$ is the rate budget,
enforced through a non-negative dual multiplier. After training, we freeze
the codec and use its posterior mean as the generator target,
$\vu_g^\star=\operatorname{sg}(\vmu_g)$, where $\operatorname{sg}$ denotes
stop-gradient. The decoder stays frozen.

\subsection{Adaptive Generator}
\label{sec:evolving-avatar-generator}

For each interval $g$, the generator takes the observed user face frames and both
participants' audio, as shown in Figure~\ref{fig:evolving-avatar-overview}. The
\emph{User Visual and Dyadic Audio Tokenizers} turn these streams into
$\mV_g\in\R^{Q_v\times D}$ and $\mA_g=[\mA_g^{\mathrm U};\mA_g^{\mathrm
A}]\in\R^{2Q_a\times D}$. Here $Q_v$ is the number of visual tokens, $Q_a$ is the
number of audio tokens per participant, and $D$ is their shared width.
At deployment, these tokens update fast weights before generation,
allowing the model to learn from the ongoing conversation.

\paragraph{Fusing and adapting context}
Across $N$ blocks, visual tokens alternate \emph{Cross-Attention} to audio with
\emph{Self-Attention}, producing context $\mH_g^\ell\in\R^{Q_v\times D}$ at block
$\ell$. To learn without motion labels, we use test-time training (TTT) with
\emph{dyadic context prediction} (DCP). Learned projections of $\mH_g^\ell$ give
keys, values, and queries. DCP predicts values from keys to capture regularities
in the observed context. \emph{Persistent TTT} updates fast weights with this
prediction error before reading the query:
\begin{equation}
  \begin{aligned}
    \mathcal{L}_{\mathrm{ctx},g}^\ell(W)
      &=\operatorname{MSE}(f_W(\mK_g^\ell),\mV_g^\ell),\\
    W_g^\ell
      &=W_{g-1}^\ell-\eta_g^\ell\nabla_W
        \mathcal{L}_{\mathrm{ctx},g}^\ell(W_{g-1}^\ell),\\
    \mD_g^\ell
      &=r_{W_g^\ell}^\ell(\mQ_g^\ell)
        -r_{W_0^\ell}^\ell(\mQ_g^\ell).
  \end{aligned}
  \label{eq:persistent-ttt}
\end{equation}
Here $W_0^\ell$ denotes learned initial weights, $\eta_g^\ell$ a positive update
rate, and $r_W^\ell$ the output for a query. The difference
$\mD_g^\ell\in\R^{Q_v\times D}$ compares the adapted and initial outputs. We add this
change to the visual tokens before the next block. Fast weights retain context across
intervals to guide motion.

\paragraph{Using activity to guide motion}
Speaking and listening call for different motions. The \emph{Avatar Activity Encoder}
therefore reads framewise audio and video features and predicts speaking/listening
probabilities. Over $F_g$ valid frames, these form $\mP_g\in\R^{F_g\times2}$. A
learned mapping gives the gate $\bm\gamma_g\in\R^D$. Shared across visual tokens, it
scales adaptation to condition motion:
\begin{equation}
  \mC_g^\ell=\mH_g^\ell+\bm\gamma_g\odot\mD_g^\ell
  \in\R^{Q_v\times D}.
  \label{eq:activity-conditioned-motion}
\end{equation}
The current context remains directly available through $\mH_g^\ell$. Activity changes
only the added adaptation term. It enters neither the fast-weight update nor the
context passed to later blocks. Training and inference both use predicted
probabilities without requiring activity labels as inputs.

\paragraph{Adapting to current articulation}
The jaw route instead adapts to current speech. Its \emph{Transient TTT} module takes
the detached tokens $\operatorname{sg}([\mA_g;\mV_g])\in\R^{(2Q_a+Q_v)\times D}$. It
projects them to width $D_j$ and applies DCP from its learned initial weights. Adding
the resulting change to the projected tokens gives
$\mC_g^{\mathrm{jaw}}\in\R^{(2Q_a+Q_v)\times D_j}$. This ungated branch discards its
fast weights after use.

\paragraph{Generating latent motion}
The adapted contexts guide two flow-matching (FM) heads~\citep{lipman2022flow}. The
\emph{Main FM Head} uses $\mC_g^{\mathrm{main}}=(\mC_g^1,\ldots,\mC_g^N)$ to generate
the complete codec latent $\widehat{\vu}_g^{\mathrm{main}}\in\R^{d_z}$. The
\emph{Jaw-Specific FM Head} uses $\mC_g^{\mathrm{jaw}}$ to generate
$\widehat{\vu}_g^{\mathrm{jaw}}\in\R^{d_{jc}}$, combining jaw and coordination
components. Each head maps its noisy latent to a motion token of width $D$ or $D_j$,
attends to its context, and predicts a latent velocity. Integrating this velocity
generates motion from noise as flow time $\tau$ runs from $0$ to $1$, while $g$
indexes intervals of the ongoing conversation.

\paragraph{Training the generator}
We train the generator to use earlier audiovisual observations when predicting
later motion. A sampled interval boundary divides each clip into a context segment $A$ and a supervised segment
$B$. We process both in time order, with DCP updates throughout and motion losses
only on $B$. Clips without $A$ train generation from the learned initial fast weights.

The frozen codec gives targets $\vu_g^{\star,\mathrm{main}}=\vu_g^\star$
and $\vu_g^{\star,\mathrm{jaw}}=\Pi_{\mathrm{jaw}}(\vu_g^\star)$, where
$\Pi_{\mathrm{jaw}}(\vu)=[\vu^j,\vu^c]$. For each branch $b$, we mix its target
with Gaussian noise $\bm\epsilon_g^b$ at flow time $\tau_g$. The velocity target
is the clean latent minus this noise. The predicted velocity also gives a clean
latent estimate:
\begin{equation}
  \begin{aligned}
    \vu_g^{\tau_g,b}
      &=(1-\tau_g)\bm\epsilon_g^b+\tau_g\vu_g^{\star,b},\\
    \mathcal{L}_{\mathrm{FM}}^b
      &=\operatorname{MSE}_{B}\!\left(
        v_{\theta_b}(\vu_g^{\tau_g,b},\tau_g,\mC_g^b),
        \vu_g^{\star,b}-\bm\epsilon_g^b\right),\\
    \widetilde{\vu}_g^b
      &=\vu_g^{\tau_g,b}+(1-\tau_g)
        v_{\theta_b}(\vu_g^{\tau_g,b},\tau_g,\mC_g^b).
  \end{aligned}
  \label{eq:dual-flow-matching}
\end{equation}
Here $\operatorname{MSE}_B$ averages errors over $B$, and $\widetilde{\vu}_g^b$
estimates the clean latent. To supervise motion changes,
$\mathcal{L}_{\Delta}^b$ matches differences between adjacent estimates to target differences,
covering the full main latent and only the private jaw latent.
$\mathcal{L}_{\mathrm{dec}}^{\mathrm{jaw}}$ compares decoded jaw parameters with ground truth.
These motion losses use $B$, while activity cross-entropy
$\mathcal{L}_{\mathrm{act}}$ uses all valid clip frames:
\begin{equation}
  \begin{aligned}
  \mathcal{L}_{\mathrm{gen}}
  &=\mathcal{L}_{\mathrm{FM}}^{\mathrm{main}}
   +\mathcal{L}_{\mathrm{FM}}^{\mathrm{jaw}}
   +\lambda_{\Delta}\!\left(
     \mathcal{L}_{\Delta}^{\mathrm{main}}
     +\mathcal{L}_{\Delta}^{\mathrm{jaw}}\right)\\
  &\quad+\lambda_{\mathrm{dec}}\mathcal{L}_{\mathrm{dec}}^{\mathrm{jaw}}
   +\lambda_{\mathrm{act}}\mathcal{L}_{\mathrm{act}}.
  \end{aligned}
  \label{eq:stage2-training}
\end{equation}
Motion gradients pass through DCP updates to train the context features and initial
fast weights for motion prediction. Detached jaw inputs prevent its losses from
updating the main branch. Detached activity probabilities leave the activity
predictor supervised only by speaking/listening labels.

\paragraph{Decoding and streaming}
\label{sec:training-streaming}
Both branches update per complete interval and reuse adapted conditions throughout
generation. \emph{Codec-Defined Routing} combines main expression and neck latents
with the separate jaw latent. Expression and neck share the main coordination
component, while jaw uses its own. The frozen \emph{Region-wise Codec Decoder}
decodes the avatar motion:
\begin{equation}
  \begin{aligned}
    \mathcal{R}_g^{\mathrm{gen}}&=
      \left(
        [\widehat{\vu}_g^{\mathrm{main},e},\widehat{\vu}_g^{\mathrm{main},c}],
        [\widehat{\vu}_g^{\mathrm{main},n},\widehat{\vu}_g^{\mathrm{main},c}],
        \widehat{\vu}_g^{\mathrm{jaw}}
      \right),\\
    \mX_g&=D_\psi((\mathcal{R}_k^{\mathrm{gen}})_{k\leq g})
       \in\R^{F_g\times106}.
  \end{aligned}
  \label{eq:codec-defined-routing}
\end{equation}
DCP updates fast weights for each complete interval while offline-trained
parameters stay fixed. Main weights guide current and later motion and reset between
conversations. Jaw updates last one interval. Algorithm~\ref{alg:streaming-inference}
gives the full streaming procedure, including incomplete final intervals.

\section{\benchname}
\label{sec:interhead-bench}

\benchname{} provides data and an evaluation protocol for interactive 3D head
generation. Its construction framework, \factoryname{}, unifies two recording
formats. The resulting benchmark supports comparisons of motion accuracy,
conversational behavior, and perceptual quality.

\begin{figure}[t]
    \centering
    \includegraphics[width=1\linewidth]{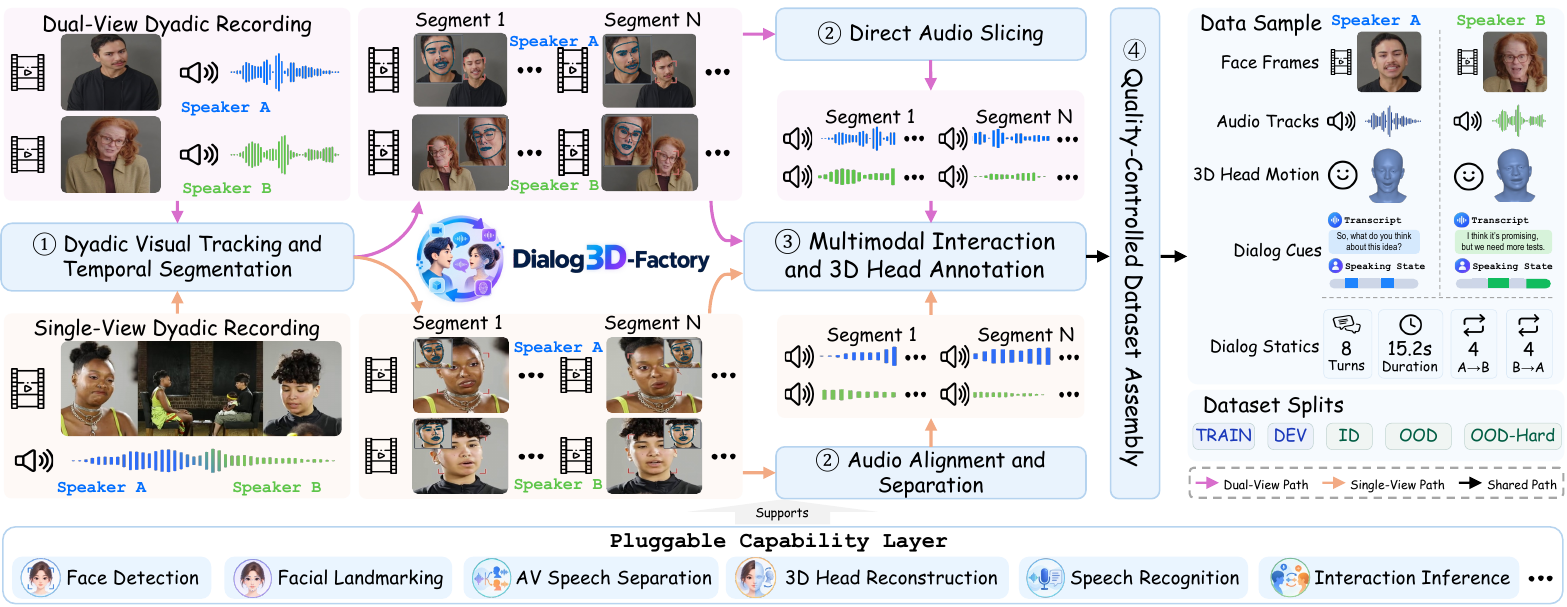}
    \caption{\factoryname{} processes dual-view recordings with separate audio and
    single-view recordings with mixed audio. A shared workflow extracts face
    video, participant speech, FLAME motion, transcripts, speaking and
    interaction states, and turn statistics on a common timeline.}
    \label{fig:dialog3d-factory}
\end{figure}

\paragraph{Dataset construction framework}
\factoryname{} supports two recording formats. \emph{Dual-view} data
provide separate video and audio for each participant. \emph{Single-view}
recordings contain both participants in one video with mixed audio.
The four steps in Figure~\ref{fig:dialog3d-factory} unify both formats.

\textbf{\ding{182}~Track and segment.} We standardize the media and track one
face per dual-view stream or two faces in a single-view stream. We split at
unrecoverable tracking gaps and discard short segments.
\textbf{\ding{183}~Recover participant audio.} For each segment, we slice the
separate dual-view audio tracks, or separate the single-view mixture and
match each speech track to its face.
\textbf{\ding{184}~Annotate motion and interaction.} From these participant
streams, we reconstruct framewise FLAME motion, transcribe speech, and infer
speaking states. Combining the two participants' states gives interaction
states and turn statistics.
\textbf{\ding{185}~Validate and assemble.} We check timing and required outputs
before assigning accepted segments to dataset splits. Processing components
can be replaced while keeping the same output format. Details are provided in Appendix~\ref{app:dataset-construction}.

\begin{table*}[t]
\centering

\caption{
Generation-time conditioning context and parameter sizes of the compared
methods. U and A denote User and Avatar, respectively. Parameter counts are
reported in millions (M).
}
\label{tab:baseline-contexts}

\setlength{\tabcolsep}{4pt}
\renewcommand{\arraystretch}{1.08}
\footnotesize

\resizebox{\linewidth}{!}{%
\begin{tabular}{
c
c
!{\color{TableSeparator}\vrule width 0.45pt}
*{4}{c}
!{\color{TableSeparator}\vrule width 0.45pt}
*{2}{c}
}

\toprule

\multicolumn{1}{c}{%
  \multirow[c]{2}{*}[-0.5ex]{\textbf{Method}}%
}
&
\multicolumn{1}{c!{\color{TableSeparator}\vrule width 0.45pt}}{%
  \multirow[c]{2}{*}[-0.5ex]{\textbf{Venue}}%
}
&
\multicolumn{4}{c!{\color{TableSeparator}\vrule width 0.45pt}}{%
  \textbf{Context}%
}
&
\multicolumn{2}{c}{\textbf{Parameter Size}}
\\

\cmidrule(l{1.5pt}r{1.5pt}){3-6}
\cmidrule(l{1.5pt}r{1.5pt}){7-8}

&
&
\textbf{U-Video}
&
\textbf{U-FLAME}
&
\textbf{U-Audio}
&
\textbf{A-Audio}
&
\textbf{Total (M)}
&
\textbf{Trainable (M)}
\\

\midrule

\MethodRef{DiffPoseTalk}{sun2024diffposetalk}
& TOG '24
& \ContextNo & \ContextNo & \ContextNo & \ContextYes
& 129.32 & 110.55
\\

\MethodRef{ARTalk}{chu2025artalk}
& SIGGRAPH Asia '25
& \ContextNo & \ContextNo & \ContextNo & \ContextYes
& 382.87 & 37.89
\\

\MethodRef{UniLS}{chu2026unils}
& CVPR '26
& \ContextNo & \ContextNo & \ContextYes & \ContextYes
& 422.88 & 27.38
\\

\MethodRef{DualTalk}{peng2025dualtalk}
& CVPR '25
& \ContextNo & \ContextYes & \ContextYes & \ContextYes
& 647.27 & 638.85
\\

\rowcolor{black!5}
\MethodOurs
& This work
& \ContextYes & \ContextNo & \ContextYes & \ContextYes
& 159.57 & 54.82
\\

\bottomrule

\end{tabular}%
}
\end{table*}

\begin{wrapfigure}{r}{0.5\textwidth}
    \centering
    \includegraphics[width=\linewidth]{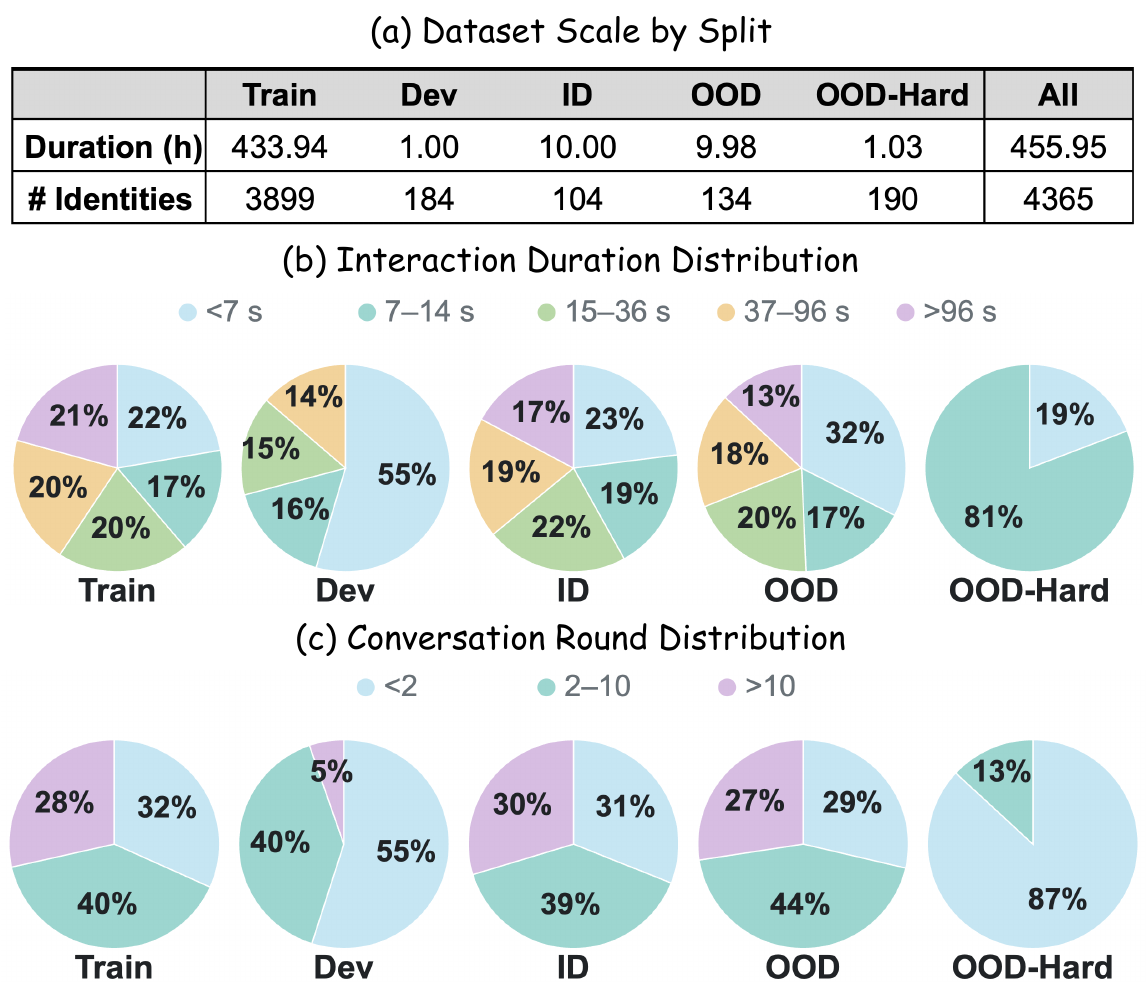}
    \caption{\benchname{} composition by split, including scale and the
    distributions of interaction duration and conversational turns.}
    \label{fig:dataset-stats}
\end{wrapfigure}

\paragraph{Dataset composition and statistics}
We build Train, Dev, ID, and OOD from dual-view Seamless Interaction
recordings~\citep{agrawal2025seamless}, and OOD-Hard from single-view RealTalk
recordings~\citep{geng2023affective}.
ID holds out samples while allowing overlap with Train in participants or
recorded interactions. OOD holds out both participants and interactions.
OOD-Hard tests transfer to a different recording domain. Together, the splits
contain 455.95 hours of interaction and 4,365 source participant IDs.
Figure~\ref{fig:dataset-stats} shows their scale, durations, and turn counts.

\paragraph{Baselines}
To compare methods on these data, we adapt four baselines to predict the same
106-D Avatar FLAME motion. Their inputs follow the task definition in
Section~\ref{sec:preliminaries} and are listed in
Table~\ref{tab:baseline-contexts}. DiffPoseTalk and ARTalk use Avatar audio.
UniLS adds User audio, and DualTalk also uses precomputed user FLAME motion.
\methodname{} instead uses causally available user face frames
and both audio streams.
These input differences guide our interpretation of the results.

\paragraph{Quantitative evaluation metrics}\label{sec:bench-metrics}
We combine metrics from prior
work~\citep{richard2021meshtalk,xing2023codetalker,peng2025dualtalk,chu2026unils}
in \textbf{parameter space} and \textbf{mesh space} for a shared FLAME evaluation.
\textbf{\ding{182}~Parameter space.} MSE measures framewise error, and FD compares motion
means and covariances. P-FD compares user--avatar joint statistics, while rPCC
measures correlation error. SID measures diversity of generated motion across
reference clusters. PDD and JDD measure neck and jaw amplitude errors.
\textbf{\ding{183}~Mesh space.} With neutral FLAME identity and zero neck rotation,
LVE measures lip error, MHD measures full-mesh error, and FDD compares
upper-face motion amplitude.
We report all-frame and avatar speaking/listening scores. Lower is better
except for SID, interpreted alongside accuracy.
Appendix~\ref{app:companion-metrics} provides details.

\paragraph{Perceptual evaluation protocol}\label{sec:bench-perceptual}
We assess perceived quality during interaction through a blind,
forced-choice A/B study. Raters compare lip synchronization, motion
naturalness, turn-taking coherence, audiovisual responsiveness, and overall
conversational realism. These criteria cover both the animation itself and
its fit to the interaction. Study details appear in
Appendix~\ref{app:perceptual-evaluation}.

\section{Experimental Results}

\begin{table*}[t]
\centering

\caption{\textbf{Parameter-space comparison.} Speaking and listening results. Green bold
and blue underlined values mark the best and second best in each split and
state. Lower is better except SID.}
\label{tab:parameter-space-results}

\setlength{\tabcolsep}{2.2pt}
\renewcommand{\arraystretch}{1.08}
\footnotesize

\resizebox{\linewidth}{!}{%
\begin{tabular}{
Y
!{\color{TableSeparator}\vrule width 0.45pt}
*{17}{Z}
}

\toprule


\multicolumn{1}{
  c!{\color{TableSeparator}\vrule width 0.45pt}
}{%
  \multirow[c]{2}{*}[-2.5mm]{\textbf{Method}}%
}
&
\multicolumn{3}{c}{\textbf{FD} $\downarrow$}
&
\multicolumn{3}{c}{\textbf{P-FD} $\downarrow$}
&
\multicolumn{3}{c}{\textbf{MSE} $\downarrow$}
&
\multicolumn{3}{c}{\textbf{rPCC} $\downarrow$}
&
\multicolumn{3}{c}{\textbf{SID} $\uparrow$}
&
\multicolumn{1}{c}{%
  \multirow[c]{2}{*}[-2.5mm]{%
    \textbf{PDD} $\downarrow$%
  }%
}
&
\multicolumn{1}{c}{%
  \multirow[c]{2}{*}[-2.5mm]{%
    \textbf{JDD} $\downarrow$%
  }%
}
\\


\cmidrule(l{1.5pt}r{1.5pt}){2-4}
\cmidrule(l{1.5pt}r{1.5pt}){5-7}
\cmidrule(l{1.5pt}r{1.5pt}){8-10}
\cmidrule(l{1.5pt}r{1.5pt}){11-13}
\cmidrule(l{1.5pt}r{1.5pt}){14-16}


&
\PlainMetricHead{EXP} 
&
\makecell[c]{
  \textbf{JAW}\\
  {\scriptsize$(\times 10^{3})$}
}
&
\makecell[c]{
  \textbf{POSE}\\
  {\scriptsize$(\times 10^{2})$}
}
&
\PlainMetricHead{EXP} 
&
\makecell[c]{
  \textbf{JAW}\\
  {\scriptsize$(\times 10^{3})$}
}
&
\makecell[c]{
  \textbf{POSE}\\
  {\scriptsize$(\times 10^{2})$}
}
&
\makecell[c]{
  \textbf{EXP}\\
  {\scriptsize$(\times 10^{1})$}
}
&
\makecell[c]{
  \textbf{JAW}\\
  {\scriptsize$(\times 10^{3})$}
}
&
\makecell[c]{
  \textbf{POSE}\\
  {\scriptsize$(\times 10^{2})$}
}
&
\makecell[c]{
  \textbf{EXP}\\
  {\scriptsize$(\times 10^{2})$}
}
&
\makecell[c]{
  \textbf{JAW}\\
  {\scriptsize$(\times 10^{1})$}
}
&
\makecell[c]{
  \textbf{POSE}\\
  {\scriptsize$(\times 10^{1})$}
}
&
\PlainMetricHead{EXP}  
&
\PlainMetricHead{JAW}
&
\PlainMetricHead{POSE}
&
&
\\

\midrule


\DatasetBand{In-Distribution Test Set (ID)}

\StateRow{Speaking}

\DataRow
  {\MethodRef{DiffPoseTalk}{sun2024diffposetalk}}
  {48.45 & 22.62 & 24.60}
  {50.30 & 22.93 & 25.47}
  {6.01 & 10.01 & 11.61}
  {20.11 & 1.95 & 3.31}
  {1.84 & 1.36 & \second{1.35}}
  {9.38}
  {2.88}

\DataRow
  {\MethodRef{ARTalk}{chu2025artalk}}
  {21.11 & 3.66 & 18.89}
  {22.12 & 3.75 & 19.52}
  {2.77 & 1.70 & 8.68}
  {\second{18.83} & 1.30 & \second{3.29}}
  {2.18 & 1.69 & \best{1.38}}
  {\best{5.67}}
  {0.95}

\DataRow
  {\MethodRef{UniLS}{chu2026unils}}
  {\second{15.99} & \best{2.36} & 12.31}
  {17.05 & \best{2.45} & 12.76}
  {2.32 & \best{1.30} & 5.36}
  {\best{14.73} & \best{1.07} & 3.37}
  {\best{2.84} & \best{2.20} & 1.13}
  {7.36}
  {\best{0.84}}

\DataRow
  {\MethodRef{DualTalk}{peng2025dualtalk}}
  {16.80 & 4.17 & \second{12.05}}
  {\second{16.97} & 4.23 & \second{12.16}}
  {\best{1.74} & \second{1.59} & \best{4.21}}
  {20.69 & 1.31 & 4.13}
  {0.31 & 0.49 & 0.25}
  {11.37}
  {2.08}

\DataRow
  {\MethodOurs}
  {\best{15.49} & \second{2.40} & \best{9.39}}
  {\best{16.64} & \second{2.56} & \best{9.87}}
  {\second{2.27} & 1.78 & \second{4.61}}
  {19.01 & \second{1.25} & \best{3.17}}
  {\second{2.30} & \second{2.13} & \second{1.35}}
  {\second{7.09}}
  {\second{0.92}}

\StateRow{Listening}

\DataRow
  {\MethodRef{DiffPoseTalk}{sun2024diffposetalk}}
  {46.68 & 27.37 & 23.93}
  {48.52 & 27.69 & 24.88}
  {5.79 & 11.75 & 11.22}
  {\second{19.78} & 2.02 & \second{3.06}}
  {1.79 & 1.33 & \best{1.39}}
  {9.88}
  {3.03}

\DataRow
  {\MethodRef{ARTalk}{chu2025artalk}}
  {24.74 & 4.40 & 24.21}
  {25.70 & 4.50 & 24.86}
  {2.98 & 1.97 & 9.94}
  {23.47 & \second{1.76} & 3.42}
  {1.41 & 1.33 & 0.91}
  {\best{5.99}}
  {1.34}

\DataRow
  {\MethodRef{UniLS}{chu2026unils}}
  {18.61 & \second{3.60} & 13.07}
  {19.74 & \second{3.73} & 13.53}
  {2.50 & 2.00 & 5.45}
  {\best{17.34} & \best{1.65} & 3.25}
  {\best{1.99} & \second{1.65} & 0.95}
  {7.08}
  {\best{1.10}}

\DataRow
  {\MethodRef{DualTalk}{peng2025dualtalk}}
  {\second{17.64} & 5.07 & \second{11.06}}
  {\best{17.78} & 5.11 & \second{11.14}}
  {\best{1.82} & \best{1.95} & \best{3.88}}
  {24.94 & 2.01 & 3.65}
  {0.33 & 0.54 & 0.28}
  {10.31}
  {1.99}

\DataRow
  {\MethodOurs}
  {\best{17.01} & \best{3.17} & \best{8.99}}
  {\second{18.13} & \best{3.33} & \best{9.46}}
  {\second{2.30} & \best{1.95} & \second{4.18}}
  {22.45 & 1.78 & \best{2.88}}
  {\second{1.95} & \best{1.87} & \second{1.29}}
  {\second{6.97}}
  {\second{1.23}}


\DatasetBand{Out-of-Distribution Test Set (OOD)}

\StateRow{Speaking}

\DataRow
  {\MethodRef{DiffPoseTalk}{sun2024diffposetalk}}
  {47.11 & 22.89 & 31.18}
  {48.81 & 23.11 & 32.10}
  {5.77 & 9.58 & 13.90}
  {22.16 & 1.88 & 3.66}
  {1.78 & 1.39 & \second{1.27}}
  {9.33}
  {3.06}

\DataRow
  {\MethodRef{ARTalk}{chu2025artalk}}
  {21.82 & 2.40 & 22.86}
  {22.77 & 2.46 & 23.52}
  {2.75 & \second{1.14} & 9.82}
  {21.17 & 1.43 & \second{3.63}}
  {2.13 & 1.82 & 1.25}
  {\best{5.68}}
  {\second{0.68}}

\DataRow
  {\MethodRef{UniLS}{chu2026unils}}
  {\second{15.41} & \best{2.14} & 13.58}
  {16.37 & \best{2.21} & 14.00}
  {2.15 & \best{1.06} & 5.68}
  {\best{14.70} & \best{1.10} & 3.69}
  {\best{2.69} & \best{2.15} & 1.02}
  {7.00}
  {\best{0.65}}

\DataRow
  {\MethodRef{DualTalk}{peng2025dualtalk}}
  {15.80 & 3.23 & \second{12.86}}
  {\second{15.91} & 3.26 & \second{12.94}}
  {\best{1.61} & 1.20 & \best{4.41}}
  {19.16 & 1.42 & 4.21}
  {0.22 & 0.43 & 0.17}
  {11.10}
  {1.64}

\DataRow
  {\MethodOurs}
  {\best{14.10} & \second{2.24} & \best{9.46}}
  {\best{15.21} & \second{2.35} & \best{9.94}}
  {\second{2.05} & 1.45 & \second{4.59}}
  {\second{15.32} & \second{1.24} & \best{3.17}}
  {\second{2.33} & \second{2.06} & \best{1.35}}
  {\second{6.81}}
  {0.73}

\StateRow{Listening}

\DataRow
  {\MethodRef{DiffPoseTalk}{sun2024diffposetalk}}
  {46.78 & 25.23 & 31.53}
  {48.37 & 25.46 & 32.50}
  {5.64 & 10.43 & 13.63}
  {22.08 & 2.08 & \second{3.44}}
  {1.73 & 1.34 & \second{1.25}}
  {10.33}
  {3.22}

\DataRow
  {\MethodRef{ARTalk}{chu2025artalk}}
  {22.88 & 3.01 & 29.52}
  {23.75 & 3.09 & 30.20}
  {2.71 & \best{1.36} & 11.57}
  {24.57 & \second{1.84} & 3.70}
  {1.38 & 1.39 & 0.82}
  {\best{6.07}}
  {0.99}

\DataRow
  {\MethodRef{UniLS}{chu2026unils}}
  {17.53 & \second{2.97} & 15.60}
  {18.51 & \second{3.07} & 16.05}
  {2.26 & 1.54 & 6.17}
  {\best{16.63} & \best{1.72} & 3.69}
  {\second{1.89} & \second{1.61} & 0.81}
  {6.93}
  {\best{0.87}}

\DataRow
  {\MethodRef{DualTalk}{peng2025dualtalk}}
  {\second{16.04} & 3.86 & \second{12.97}}
  {\second{16.13} & 3.88 & \second{13.03}}
  {\best{1.64} & \second{1.43} & \best{4.45}}
  {22.69 & 2.13 & 3.82}
  {0.24 & 0.44 & 0.19}
  {10.12}
  {1.62}

\DataRow
  {\MethodOurs}
  {\best{14.61} & \best{2.37} & \best{9.90}}
  {\best{15.70} & \best{2.49} & \best{10.45}}
  {\second{1.99} & 1.47 & \second{4.52}}
  {\second{18.92} & 1.96 & \best{3.04}}
  {\best{2.05} & \best{1.83} & \best{1.29}}
  {\second{6.61}}
  {\second{0.89}}


\DatasetBand{Hard Out-of-Distribution Test Set (OOD-Hard)}

\StateRow{Speaking}

\DataRow
  {\MethodRef{DiffPoseTalk}{sun2024diffposetalk}}
  {54.15 & 23.00 & 26.81}
  {55.22 & 23.16 & 27.13}
  {6.07 & 9.13 & 10.30}
  {29.85 & 2.22 & \second{5.56}}
  {1.59 & 1.16 & \second{0.98}}
  {8.51}
  {2.65}

\DataRow
  {\MethodRef{ARTalk}{chu2025artalk}}
  {25.36 & 5.59 & 40.46}
  {26.01 & 5.65 & 40.72}
  {2.89 & 2.24 & 14.32}
  {25.78 & 2.15 & 6.07}
  {1.23 & 0.88 & 0.46}
  {4.26}
  {0.90}

\DataRow
  {\MethodRef{UniLS}{chu2026unils}}
  {20.87 & 3.84 & 26.15}
  {21.64 & 3.91 & 26.38}
  {2.56 & 1.88 & 9.33}
  {19.01 & 2.28 & 6.04}
  {\second{1.83} & \second{1.58} & 0.55}
  {\second{4.01}}
  {\second{0.81}}

\DataRow
  {\MethodRef{DualTalk}{peng2025dualtalk}}
  {\second{13.83} & \second{3.17} & \second{8.77}}
  {\second{13.89} & \second{3.18} & \second{8.81}}
  {\best{1.39} & \best{1.10} & \best{2.97}}
  {\second{16.73} & \second{1.49} & 6.46}
  {0.08 & 0.21 & 0.13}
  {5.70}
  {1.48}

\DataRow
  {\MethodOurs}
  {\best{12.72} & \best{2.47} & \best{8.14}}
  {\best{13.50} & \best{2.54} & \best{8.40}}
  {\second{1.75} & \second{1.39} & \second{3.46}}
  {\best{15.44} & \best{1.34} & \best{5.45}}
  {\best{2.36} & \best{1.93} & \best{1.17}}
  {\best{3.38}}
  {\best{0.68}}

\StateRow{Listening}

\DataRow
  {\MethodRef{DiffPoseTalk}{sun2024diffposetalk}}
  {51.26 & 25.36 & 27.29}
  {52.49 & 25.56 & 27.69}
  {5.77 & 9.74 & 10.35}
  {30.12 & 2.17 & \second{5.25}}
  {\second{1.60} & 1.11 & \second{1.10}}
  {10.56}
  {3.16}

\DataRow
  {\MethodRef{ARTalk}{chu2025artalk}}
  {22.59 & 4.00 & 35.94}
  {23.34 & 4.08 & 36.27}
  {2.58 & 1.64 & 12.80}
  {28.61 & \second{2.13} & 5.80}
  {1.26 & 1.18 & 0.56}
  {5.84}
  {\second{0.95}}

\DataRow
  {\MethodRef{UniLS}{chu2026unils}}
  {22.76 & 3.92 & 22.61}
  {23.70 & 4.03 & 22.88}
  {2.70 & 1.84 & 8.10}
  {22.93 & 2.86 & 5.86}
  {1.47 & \second{1.29} & 0.58}
  {\second{4.63}}
  {1.05}

\DataRow
  {\MethodRef{DualTalk}{peng2025dualtalk}}
  {\second{13.50} & \second{3.25} & \second{8.61}}
  {\best{13.57} & \second{3.28} & \second{8.65}}
  {\best{1.36} & \best{1.14} & \best{2.92}}
  {\second{18.77} & 2.44 & 6.02}
  {0.13 & 0.22 & 0.13}
  {4.81}
  {1.30}

\DataRow
  {\MethodOurs}
  {\best{13.39} & \best{2.54} & \best{8.25}}
  {\second{14.25} & \best{2.63} & \best{8.48}}
  {\second{1.75} & \second{1.31} & \second{3.34}}
  {\best{16.82} & \best{1.84} & \best{4.98}}
  {\best{2.09} & \best{1.69} & \best{1.11}}
  {\best{3.85}}
  {\best{0.85}}

\bottomrule

\end{tabular}%
}

\end{table*}

\begin{table*}[t]
\centering

\caption{\textbf{Mesh-space comparison.} Speaking/listening results. LVE and MHD are in
millimeters. Green bold and blue underline denote the best and second best per
split and state. Lower is better.}
\label{tab:mesh-space-results}

\setlength{\tabcolsep}{1.8pt}
\renewcommand{\arraystretch}{1.08}
\footnotesize

\resizebox{\linewidth}{!}{%
\begin{tabular}{
Y
!{\color{TableSeparator}\vrule width 0.45pt}
*{6}{c}
!{\color{TableSeparator}\vrule width 0.45pt}
*{6}{c}
!{\color{TableSeparator}\vrule width 0.45pt}
*{6}{c}
}
\toprule
& \multicolumn{6}{c!{\color{TableSeparator}\vrule width 0.45pt}}{\textbf{ID}}
& \multicolumn{6}{c!{\color{TableSeparator}\vrule width 0.45pt}}{\textbf{OOD}}
& \multicolumn{6}{c}{\textbf{OOD-Hard}} \\
\cmidrule(l{1.5pt}r{1.5pt}){2-7}
\cmidrule(l{1.5pt}r{1.5pt}){8-13}
\cmidrule(l{1.5pt}r{1.5pt}){14-19}
\multicolumn{1}{c!{\color{TableSeparator}\vrule width 0.45pt}}{\textbf{Method}}
& \multicolumn{3}{c}{\textbf{Speaking}}
& \multicolumn{3}{c!{\color{TableSeparator}\vrule width 0.45pt}}{\textbf{Listening}}
& \multicolumn{3}{c}{\textbf{Speaking}}
& \multicolumn{3}{c!{\color{TableSeparator}\vrule width 0.45pt}}{\textbf{Listening}}
& \multicolumn{3}{c}{\textbf{Speaking}}
& \multicolumn{3}{c}{\textbf{Listening}} \\
\cmidrule(l{1.5pt}r{1.5pt}){2-4}
\cmidrule(l{1.5pt}r{1.5pt}){5-7}
\cmidrule(l{1.5pt}r{1.5pt}){8-10}
\cmidrule(l{1.5pt}r{1.5pt}){11-13}
\cmidrule(l{1.5pt}r{1.5pt}){14-16}
\cmidrule(l{1.5pt}r{1.5pt}){17-19}
& \textbf{FDD} $\downarrow$ & \textbf{LVE} $\downarrow$ & \textbf{MHD} $\downarrow$
& \textbf{FDD} $\downarrow$ & \textbf{LVE} $\downarrow$ & \textbf{MHD} $\downarrow$
& \textbf{FDD} $\downarrow$ & \textbf{LVE} $\downarrow$ & \textbf{MHD} $\downarrow$
& \textbf{FDD} $\downarrow$ & \textbf{LVE} $\downarrow$ & \textbf{MHD} $\downarrow$
& \textbf{FDD} $\downarrow$ & \textbf{LVE} $\downarrow$ & \textbf{MHD} $\downarrow$
& \textbf{FDD} $\downarrow$ & \textbf{LVE} $\downarrow$ & \textbf{MHD} $\downarrow$ \\
\midrule

\MethodRef{DiffPoseTalk}{sun2024diffposetalk}
& 39.26 & 18.78 & 4.10 & 40.74 & 19.97 & 4.31
& 38.37 & 18.10 & 4.02 & 40.82 & 18.60 & 4.11
& 37.57 & 17.94 & 3.97 & 44.10 & 18.28 & 4.01
\\

\MethodRef{ARTalk}{chu2025artalk}
& \second{32.97} & 9.33 & 2.21 & \second{35.90} & 11.77 & 2.75
& \second{30.82} & \second{8.18} & 1.98 & 30.92 & 9.85 & 2.32
& \second{20.91} & 10.79 & 2.49 & \second{19.84} & 10.10 & 2.34
\\

\MethodRef{UniLS}{chu2026unils}
& \best{29.21} & \best{7.62} & \best{1.79} & \best{31.11} & \second{9.92} & \second{2.21}
& \best{27.37} & \best{7.29} & \second{1.71} & \best{28.19} & 8.91 & 1.97
& 21.84 & 10.28 & 2.27 & 21.78 & 10.82 & 2.37
\\

\MethodRef{DualTalk}{peng2025dualtalk}
& 68.51 & \second{8.12} & \second{1.84} & 61.57 & \best{9.26} & \best{2.03}
& 64.35 & \best{7.29} & \best{1.65} & 55.63 & \best{8.03} & \best{1.75}
& 42.83 & \best{7.65} & \best{1.67} & 38.31 & \best{7.64} & \best{1.67}
\\

\MethodOurs
& 37.63 & 9.35 & 2.10 & 36.23 & 10.08 & 2.22
& 33.96 & 8.55 & 1.90 & \second{29.68} & \second{8.84} & \second{1.93}
& \best{19.09} & \second{8.83} & \second{1.92} & \best{18.13} & \second{8.76} & \second{1.89}
\\

\bottomrule
\end{tabular}%
}
\end{table*}

\paragraph{Our gains span speaking, listening, and challenging distribution shifts}
Our expression and neck FD are lowest in both states on every split
(Table~\ref{tab:parameter-space-results}). On OOD-Hard, jaw FD also leads,
and speaking/listening FDD reaches 19.09/18.13 versus 20.91/19.84 for
ARTalk (Table~\ref{tab:mesh-space-results}). These results show closer
agreement in motion statistics and upper-face amplitude. DualTalk has
lower MSE and LVE/MHD, but lower motion coverage (SID). These point errors
measure agreement with one recorded response, whereas an interaction can
admit several plausible responses. The 86\% preference for our method
against DualTalk on OOD-Hard (Figure~\ref{fig:qualitative-eval}) supports
separating reference agreement from perceived conversational quality.
UniLS retains lower FDD on ID/OOD, revealing a remaining amplitude gap:
matching coefficient statistics does not ensure equally accurate geometric
variation after FLAME decoding. Parameter metrics also assess neck motion,
which the expression-and-jaw meshes omit. Together, the two spaces and
human judgments assess different aspects of conversational motion.
All-frame results in both parameter and mesh spaces appear in
Appendix~\ref{app:all-frame-results}.

\paragraph{Learning throughout a conversation improves expression generation}
On ID, full TTT lowers expression P-FD by 20.4\% against No write, 4.2\%
against Freeze-1, and 8.6\% against No carry
(Table~\ref{tab:ablation-ttt}). Freeze-1 stops persistent updates after
the first, while No carry clears fast state after each group. These
same-checkpoint comparisons support both continued updates and retention
of learned context. Neck P-FD instead increases by 8.9\% against No write.
DCP updates fast weights through context prediction, which does not
directly constrain each region's motion statistics. Across five
equal-duration intervals (Figure~\ref{fig:interval-performance}),
expression P-FD falls by 2.7\% on OOD and 7.1\% on OOD-Hard from first to
last, while all four baselines worsen on OOD. The interval results show
improvement over time, while the matched controls test the contribution of
online learning to motion generation.

\paragraph{Human judgments favor our conversational behavior}
Overall-realism preference over UniLS rises from 58\% on ID to 66\% on OOD
and 72\% on OOD-Hard (Figure~\ref{fig:qualitative-eval}). On OOD-Hard,
preference reaches 90\% against DiffPoseTalk and ARTalk and 86\% against
DualTalk. These judgments favor our behavior under distribution shift,
complementing the speaking and listening examples. Raters judge motion
alongside user video and shared audio, making these preferences
informative about behavior within the conversation. The UniLS trend
remains descriptive because all three uncertainty intervals include equal
preference. These comparisons assess complete methods, while the matched
TTT controls test online adaptation. Additional visual comparisons appear
in Appendix~\ref{app:visual-comparisons}.

\paragraph{The components contribute complementary motion improvements}
With jaw flow disabled in both codecs, regional coding lowers ID
expression and neck P-FD by 1.9\% and 0.8\% relative to unified coding
(Table~\ref{tab:ablation-modules}), supporting separate regional latents.
Jaw P-FD remains slightly higher, so this benefit is region dependent. Jaw
FM is associated with 5.1\% lower jaw P-FD, consistent with its
articulation role. Activity conditioning lowers all six FD/P-FD scores,
including expression P-FD from 17.10 to 16.79, supporting its use in
guiding adapted motion. Jaw FM, Activity, and RGB comparisons are
exploratory across revisions. In Appendix~\ref{app:context-signals}, RGB
improves expression and neck statistics over audio alone on OOD-Hard, but
expression statistics worsen on ID/OOD. Visual context therefore offers
selective benefits, motivating better control of how it guides motion
generation.

\begin{table*}[t]
\centering
\begin{minipage}[t]{0.49\linewidth}
\vspace{0pt}
\centering
\caption{\textbf{TTT inference ablation.} No write disables updates. Freeze-1 stops
persistent updates after the first. No carry updates fast state for each
prediction and then clears it. All controls use the same offline-trained
parameters.}
\label{tab:ablation-ttt}
\setlength{\tabcolsep}{2.2pt}
\renewcommand{\arraystretch}{1.08}
\footnotesize
\resizebox{\linewidth}{!}{%
\begin{tabular}{c*{6}{c}}
\toprule
\multirow[c]{2}{*}[-2.5mm]{\textbf{Variant}} & \multicolumn{3}{c}{\textbf{FD} $\downarrow$} & \multicolumn{3}{c}{\textbf{P-FD} $\downarrow$} \\
\cmidrule(l{1.5pt}r{1.5pt}){2-4}\cmidrule(l{1.5pt}r{1.5pt}){5-7}
 & \PlainMetricHead{EXP} & \makecell[c]{\textbf{JAW}\\{\scriptsize$(\times 10^{3})$}} & \makecell[c]{\textbf{POSE}\\{\scriptsize$(\times 10^{2})$}} & \PlainMetricHead{EXP} & \makecell[c]{\textbf{JAW}\\{\scriptsize$(\times 10^{3})$}} & \makecell[c]{\textbf{POSE}\\{\scriptsize$(\times 10^{2})$}} \\
\midrule
\textbf{Full model} & \best{15.77} & \best{2.64} & 8.89 & \best{16.79} & \best{2.79} & 9.32 \\
No write & 19.91 & 2.67 & 8.03 & 21.09 & 2.81 & 8.56 \\
Freeze-1 & 16.51 & 2.64 & 8.85 & 17.53 & 2.79 & 9.29 \\
No carry & 17.30 & 2.64 & 8.69 & 18.36 & 2.79 & 9.16 \\
\bottomrule
\end{tabular}%
}
\end{minipage}\hfill%
\begin{minipage}[t]{0.49\linewidth}
\vspace{0pt}
\centering
\caption{\textbf{Module ablation.} Unified Codec disables jaw flow and regional
partitioning. w/o Activity removes predicted-state features, supervision, and
innovation gating. Comparisons use independently trained variants.}
\label{tab:ablation-modules}
\setlength{\tabcolsep}{2.2pt}
\renewcommand{\arraystretch}{1.08}
\footnotesize
\resizebox{\linewidth}{!}{%
\begin{tabular}{c*{6}{c}}
\toprule
\multirow[c]{2}{*}[-2.5mm]{\textbf{Variant}} & \multicolumn{3}{c}{\textbf{FD} $\downarrow$} & \multicolumn{3}{c}{\textbf{P-FD} $\downarrow$} \\
\cmidrule(l{1.5pt}r{1.5pt}){2-4}\cmidrule(l{1.5pt}r{1.5pt}){5-7}
 & \PlainMetricHead{EXP} & \makecell[c]{\textbf{JAW}\\{\scriptsize$(\times 10^{3})$}} & \makecell[c]{\textbf{POSE}\\{\scriptsize$(\times 10^{2})$}} & \PlainMetricHead{EXP} & \makecell[c]{\textbf{JAW}\\{\scriptsize$(\times 10^{3})$}} & \makecell[c]{\textbf{POSE}\\{\scriptsize$(\times 10^{2})$}} \\
\midrule
\textbf{Full model} & \best{15.77} & \best{2.64} & \best{8.89} & \best{16.79} & \best{2.79} & \best{9.32} \\
w/o Jaw FM & 15.79 & 2.79 & 8.89 & 16.81 & 2.94 & 9.32 \\
Unified Codec & 16.10 & 2.79 & 8.99 & 17.13 & 2.93 & 9.40 \\
w/o Activity & 16.09 & 2.76 & 8.94 & 17.10 & 2.90 & 9.37 \\
\bottomrule
\end{tabular}%
}
\end{minipage}
\end{table*}

\begin{figure}[t]
  \centering
  \begin{minipage}[t]{0.49\linewidth}
    \vspace{0pt}
    \input{figures/interval-performance}
  \end{minipage}\hfill
  \begin{minipage}[t]{0.49\linewidth}
    \vspace{0pt}
    \input{figures/qualitative-eval}
  \end{minipage}
\end{figure}
\section{Conclusion}

We introduced \methodname{}, a causal generator that learns from user face
video and dyadic audio while generating coordinated speaking and listening
motion. Dyadic Context Prediction enables test-time training without target
motion labels at deployment. Persistent adaptation retains patterns learned
within each conversation, while transient jaw adaptation responds to current
articulation. We introduced \benchname{}, which unifies single-view and
dual-view conversation videos for evaluating speaking and listening across
distributions. Experiments show improved motion statistics and human ratings
under distribution shifts, while matched ablations support continued learning
within conversations. Together, these contributions extend interactive 3D head
generation from conditioning on context to learning from it throughout each
conversation.

\paragraph{Limitations and future work}
Recorded conversations do not capture how users respond to generated motion.
Future work could study continual learning in live interactions, distinguishing
stable expressive habits from temporary emotional responses to decide what to
retain and when to revise it. For psychological support, the key question is
whether adaptation leads to nonverbal responses appropriate to users' changing
needs. Clinician-guided studies could assess perceived support alongside motion
quality and track their relationship throughout longer conversations.

\subsection*{AI use statement}

Generative AI tools helped refine the language and improve readability. The
authors take responsibility for the accuracy and integrity of all content,
scientific claims, and conclusions.

\subsection*{Ethics statement}

All original audiovisual recordings used in this study come from publicly
available datasets. Their use and redistribution remain subject to the source
datasets' licenses and access conditions.

\subsection*{Reproducibility statement}

Appendix~\ref{app:method-details} documents model architecture, training
objectives, and streaming inference. Appendices~\ref{app:dataset-construction}
and~\ref{app:companion-metrics} detail dataset construction and evaluation
metric definitions, respectively.

\subsubsection*{Acknowledgments}

We thank the reviewers and chairs for their time and thoughtful feedback on our work.

\bibliography{iclr2027_conference}
\bibliographystyle{iclr2027_conference}

\appendix

\clearpage
\addtocontents{toc}{\protect\setcounter{tocdepth}{2}}
\begingroup
\edef\AppendixSavedTocDepth{\number\value{tocdepth}}
\renewcommand{\contentsname}{Appendix Contents}
\hypersetup{linktoc=all}

\makeatletter
\renewcommand{\l@section}[2]{%
  \ifnum\c@tocdepth>\z@
    \addvspace{0.8em}%
    \begingroup
      \bfseries
      \@dottedtocline{1}{0em}{2.2em}{#1}{#2}%
    \endgroup
  \fi
}
\renewcommand{\l@subsection}{\@dottedtocline{2}{2.2em}{2.8em}}
\makeatother

\pdfbookmark[1]{Appendix Contents}{appendix.contents}
\tableofcontents
\setcounter{tocdepth}{\AppendixSavedTocDepth}
\endgroup

\clearpage

\section{Metric definitions and properties}
\label{app:companion-metrics}

\paragraph{Inputs and shapes}
For a directed record and evaluation view, $T$ denotes the selected frame
count. Predicted avatar, reference avatar, and aligned user motion are
$\mX,\mY,\mU\in\R^{T\times106}$. For component $c\in\{e,n,j\}$, their slices
$\mX^c,\mY^c,\mU^c\in\R^{T\times d_c}$ contain expression, neck pose, or jaw
pose, with $d_e=100$ and $d_n=d_j=3$. A frame vector is $\vx_t^c\in\R^{d_c}$.
User FLAME motion supplies evaluation context only for P-FD and rPCC. These
parameters are never supplied to \methodname.

\paragraph{Views and aggregation}
Speaking includes avatar speech and overlap, whereas listening includes avatar
listening and silence. All-frame scores are recomputed from every valid frame,
not averaged from these two views. We compute each scalar metric per directed
record, then average equally over records with a defined score. Lower is
better except for SID, whose higher value indicates broader occupancy rather
than fidelity. Table multipliers only rescale the displayed scores for
readability.

\subsection{Parameter-space metrics}

We compute MSE, FD, P-FD, rPCC, and SID following
DualTalk~\citep{peng2025dualtalk}, and adapt PDD and JDD from
UniLS~\citep{chu2026unils}. Shared FLAME coordinates and evaluation
conventions make comparisons consistent across methods, although scores may
differ from original reports.

\paragraph{MSE: aligned coefficient error}
MSE averages squared differences across matching frames and channels. It
measures fidelity to the recorded coefficients, whose scales differ by
component. It does not establish whether another response is perceptually
plausible or appropriate to the conversation:
\begin{equation*}
  D_{\mathrm{MSE}}^c
  =\frac{1}{T d_c}\lVert\mX^c-\mY^c\rVert_F^2.
\end{equation*}

\paragraph{FD: marginal motion statistics}
For $\mZ,\mW\in\R^{T\times d}$, let $\vmu_Z\in\R^d$ denote the mean over
frames and $\mSigma_Z\in\R^{d\times d}$ the sample covariance, set to zero for
$T=1$ and otherwise using denominator $T-1$. FD compares coefficient means and
covariances through the Gaussian moment functional from
FID~\citep{heusel2017ttur}, discarding frame order and higher-order
distributional structure:
\begin{align*}
  \mathcal{F}(\mZ,\mW)
  &=\lVert\vmu_Z-\vmu_W\rVert_2^2
    +\Tr\!\left(\mSigma_Z+\mSigma_W
    -2(\mSigma_Z^{1/2}\mSigma_W\mSigma_Z^{1/2})^{1/2}\right),
  \\
  D_{\mathrm{FD}}^c&=\mathcal{F}(\mX^c,\mY^c).
\end{align*}

\paragraph{P-FD: joint interaction statistics}
Concatenated features $[\mU^c,\mX^c]$ and $[\mU^c,\mY^c]$ have shape
$T\times2d_c$. P-FD compares joint means and covariances, including same-frame
user--avatar cross-covariance. It cannot establish response timing, temporal
direction, or causal influence between participants:
\begin{equation*}
  D_{\mathrm{P\text{-}FD}}^c
  =\mathcal{F}([\mU^c,\mX^c],[\mU^c,\mY^c]).
\end{equation*}

\paragraph{rPCC: global correlation agreement}
Pearson correlation $\rho$ acts on $\operatorname{vec}\mX^c\in\R^{Td_c}$,
which flattens time and channels. rPCC compares predicted and reference
user--avatar correlations. This scalar cannot resolve individual channels,
motion amplitude, lagged responses, or causal influence:
\begin{equation*}
  D_{\mathrm{rPCC}}^c
  =\left|\rho(\operatorname{vec}\mY^c,\operatorname{vec}\mU^c)
  -\rho(\operatorname{vec}\mX^c,\operatorname{vec}\mU^c)\right|.
\end{equation*}

\paragraph{SID: occupancy of target motion cells}
Fixed-seed K-means fits target centroids $\vc_k\in\R^{d_c}$ for each record,
view, and component, with $K_e=40$ and $K_n=K_j=10$. Each predicted frame is
assigned to its nearest centroid, yielding occupancies $\vp\in\R^{K_c}$. SID
measures predicted occupancy spread and evenness, ignoring agreement with
target occupancies, distances to centroids, and visit order:
\begin{equation*}
  a_t=\argmin_k\lVert\vx_t^c-\vc_k\rVert_2^2,
  \qquad
  p_k=\frac1T\sum_{t=1}^T\1_{\{a_t=k\}},
  \qquad
  S_{\mathrm{SID}}^c=-\sum_{k=1}^{K_c}p_k\log_2(p_k+10^{-6}).
\end{equation*}

\paragraph{PDD and JDD: motion amplitude}
For $\mX^c_{:,k}\in\R^T$, $\widehat\sigma_t$ denotes sample standard deviation
with denominator $T-1$. PDD and JDD compare neck and jaw dispersion, ignoring
constant offsets and frame order. Low values do not imply correct velocity,
smoothness, or cross-channel coordination:
\begin{equation*}
  D_{\mathrm{DD}}^c=\frac{100}{d_c}\sum_{k=1}^{d_c}
  \left|\widehat\sigma_t(\mX^c_{:,k})-\widehat\sigma_t(\mY^c_{:,k})\right|,
  \qquad
  \mathrm{PDD}=D_{\mathrm{DD}}^n,\quad \mathrm{JDD}=D_{\mathrm{DD}}^j.
\end{equation*}

\paragraph{Numerical conventions}
For FD and P-FD, rank deficiency or a non-finite matrix square root triggers
$\epsilon I$ stabilization, with $\epsilon=10^{-6}$. We add it to both
covariances inside the square-root term only, retaining their original traces
elsewhere. This can yield tiny negative scores. We omit rPCC when a Pearson
denominator is numerically zero. SID requires at least $K_c$ distinct target
frames and successful codebook fitting. Dispersion metrics require at least
two valid frames in the view.

\subsection{Mesh-space metrics}

Following UniLS~\citep{chu2026unils}, FLAME produces meshes
$\mV^X,\mV^Y\in\R^{T\times V\times3}$ with $V$ vertices in millimeters. We
zero identity, root, eye, and neck coefficients for both meshes. Mesh metrics
compare expression and jaw geometry, while parameter-space metrics evaluate
neck motion.

\paragraph{LVE: largest lip deviation}
Let $\sL\subseteq\{1,\ldots,V\}$ index lip vertices with positions
$\mV_{t,v}^X,\mV_{t,v}^Y\in\R^3$. LVE is the root mean square of the largest
lip-vertex displacement per frame. It emphasizes localized errors and is
sensitive to outliers, but does not directly measure audiovisual
synchronization:
\begin{equation*}
  D_{\mathrm{LVE}}=\left[
    \frac1T\sum_{t=1}^T\max_{v\in\sL}
    \lVert\mV_{t,v}^X-\mV_{t,v}^Y\rVert_2^2
  \right]^{1/2}.
\end{equation*}

\paragraph{MHD: full-mesh reference error}
MHD computes the root mean square error across matching vertices and frames,
in millimeters like LVE. It is not a Hausdorff distance. Its spatial average
measures reference fidelity, but can dilute local errors and cannot establish
perceptual plausibility:
\begin{equation*}
  D_{\mathrm{MHD}}=\left[
    \frac1{TV}\sum_{t=1}^T\sum_{v=1}^V
    \lVert\mV_{t,v}^X-\mV_{t,v}^Y\rVert_2^2
  \right]^{1/2}.
\end{equation*}

\paragraph{FDD: projected upper-face amplitude}
Let $(h_1,\ldots,h_M)$ be the upper-face index list of $M=884$ entries,
including repeats, and $\vone=(1,1,1)^\top\in\R^3$. FDD compares sample
standard deviations of projections $\vone^\top\mV_{t,h_m}\in\R$, which sum the
three coordinates. It ignores constant offsets and temporal order, and
coordinate changes can cancel, so it does not measure full three-dimensional
dynamics:
\begin{equation*}
  D_{\mathrm{FDD}}=\frac{100}{M}\sum_{m=1}^M
  \left|\widehat\sigma_t(\vone^\top\mV_{t,h_m}^X)
  -\widehat\sigma_t(\vone^\top\mV_{t,h_m}^Y)\right|.
\end{equation*}

\subsection{Interpreting the metrics together}

Paired error measures agreement with one recorded response, whereas a
conversation can admit several appropriate responses. To see the distinction,
fix the context and let $Z,Z'\in\R^d$ be independent draws from the same
conditional response distribution, with mean $\mu$ and covariance $\Sigma$.
Predicting the mean and sampling a response give different expected squared
reference errors:
\begin{equation*}
  \E\lVert Z-\mu\rVert_2^2=\Tr(\Sigma),
  \qquad
  \E\lVert Z'-Z\rVert_2^2=2\Tr(\Sigma).
\end{equation*}
A sampled response can incur more expected squared reference error than
predicting the conditional mean. After normalization, these identities apply
to parameter MSE and squared MHD, but not directly to LVE or expected MHD. For
meshes, averaging decoded responses generally differs from decoding mean FLAME
coefficients. Neither construction guarantees a plausible response.

We combine paired fidelity, motion statistics, amplitude, and perceptual
evaluation to assess both agreement with the recorded response and the
perceived overall quality of conversational behavior.

\clearpage

\section{Additional visual comparisons}
\label{app:visual-comparisons}

\begin{figure}[!htbp]
  \centering
  \includegraphics[width=\linewidth]{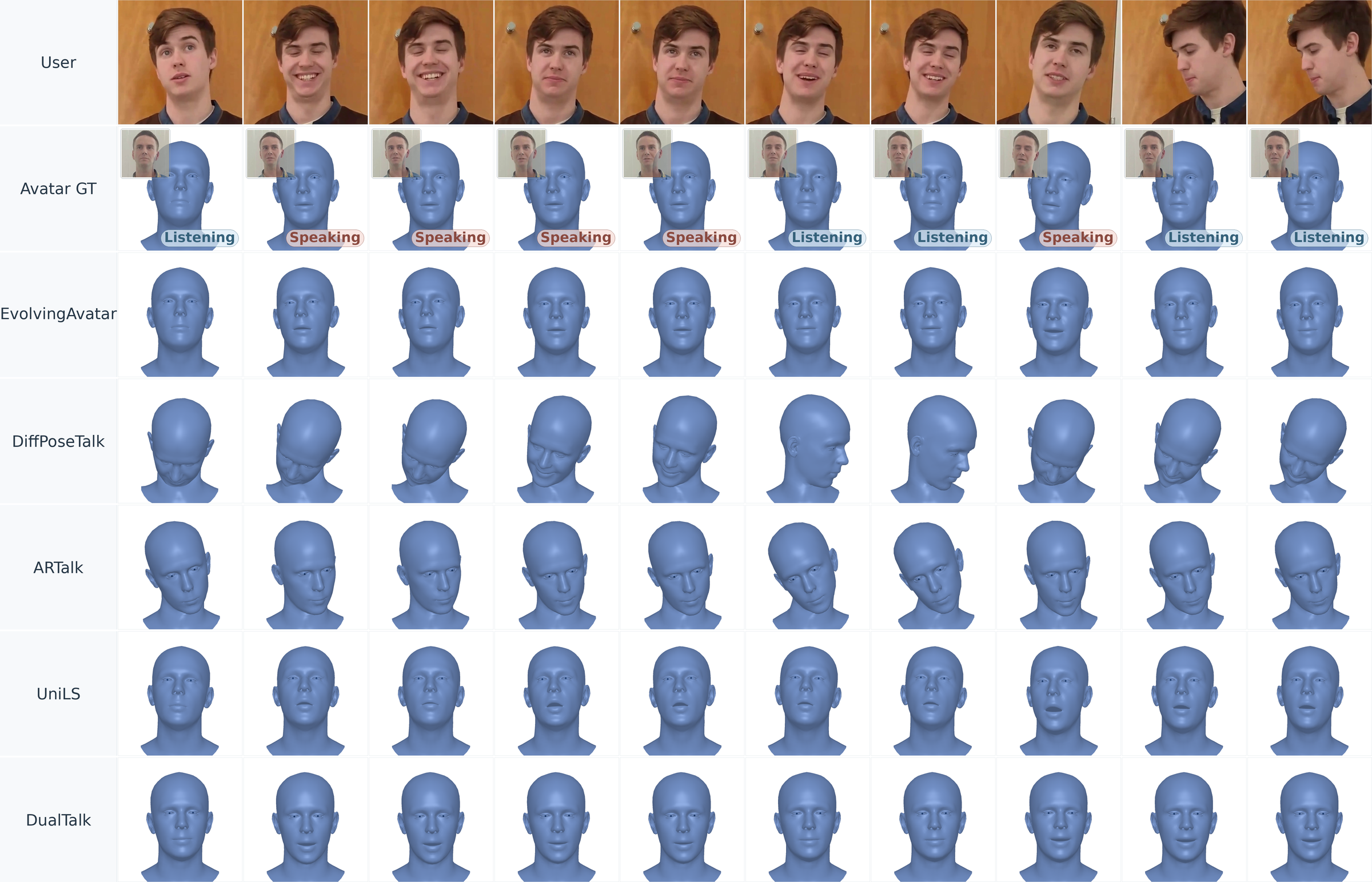}
  \caption{\textbf{Restrained neck and mouth motion across activities.} EvolvingAvatar preserves
  the reference's slightly raised chin and small speaking openings, with a more restrained mouth
  shape in the intervening listening columns. UniLS opens the mouth too widely at several of
  these selected instants, while DiffPoseTalk and ARTalk introduce marked downward or sideways
  neck poses.}
  \label{fig:additional-visual-comparisons}
  \label{fig:visual-case-046}
\end{figure}

\begin{figure}[p]
  \centering
  \includegraphics[width=\linewidth]{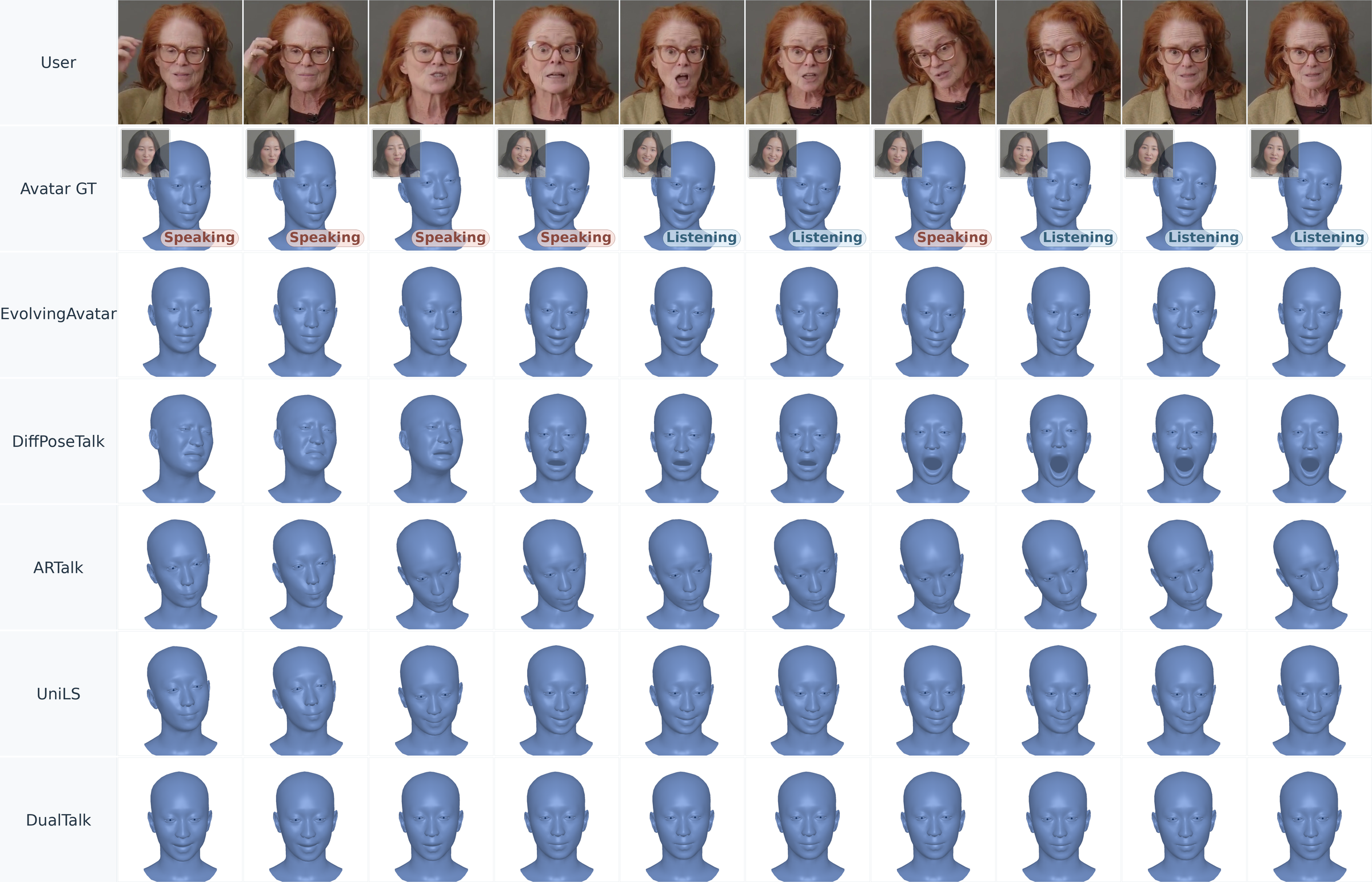}
  \caption{\textbf{Smiling while listening.} EvolvingAvatar preserves the reference's smile and
  moderate mouth opening in columns 4--6, across speaking and listening. UniLS and DualTalk show
  weaker openings there, while DiffPoseTalk exaggerates mouth opening in later sampled listening
  frames.}
  \label{fig:visual-case-001}
\end{figure}

\begin{figure}[p]
  \centering
  \includegraphics[width=\linewidth]{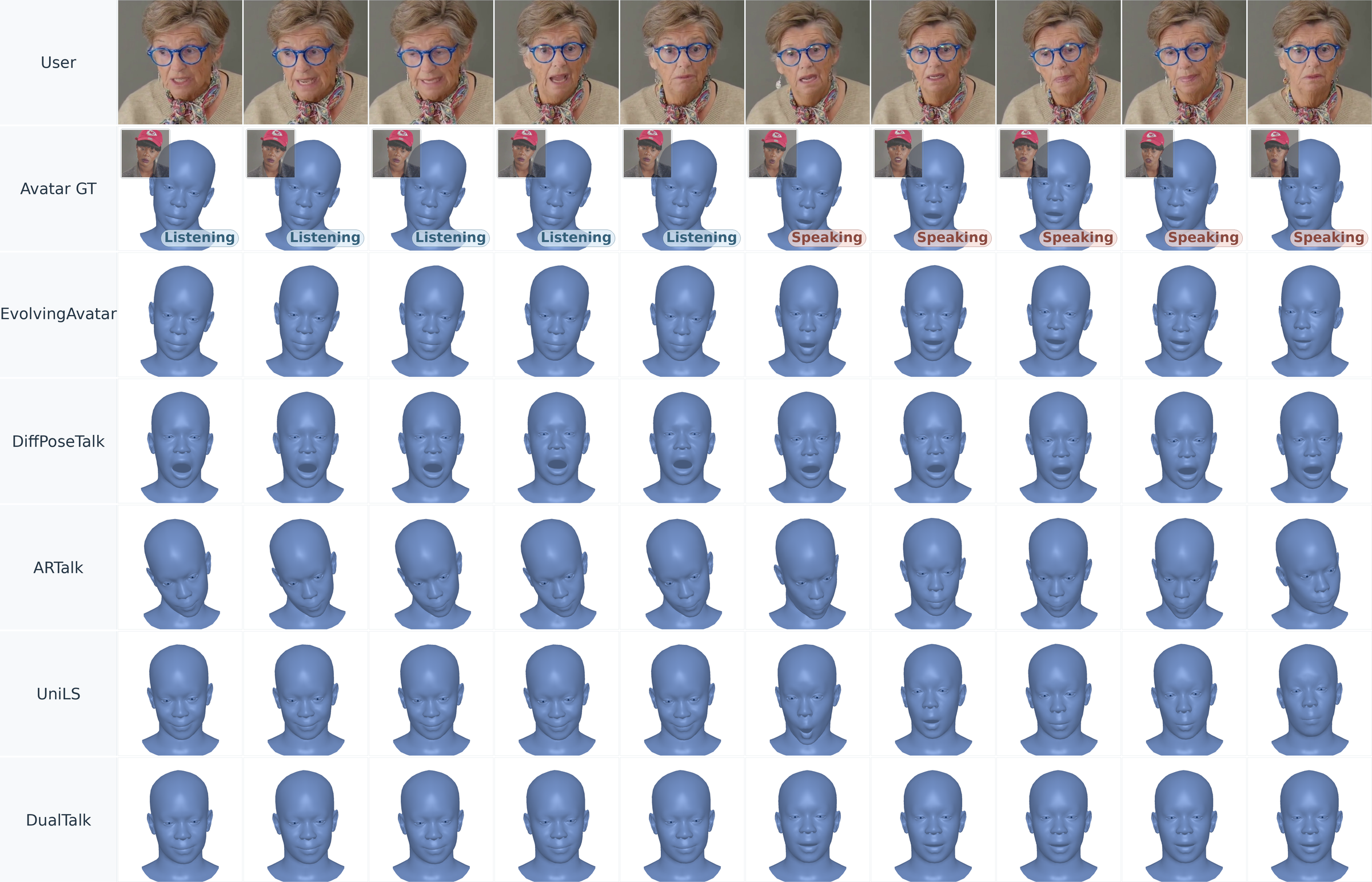}
  \caption{\textbf{From listening to speaking.} EvolvingAvatar combines the tilted listening
  posture in columns 1--5 with the larger speaking mouth openings in columns 7--10. UniLS
  changes the mouth opening less consistently with Avatar GT, while DualTalk largely retains a
  neutral expression.}
  \label{fig:visual-case-003}
\end{figure}

\begin{figure}[p]
  \centering
  \includegraphics[width=\linewidth]{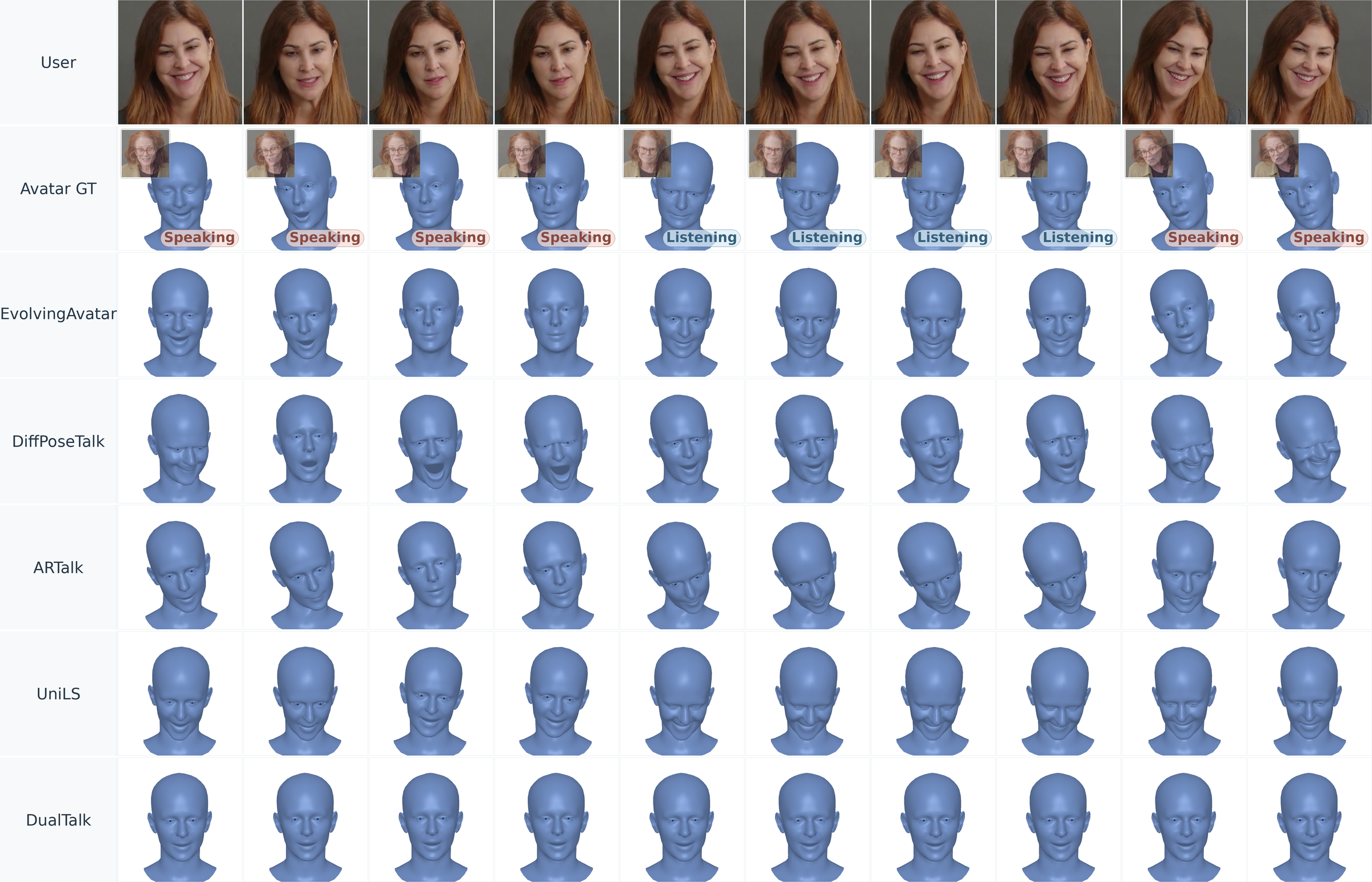}
  \caption{\textbf{Coordinating changes in expression and neck pose.} EvolvingAvatar follows the
  reference's slight downward tilt during listening in columns 5--8 and sideways lean in the
  final speaking frames. UniLS exaggerates the downward tilt during listening, while DualTalk
  remains largely frontal and DiffPoseTalk adds a broad grin where the reference expression is
  restrained.}
  \label{fig:visual-case-033}
\end{figure}

\begin{figure}[p]
  \centering
  \includegraphics[width=\linewidth]{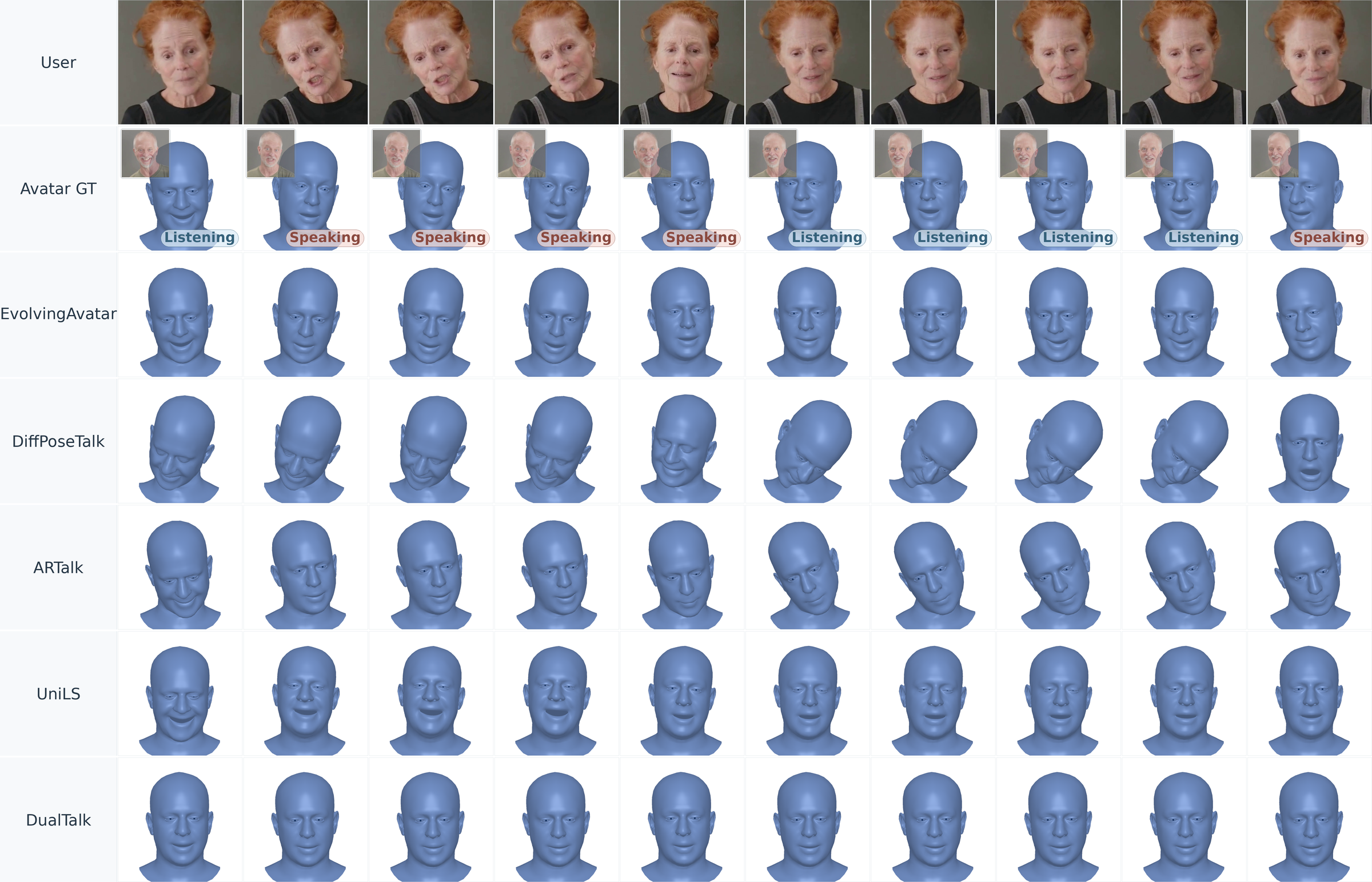}
  \caption{\textbf{Moderate mouth openings and neck motion.} EvolvingAvatar captures the
  speaking openings and tilt in columns 2--4, then retains a listening smile. UniLS exaggerates
  those speaking openings and closes the eyes in columns 6--9, while DiffPoseTalk adds a larger
  sideways lean.}
  \label{fig:visual-case-036}
\end{figure}

\begin{figure}[p]
  \centering
  \includegraphics[width=\linewidth]{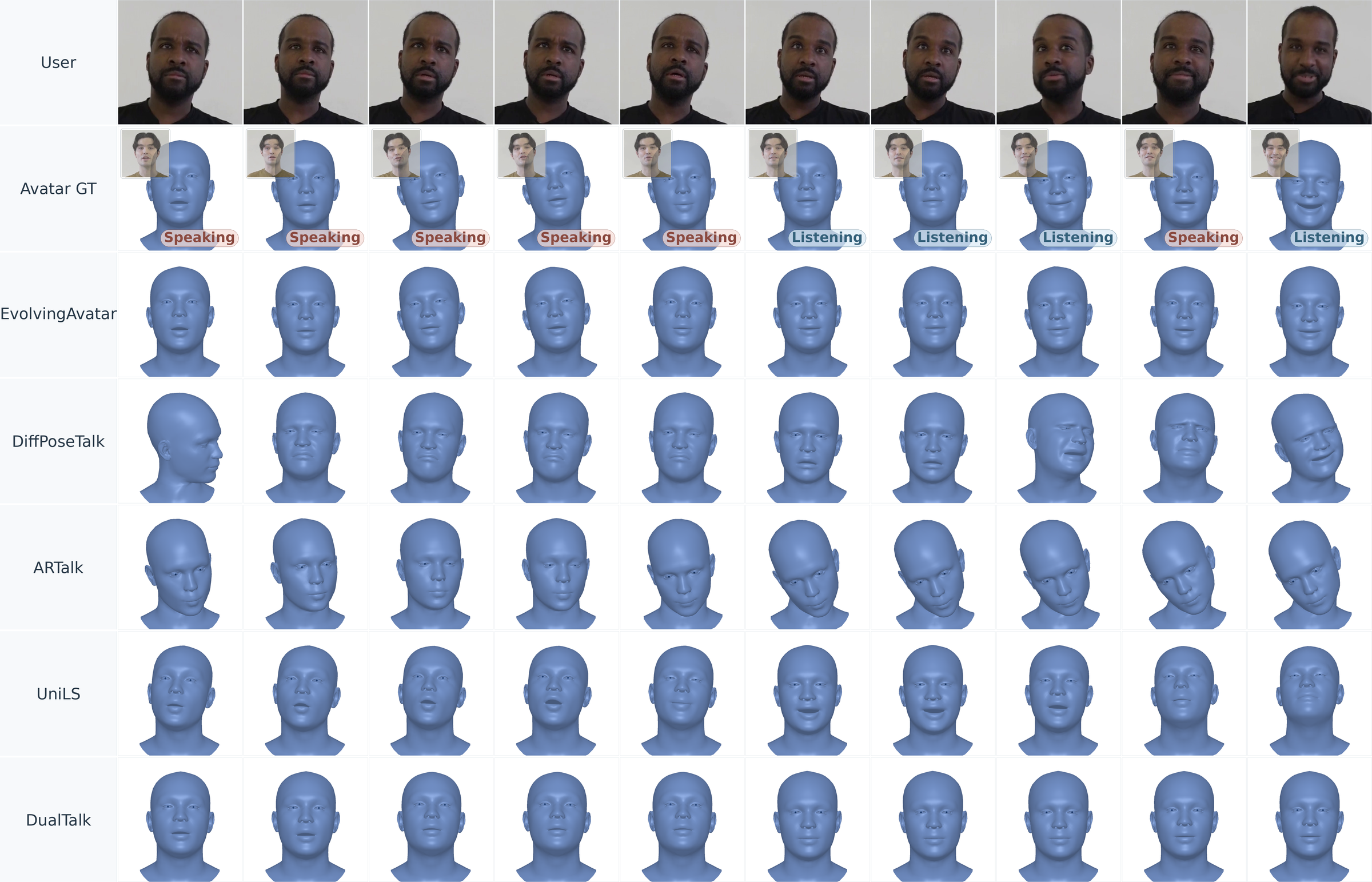}
  \caption{\textbf{Matching neck orientation and articulation together.} EvolvingAvatar retains
  the reference's near-upright orientation and modest changes in mouth opening across the
  selected speaking and listening instants. UniLS exaggerates the upward tilt or mouth opening
  at several instants, while ARTalk introduces a pronounced downward tilt that is absent from
  the reference.}
  \label{fig:visual-case-006}
\end{figure}

\begin{figure}[p]
  \centering
  \includegraphics[width=\linewidth]{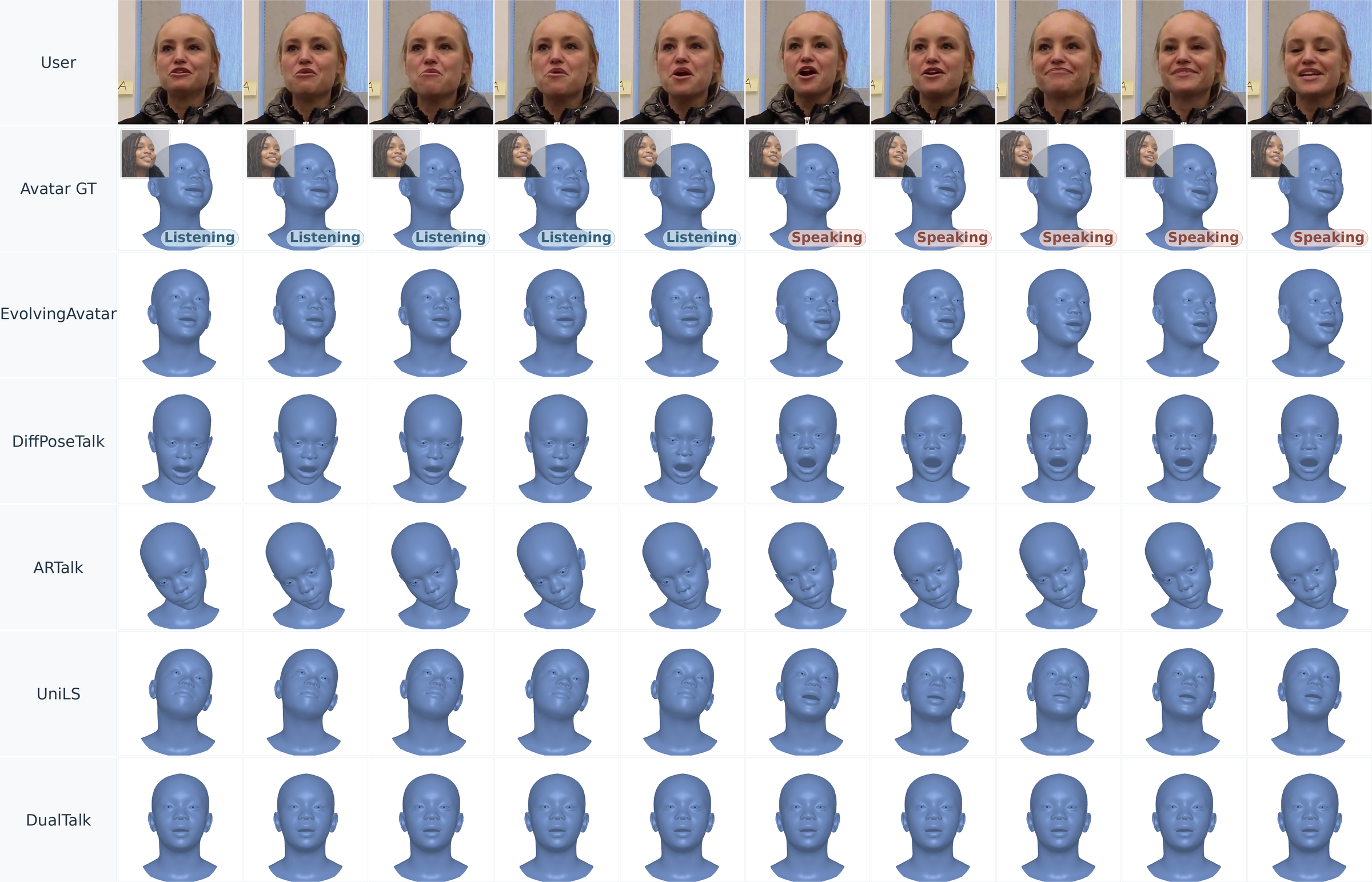}
  \caption{\textbf{An expressive listener.} EvolvingAvatar follows the reference's open-mouth
  smile and upward, sideways orientation during both activities. UniLS largely closes the mouth
  in the first five listening frames, while ARTalk tilts the face downward and DualTalk retains
  a more frontal pose.}
  \label{fig:visual-case-008}
\end{figure}

\begin{figure}[p]
  \centering
  \includegraphics[width=\linewidth]{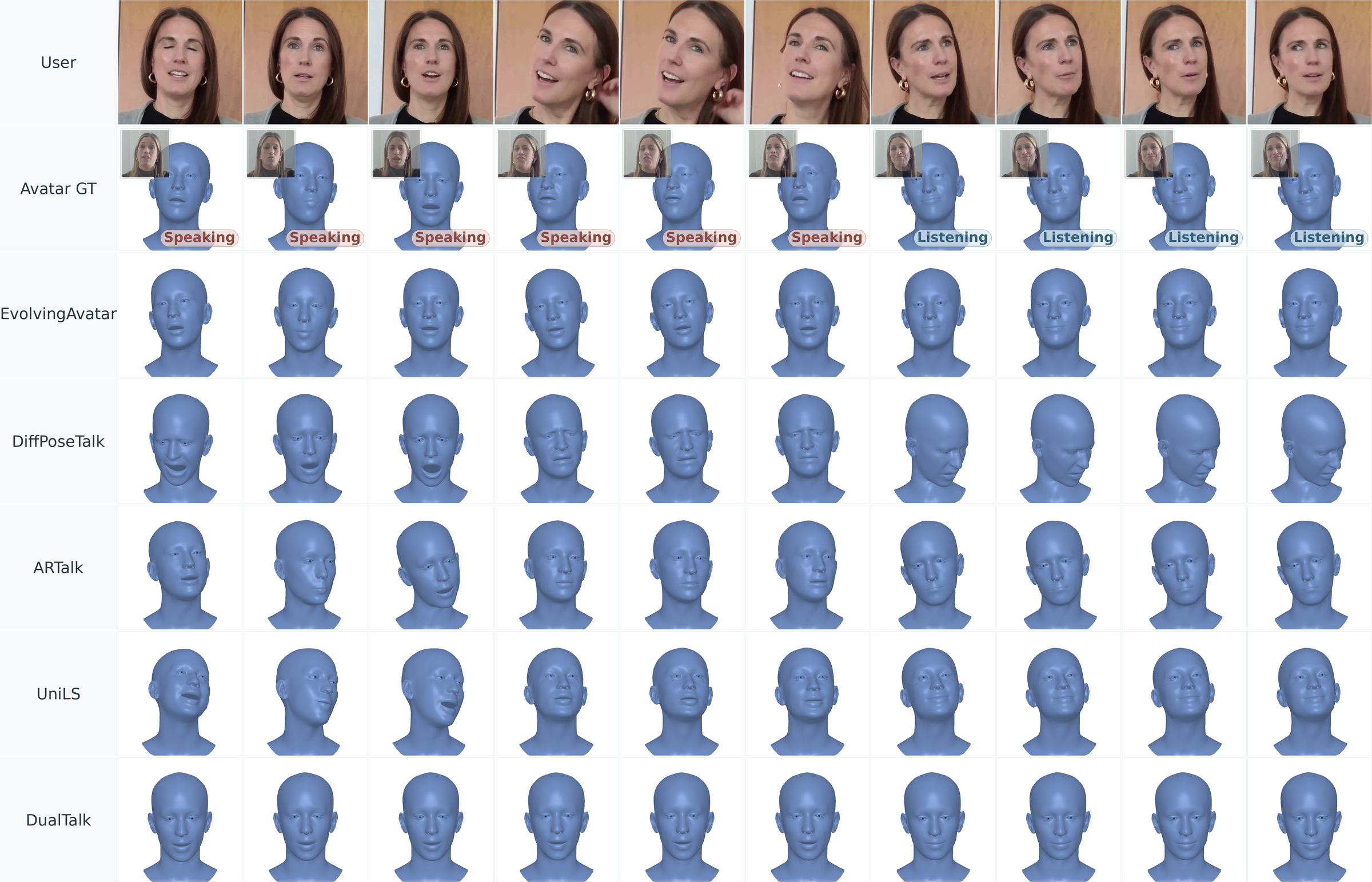}
  \caption{\textbf{From varied articulation to a listening smile.} EvolvingAvatar captures the
  speaking openings in columns 1--6 and closed-mouth listening smile in columns 7--10. UniLS
  exaggerates upward tilt and early openings, while DiffPoseTalk bows and turns the face during
  listening.}
  \label{fig:visual-case-009}
\end{figure}

\begin{figure}[p]
  \centering
  \includegraphics[width=\linewidth]{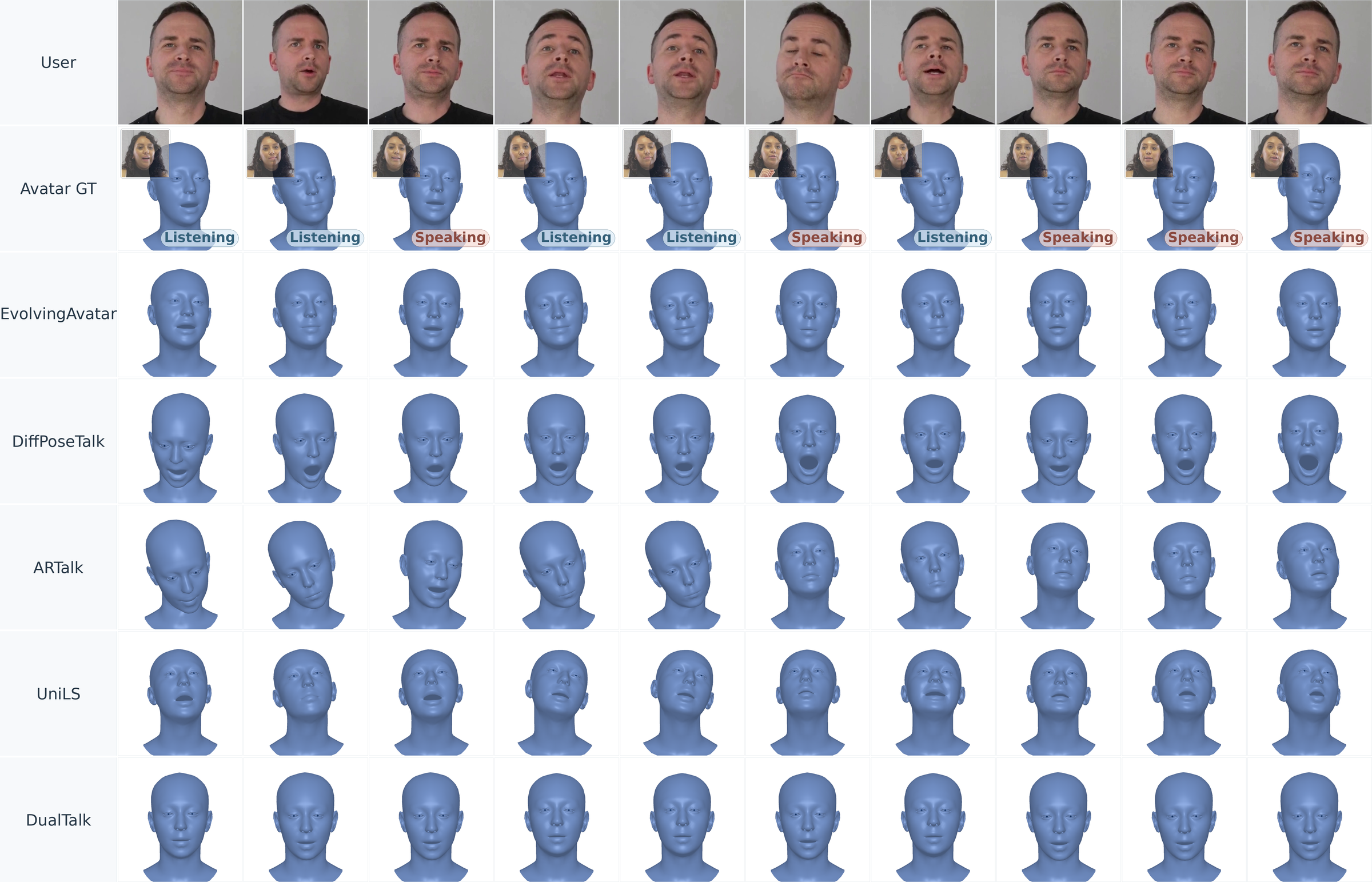}
  \caption{\textbf{Changing articulation within each activity.} EvolvingAvatar follows the
  reference's alternation between smaller and larger mouth openings while retaining a moderate
  neck pose across the sampled instants. UniLS raises the chin too far and opens the mouth in
  several near-closed reference frames, while DiffPoseTalk over-opens the mouth across most of
  the displayed columns.}
  \label{fig:visual-case-034}
\end{figure}

\begin{figure}[p]
  \centering
  \includegraphics[width=\linewidth]{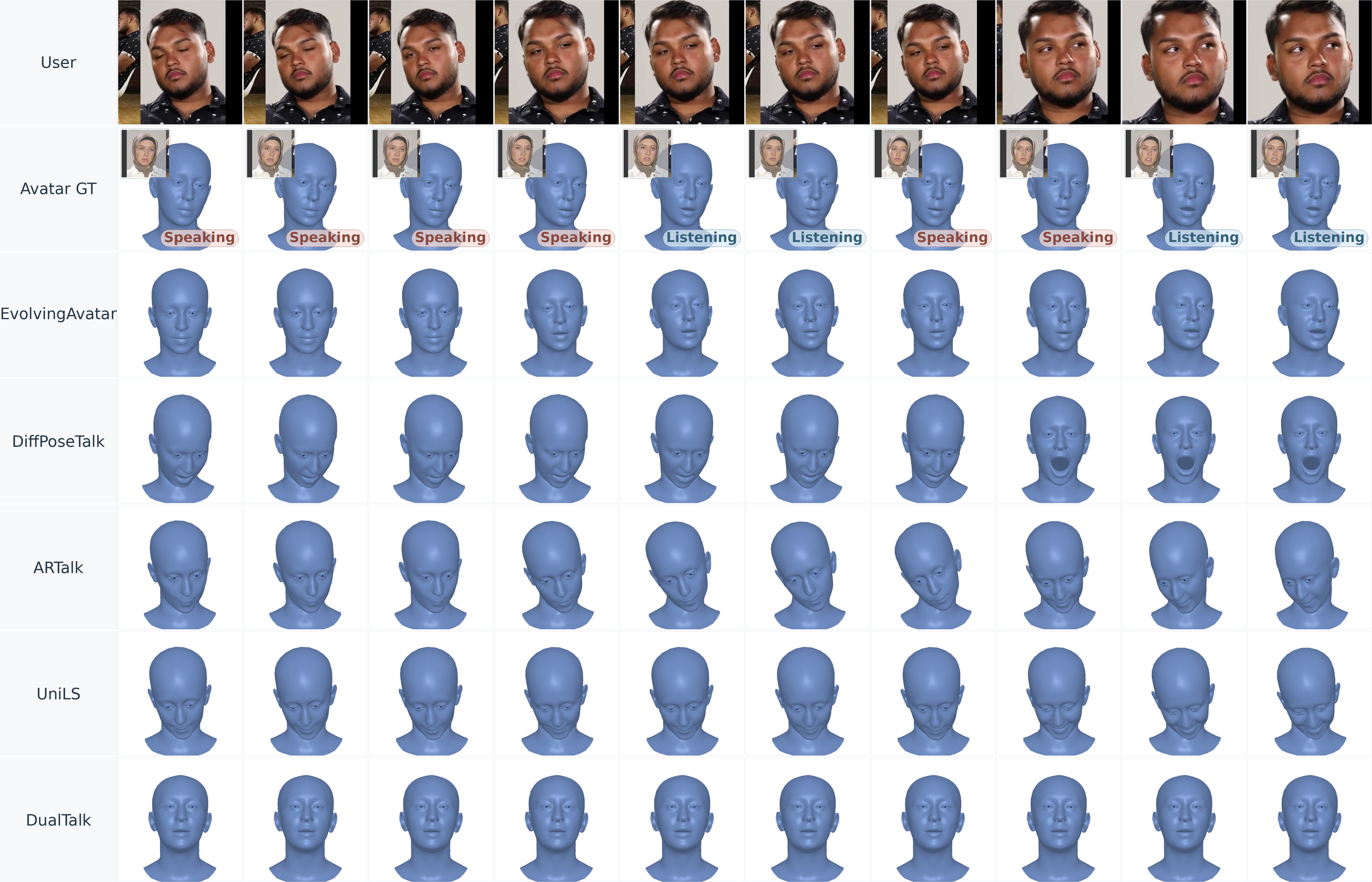}
  \caption{\textbf{Subtle articulation with an upright neck pose.} EvolvingAvatar preserves the
  near-upright neck pose and the mild increase in mouth opening in the later columns. UniLS and
  ARTalk lower the face, while DiffPoseTalk replaces the small final openings with larger ones.}
  \label{fig:visual-case-012}
\end{figure}

\begin{figure}[p]
  \centering
  \includegraphics[width=\linewidth]{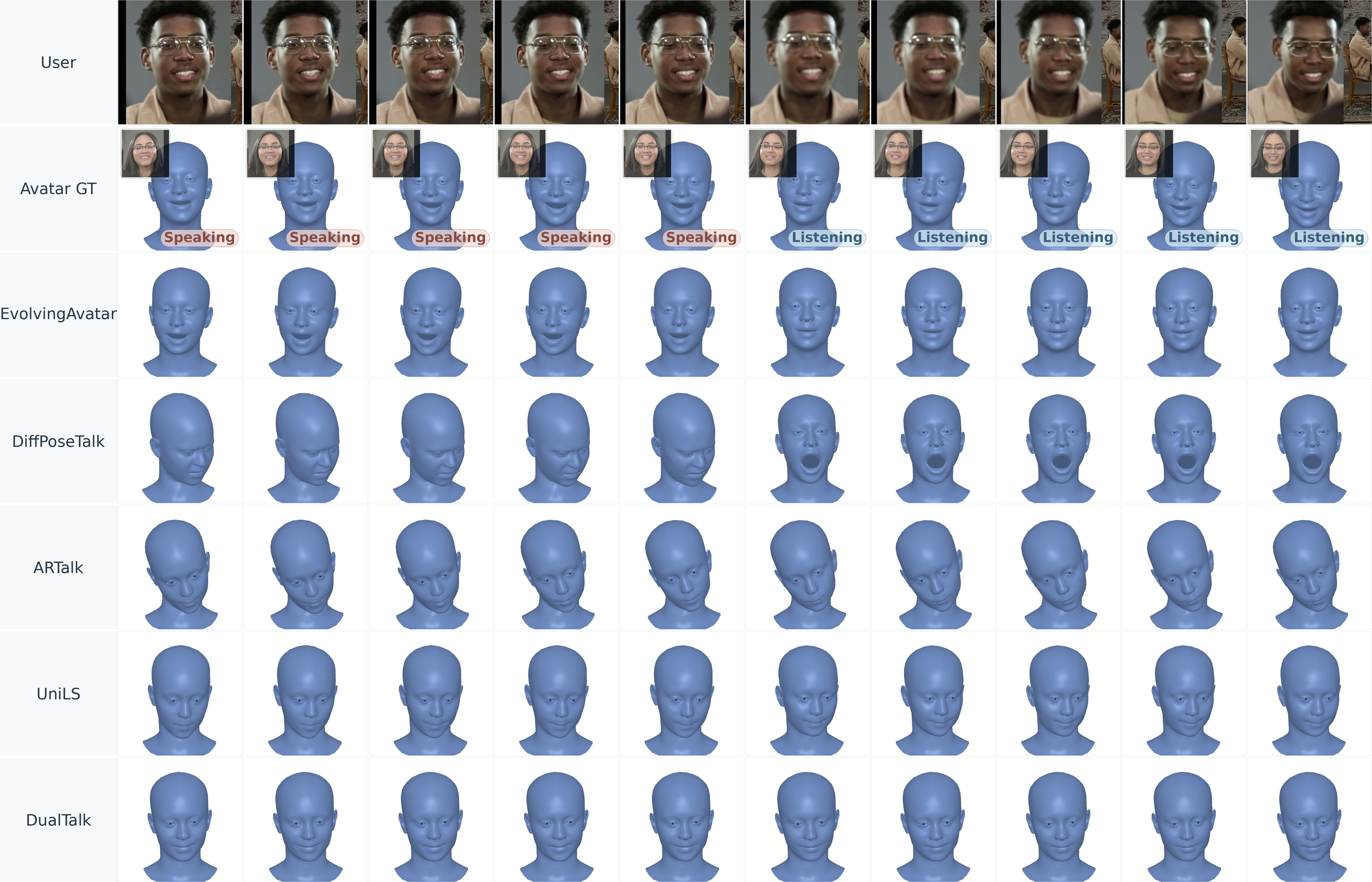}
  \caption{\textbf{A smile across speaking and listening.} EvolvingAvatar combines a
  near-frontal pose with larger speaking openings and a softer listening smile, retaining
  visible expression across both activities. UniLS and DualTalk understate the reference
  speaking expression in the first five columns, while DiffPoseTalk produces excessive mouth
  opening in the subsequent listening frames.}
  \label{fig:visual-case-014}
\end{figure}

\begin{figure}[p]
  \centering
  \includegraphics[width=\linewidth]{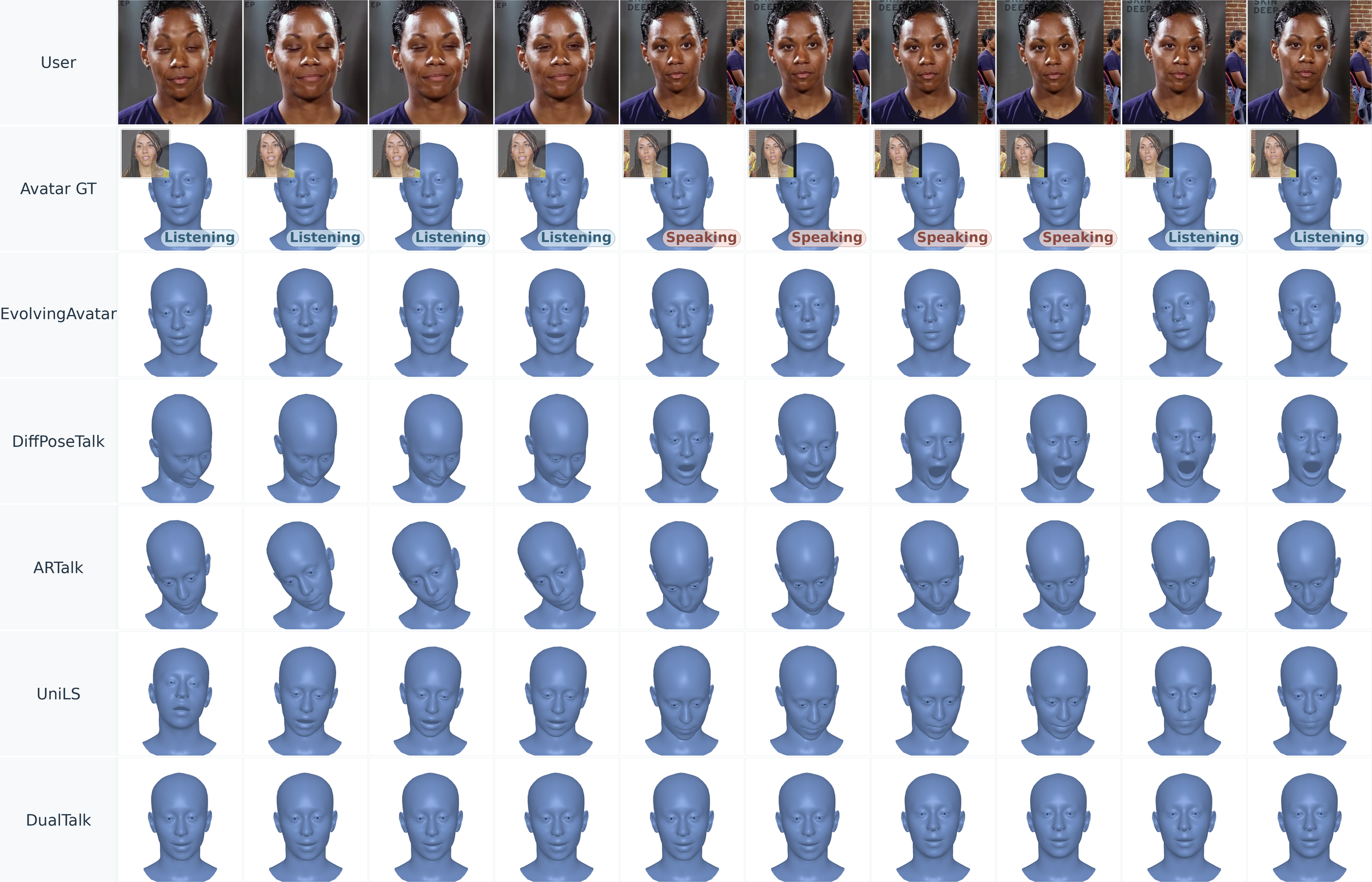}
  \caption{\textbf{Expressive listening with an upright pose.} EvolvingAvatar captures the
  listening smile in columns 2--4 and stays near the reference orientation during speaking.
  UniLS and ARTalk lower the face during speaking, while DualTalk shows less of the reference
  variation in mouth shape.}
  \label{fig:visual-case-016}
\end{figure}

\begin{figure}[p]
  \centering
  \includegraphics[width=\linewidth]{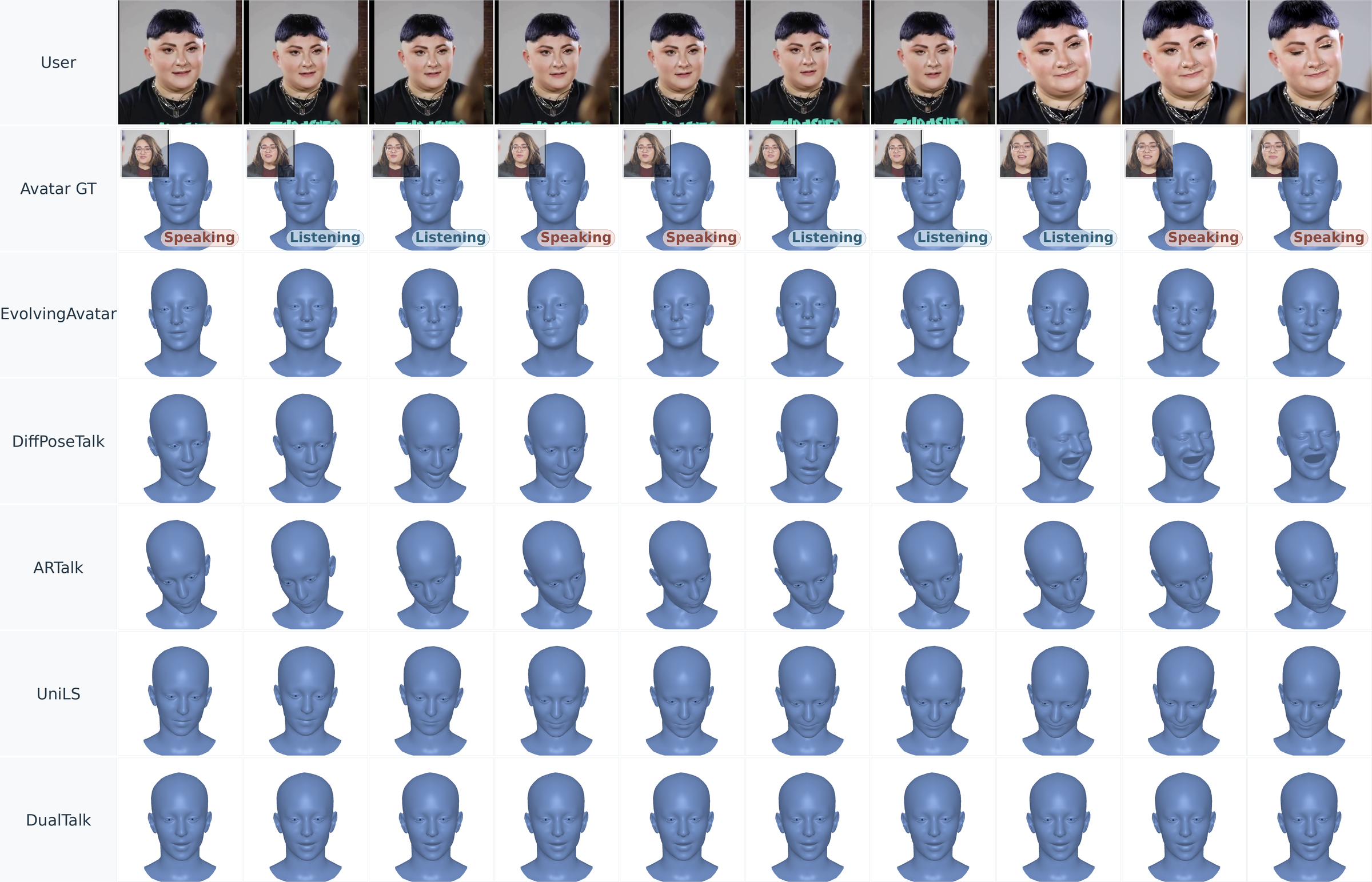}
  \caption{\textbf{A frontal smile with varying mouth opening.} EvolvingAvatar preserves the
  reference's upright pose and smile, including the larger openings in columns 8--9. UniLS and
  ARTalk lower the face, while DiffPoseTalk adds a pronounced sideways turn and wider grin in
  the final columns.}
  \label{fig:visual-case-019}
\end{figure}

\begin{figure}[p]
  \centering
  \includegraphics[width=\linewidth]{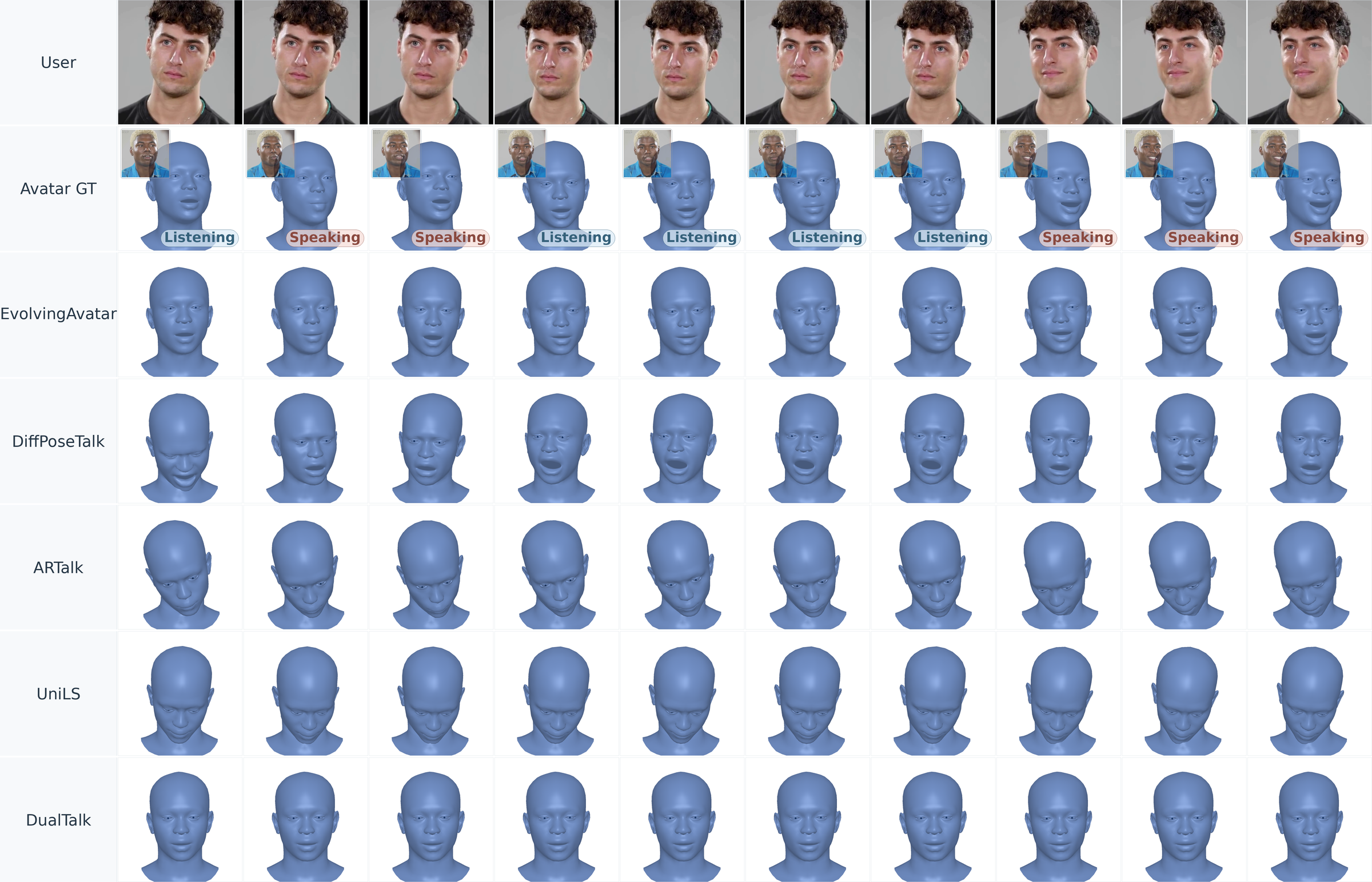}
  \caption{\textbf{Expression changes with restrained neck motion.} EvolvingAvatar follows the
  reference's smaller openings in columns 6--7 and broader speaking smile in columns 8--10.
  UniLS and ARTalk bow the head, while DualTalk shows less change in mouth shape than the
  reference.}
  \label{fig:visual-case-031}
\end{figure}

\begin{figure}[p]
  \centering
  \includegraphics[width=\linewidth]{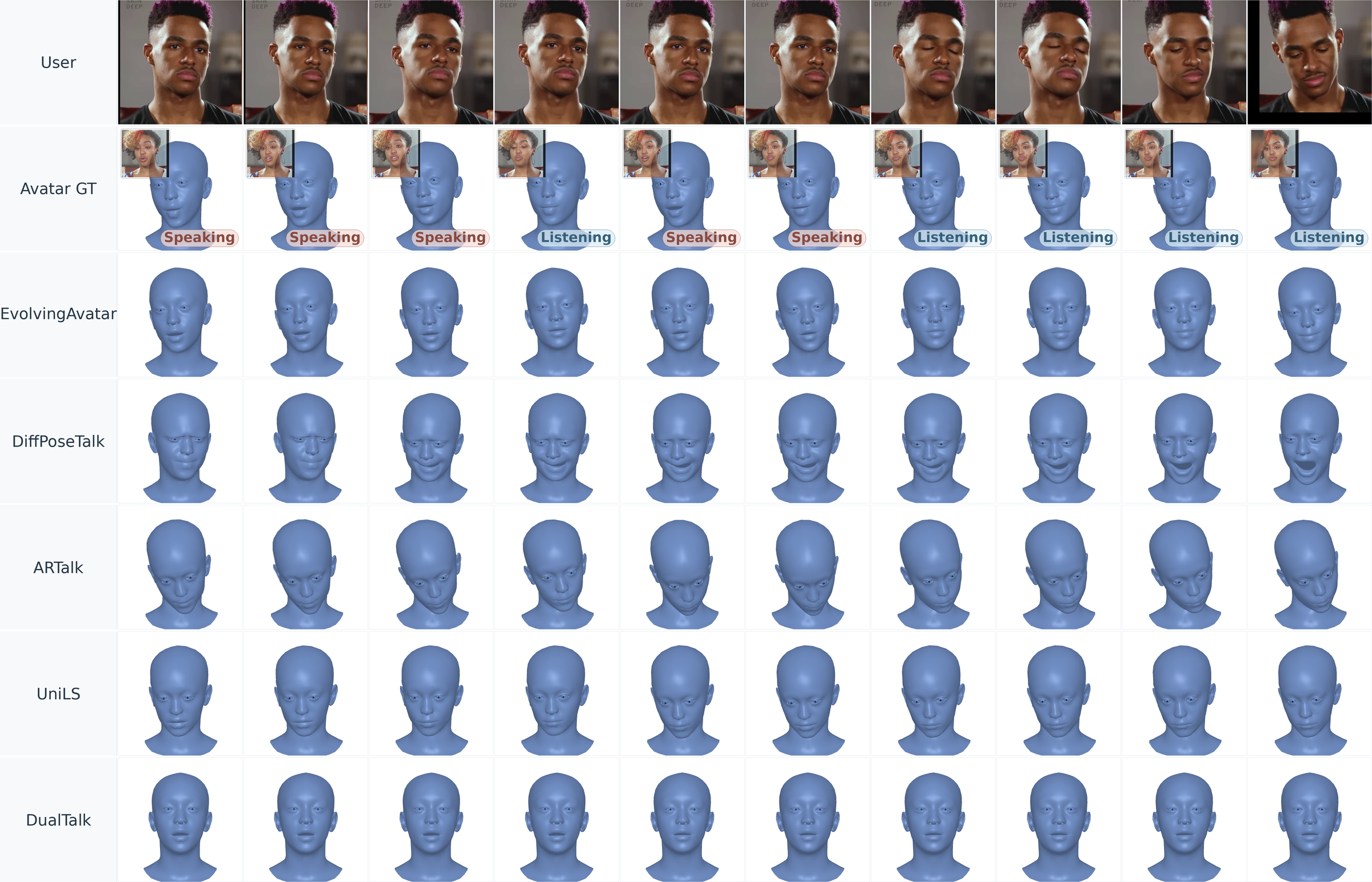}
  \caption{\textbf{Reducing articulation during listening.} EvolvingAvatar remains nearly
  upright while following the smaller listening openings in columns 7--10. UniLS and ARTalk
  lower the face, while DualTalk retains a parted mouth and DiffPoseTalk produces a wider final
  opening than Avatar GT.}
  \label{fig:visual-case-040}
\end{figure}

Figures~\ref{fig:visual-case-046}--\ref{fig:visual-case-040} provide 15 additional visual
comparisons. From top to bottom, rows show the User video, Avatar GT, EvolvingAvatar,
DiffPoseTalk, ARTalk, UniLS, and DualTalk. Columns show the same time points across
methods, ordered from left to right. Insets in the Avatar GT row show the target
person's face, and colored labels indicate whether the Avatar is speaking or listening.

Across these cases, EvolvingAvatar more closely follows reference expressions, neck
poses, and mouth openings during speaking and listening. It preserves smiles during
listening, while several baselines show excessive neck deviations, exaggerated mouth
openings, or weaker expressions.

EvolvingAvatar follows changes in articulation across speaking and listening. In
Figure~\ref{fig:visual-case-003}, the smaller listening openings in columns 1--5 give
way to larger speaking openings in columns 7--10. Figures~\ref{fig:visual-case-009}
and~\ref{fig:visual-case-014} show the reverse change, from speaking articulation to a
listening smile. Our outputs follow both patterns and retain smiles even with small
mouth openings. Several baselines show less expression change or larger mouth openings
than the reference at matching times.

Expressive listening is also visible in these examples. In
Figures~\ref{fig:visual-case-001}, \ref{fig:visual-case-008},
and~\ref{fig:visual-case-016}, the reference Avatar smiles while listening, sometimes
with parted lips. EvolvingAvatar preserves these expressions, whereas some baseline
outputs understate the smile or mouth opening. Together with the smaller listening
openings in Figure~\ref{fig:visual-case-040}, these cases illustrate why the desired
response depends on the ongoing exchange rather than a fixed facial expression
associated with the listening state.

Neck orientation further distinguishes the displayed outputs. In
Figure~\ref{fig:visual-case-033}, EvolvingAvatar follows the reference's slight downward
listening pose and sideways speaking lean. Figures~\ref{fig:visual-case-006},
\ref{fig:visual-case-019}, and~\ref{fig:visual-case-040} instead show a nearly upright
reference orientation with changing mouth shapes. Several baselines introduce stronger
downward or sideways rotations in these cases. EvolvingAvatar better retains the
relation between neck orientation and facial expression across both speaking and
listening.

\clearpage

\section{All-frame quantitative results}
\label{app:all-frame-results}

\paragraph{We outperform baselines on all-frame motion statistics}
\methodname{} outperforms all four baselines in expression, neck, and jaw FD
across splits (Table~\ref{tab:parameter-space-results-all}). Metrics are
recomputed over all valid speaking and listening frames, including overlapping
speech and silence. This extends activity-specific comparisons to statistics over
the conversational timeline. P-FD also favors our method except for OOD-Hard
expression, where DualTalk is slightly lower (13.49 versus 13.52).

\paragraph{Our mesh motion advantage is strongest on OOD-Hard}
\methodname{} outperforms all baselines in all-frame mesh FDD on OOD-Hard
(Table~\ref{tab:mesh-space-results-all}), indicating closer upper-face motion
amplitudes. UniLS leads FDD on ID/OOD, while DualTalk has lower LVE and MHD across
splits. As in the main results, motion statistics and agreement with one reference
assess different aspects of quality. Parameter metrics also assess neck motion,
which mesh evaluation excludes. Both spaces offer complementary evidence, with
human judgments assessing perceived conversational quality.

\begin{table*}[t]
\centering

\caption{\textbf{Parameter-space comparison.} Results over all valid frames. Green bold
and blue underline indicate the best and second-best values within each split.
Lower is better except for SID.}
\label{tab:parameter-space-results-all}

\setlength{\tabcolsep}{2.2pt}
\renewcommand{\arraystretch}{1.08}
\footnotesize

\resizebox{\linewidth}{!}{%
\begin{tabular}{
Y
!{\color{TableSeparator}\vrule width 0.45pt}
*{17}{Z}
}

\toprule


\multicolumn{1}{
  c!{\color{TableSeparator}\vrule width 0.45pt}
}{%
  \multirow[c]{2}{*}[-2.5mm]{\textbf{Method}}%
}
&
\multicolumn{3}{c}{\textbf{FD} $\downarrow$}
&
\multicolumn{3}{c}{\textbf{P-FD} $\downarrow$}
&
\multicolumn{3}{c}{\textbf{MSE} $\downarrow$}
&
\multicolumn{3}{c}{\textbf{rPCC} $\downarrow$}
&
\multicolumn{3}{c}{\textbf{SID} $\uparrow$}
&
\multicolumn{1}{c}{%
  \multirow[c]{2}{*}[-2.5mm]{%
    \textbf{PDD} $\downarrow$%
  }%
}
&
\multicolumn{1}{c}{%
  \multirow[c]{2}{*}[-2.5mm]{%
    \textbf{JDD} $\downarrow$%
  }%
}
\\


\cmidrule(l{1.5pt}r{1.5pt}){2-4}
\cmidrule(l{1.5pt}r{1.5pt}){5-7}
\cmidrule(l{1.5pt}r{1.5pt}){8-10}
\cmidrule(l{1.5pt}r{1.5pt}){11-13}
\cmidrule(l{1.5pt}r{1.5pt}){14-16}


&
\PlainMetricHead{EXP} 
&
\makecell[c]{
  \textbf{JAW}\\
  {\scriptsize$(\times 10^{3})$}
}
&
\makecell[c]{
  \textbf{POSE}\\
  {\scriptsize$(\times 10^{2})$}
}
&
\PlainMetricHead{EXP} 
&
\makecell[c]{
  \textbf{JAW}\\
  {\scriptsize$(\times 10^{3})$}
}
&
\makecell[c]{
  \textbf{POSE}\\
  {\scriptsize$(\times 10^{2})$}
}
&
\makecell[c]{
  \textbf{EXP}\\
  {\scriptsize$(\times 10^{1})$}
}
&
\makecell[c]{
  \textbf{JAW}\\
  {\scriptsize$(\times 10^{3})$}
}
&
\makecell[c]{
  \textbf{POSE}\\
  {\scriptsize$(\times 10^{2})$}
}
&
\makecell[c]{
  \textbf{EXP}\\
  {\scriptsize$(\times 10^{2})$}
}
&
\makecell[c]{
  \textbf{JAW}\\
  {\scriptsize$(\times 10^{1})$}
}
&
\makecell[c]{
  \textbf{POSE}\\
  {\scriptsize$(\times 10^{1})$}
}
&
\PlainMetricHead{EXP}  
&
\PlainMetricHead{JAW}
&
\PlainMetricHead{POSE}
&
&
\\

\midrule


\DatasetBand{In-Distribution Test Set (ID)}

\DataRow
  {\MethodRef{DiffPoseTalk}{sun2024diffposetalk}}
  {45.68 & 25.60 & 23.10}
  {47.40 & 25.90 & 23.99}
  {5.83 & 11.35 & 11.38}
  {\second{18.94} & 1.90 & \second{2.90}}
  {1.85 & 1.45 & \best{1.48}}
  {9.87}
  {3.04}

\DataRow
  {\MethodRef{ARTalk}{chu2025artalk}}
  {22.99 & 3.94 & 21.71}
  {23.91 & 4.03 & 22.32}
  {2.93 & 1.86 & 9.60}
  {21.72 & \second{1.57} & 3.18}
  {1.69 & 1.55 & 1.17}
  {\best{5.93}}
  {1.26}

\DataRow
  {\MethodRef{UniLS}{chu2026unils}}
  {\second{17.19} & \second{3.05} & 12.95}
  {18.20 & \second{3.18} & 13.37}
  {2.45 & \best{1.79} & 5.60}
  {\best{15.86} & \best{1.48} & 3.16}
  {\best{2.31} & \second{1.90} & 1.04}
  {7.52}
  {\best{1.03}}

\DataRow
  {\MethodRef{DualTalk}{peng2025dualtalk}}
  {17.22 & 4.63 & \second{11.52}}
  {\second{17.35} & 4.68 & \second{11.60}}
  {\best{1.79} & \second{1.81} & \best{4.05}}
  {23.17 & 1.81 & 3.66}
  {0.32 & 0.57 & 0.28}
  {11.34}
  {2.09}

\DataRow
  {\MethodOurs}
  {\best{15.77} & \best{2.64} & \best{8.89}}
  {\best{16.79} & \best{2.79} & \best{9.32}}
  {\second{2.28} & 1.87 & \second{4.37}}
  {20.74 & 1.59 & \best{2.82}}
  {\second{2.14} & \best{2.06} & \second{1.36}}
  {\second{7.41}}
  {\second{1.13}}


\DatasetBand{Out-of-Distribution Test Set (OOD)}

\DataRow
  {\MethodRef{DiffPoseTalk}{sun2024diffposetalk}}
  {45.54 & 23.77 & 30.25}
  {47.04 & 23.97 & 31.09}
  {5.68 & 10.15 & 13.70}
  {21.32 & 1.90 & \second{3.19}}
  {1.75 & 1.47 & \second{1.34}}
  {10.27}
  {3.29}

\DataRow
  {\MethodRef{ARTalk}{chu2025artalk}}
  {21.82 & 2.54 & 26.65}
  {22.64 & 2.61 & 27.25}
  {2.72 & \best{1.24} & 11.09}
  {23.04 & \second{1.61} & 3.42}
  {1.66 & 1.66 & 1.09}
  {\best{6.02}}
  {0.90}

\DataRow
  {\MethodRef{UniLS}{chu2026unils}}
  {16.13 & \second{2.39} & 14.94}
  {16.99 & \second{2.47} & 15.33}
  {2.23 & 1.33 & 6.15}
  {\best{15.32} & \best{1.48} & 3.48}
  {\best{2.24} & \second{1.93} & 0.94}
  {7.35}
  {\best{0.78}}

\DataRow
  {\MethodRef{DualTalk}{peng2025dualtalk}}
  {\second{15.78} & 3.49 & \second{13.08}}
  {\second{15.85} & 3.51 & \second{13.14}}
  {\best{1.61} & \second{1.31} & \best{4.50}}
  {20.85 & 1.89 & 3.76}
  {0.22 & 0.47 & 0.19}
  {11.25}
  {1.73}

\DataRow
  {\MethodOurs}
  {\best{13.61} & \best{2.06} & \best{9.32}}
  {\best{14.59} & \best{2.16} & \best{9.80}}
  {\second{2.01} & 1.45 & \second{4.58}}
  {\second{16.84} & 1.67 & \best{2.75}}
  {\second{2.21} & \best{2.06} & \best{1.41}}
  {\second{7.02}}
  {\second{0.83}}


\DatasetBand{Hard Out-of-Distribution Test Set (OOD-Hard)}

\DataRow
  {\MethodRef{DiffPoseTalk}{sun2024diffposetalk}}
  {50.77 & 24.55 & 26.56}
  {51.91 & 24.71 & 26.95}
  {5.82 & 9.64 & 10.28}
  {29.64 & 2.07 & \second{5.09}}
  {\second{1.66} & 1.22 & \second{1.16}}
  {10.74}
  {3.26}

\DataRow
  {\MethodRef{ARTalk}{chu2025artalk}}
  {22.41 & 3.97 & 36.19}
  {23.09 & 4.03 & 36.49}
  {2.61 & 1.66 & 13.01}
  {27.77 & \second{2.06} & 5.67}
  {1.30 & 1.20 & 0.58}
  {5.80}
  {\second{0.94}}

\DataRow
  {\MethodRef{UniLS}{chu2026unils}}
  {21.75 & 3.56 & 22.42}
  {22.58 & 3.64 & 22.67}
  {2.66 & 1.75 & 8.11}
  {21.91 & 2.69 & 5.81}
  {1.62 & \second{1.43} & 0.61}
  {\second{4.65}}
  {0.99}

\DataRow
  {\MethodRef{DualTalk}{peng2025dualtalk}}
  {\second{13.43} & \second{3.14} & \second{8.65}}
  {\best{13.49} & \second{3.17} & \second{8.69}}
  {\best{1.36} & \best{1.11} & \best{2.94}}
  {\second{18.62} & 2.34 & 5.98}
  {0.12 & 0.24 & 0.13}
  {5.23}
  {1.34}

\DataRow
  {\MethodOurs}
  {\best{12.77} & \best{2.25} & \best{7.98}}
  {\second{13.52} & \best{2.32} & \best{8.20}}
  {\second{1.75} & \second{1.26} & \second{3.36}}
  {\best{16.45} & \best{1.73} & \best{4.76}}
  {\best{2.24} & \best{1.83} & \best{1.23}}
  {\best{3.85}}
  {\best{0.82}}

\bottomrule

\end{tabular}%
}

\end{table*}

\begin{table*}[t]
\centering

\caption{\textbf{Mesh-space comparison.} All-frame results. LVE and MHD are measured in
millimeters. Green bold and blue underlined values mark the best and second
best in each split. Lower is better.}
\label{tab:mesh-space-results-all}

\setlength{\tabcolsep}{2.8pt}
\renewcommand{\arraystretch}{1.08}
\footnotesize

\resizebox{0.75\linewidth}{!}{%
\begin{tabular}{
Y
!{\color{TableSeparator}\vrule width 0.45pt}
*{9}{c}
}

\toprule

\multicolumn{1}{
  c!{\color{TableSeparator}\vrule width 0.45pt}
}{%
  \multirow[c]{2}{*}[-0.5ex]{\textbf{Method}}%
}
&
\multicolumn{3}{c}{\textbf{ID}}
&
\multicolumn{3}{c}{\textbf{OOD}}
&
\multicolumn{3}{c}{\textbf{OOD-Hard}}
\\

\cmidrule(l{1.5pt}r{1.5pt}){2-4}
\cmidrule(l{1.5pt}r{1.5pt}){5-7}
\cmidrule(l{1.5pt}r{1.5pt}){8-10}

&
\textbf{FDD} $\downarrow$
&
\textbf{LVE} $\downarrow$
&
\textbf{MHD} $\downarrow$
&
\textbf{FDD} $\downarrow$
&
\textbf{LVE} $\downarrow$
&
\textbf{MHD} $\downarrow$
&
\textbf{FDD} $\downarrow$
&
\textbf{LVE} $\downarrow$
&
\textbf{MHD} $\downarrow$
\\

\midrule

\MethodRef{DiffPoseTalk}{sun2024diffposetalk}
& 40.03 & 19.77 & 4.28
& 40.37 & 18.52 & 4.09
& 44.83 & 18.31 & 4.02
\\

\MethodRef{ARTalk}{chu2025artalk}
& \second{35.22} & 11.17 & 2.62
& \second{30.51} & 9.35 & 2.22
& \second{19.51} & 10.09 & 2.34
\\

\MethodRef{UniLS}{chu2026unils}
& \best{30.39} & \second{9.33} & \second{2.10}
& \best{27.64} & \second{8.33} & \second{1.87}
& 21.65 & 10.65 & 2.34
\\

\MethodRef{DualTalk}{peng2025dualtalk}
& 66.70 & \best{8.87} & \best{1.96}
& 61.46 & \best{7.71} & \best{1.71}
& 40.10 & \best{7.60} & \best{1.66}
\\

\MethodOurs
& 37.91 & 9.83 & 2.18
& 31.53 & 8.74 & 1.92
& \best{17.87} & \second{8.71} & \second{1.89}
\\

\bottomrule

\end{tabular}%
}

\end{table*}

\clearpage

\section{Context signal comparison}
\label{app:context-signals}

\begin{table*}[t]
\centering

\caption{\textbf{Context signals in parameter space.} Speaking/listening results. Green
bold and blue underlining denote the best and second best per split and state.
Lower is better except SID.}
\label{tab:context-parameter}

\setlength{\tabcolsep}{2.2pt}
\renewcommand{\arraystretch}{1.08}
\footnotesize

\resizebox{\linewidth}{!}{%
\begin{tabular}{
Y
!{\color{TableSeparator}\vrule width 0.45pt}
*{17}{Z}
}

\toprule

\multicolumn{1}{
  c!{\color{TableSeparator}\vrule width 0.45pt}
}{%
  \multirow[c]{2}{*}[-2.5mm]{\textbf{Method}}%
}
&
\multicolumn{3}{c}{\textbf{FD} $\downarrow$}
&
\multicolumn{3}{c}{\textbf{P-FD} $\downarrow$}
&
\multicolumn{3}{c}{\textbf{MSE} $\downarrow$}
&
\multicolumn{3}{c}{\textbf{rPCC} $\downarrow$}
&
\multicolumn{3}{c}{\textbf{SID} $\uparrow$}
&
\multicolumn{1}{c}{%
  \multirow[c]{2}{*}[-2.5mm]{%
    \textbf{PDD} $\downarrow$%
  }%
}
&
\multicolumn{1}{c}{%
  \multirow[c]{2}{*}[-2.5mm]{%
    \textbf{JDD} $\downarrow$%
  }%
}
\\

\cmidrule(l{1.5pt}r{1.5pt}){2-4}
\cmidrule(l{1.5pt}r{1.5pt}){5-7}
\cmidrule(l{1.5pt}r{1.5pt}){8-10}
\cmidrule(l{1.5pt}r{1.5pt}){11-13}
\cmidrule(l{1.5pt}r{1.5pt}){14-16}

&
\PlainMetricHead{EXP} 
&
\makecell[c]{
  \textbf{JAW}\\
  {\scriptsize$(\times 10^{3})$}
}
&
\makecell[c]{
  \textbf{NECK}\\
  {\scriptsize$(\times 10^{2})$}
}
&
\PlainMetricHead{EXP} 
&
\makecell[c]{
  \textbf{JAW}\\
  {\scriptsize$(\times 10^{3})$}
}
&
\makecell[c]{
  \textbf{NECK}\\
  {\scriptsize$(\times 10^{2})$}
}
&
\makecell[c]{
  \textbf{EXP}\\
  {\scriptsize$(\times 10^{1})$}
}
&
\makecell[c]{
  \textbf{JAW}\\
  {\scriptsize$(\times 10^{3})$}
}
&
\makecell[c]{
  \textbf{NECK}\\
  {\scriptsize$(\times 10^{2})$}
}
&
\makecell[c]{
  \textbf{EXP}\\
  {\scriptsize$(\times 10^{2})$}
}
&
\makecell[c]{
  \textbf{JAW}\\
  {\scriptsize$(\times 10^{1})$}
}
&
\makecell[c]{
  \textbf{NECK}\\
  {\scriptsize$(\times 10^{1})$}
}
&
\PlainMetricHead{EXP}  
&
\PlainMetricHead{JAW}
&
\PlainMetricHead{NECK}
&
&
\\

\midrule

\DatasetBand{In-Distribution Test Set (ID)}

\StateRow{Speaking}

\DataRow
  {\MethodRef{UniLS}{chu2026unils}}
  {15.99 & 2.36 & 12.31}
  {17.05 & 2.45 & 12.76}
  {2.32 & \best{1.30} & 5.36}
  {\best{14.73} & \best{1.07} & 3.37}
  {\best{2.84} & \best{2.20} & 1.13}
  {7.36}
  {\best{0.84}}

\DataRow
  {\MethodRef{DualTalk}{peng2025dualtalk}}
  {16.80 & 4.17 & 12.05}
  {16.97 & 4.23 & 12.16}
  {\best{1.74} & \second{1.59} & \best{4.21}}
  {20.69 & 1.31 & 4.13}
  {0.31 & 0.49 & 0.25}
  {11.37}
  {2.08}

\DataRow
  {\textbf{Ours (audio)}}
  {\best{13.71} & \second{2.21} & \second{9.34}}
  {\best{14.90} & \second{2.35} & \second{9.83}}
  {\second{2.15} & 1.69 & 4.70}
  {\second{15.38} & \second{1.22} & 3.29}
  {\second{2.47} & 2.13 & \second{1.42}}
  {\best{6.86}}
  {0.95}

\DataRow
  {\textbf{Ours (FLAME)}}
  {\second{14.24} & \best{2.13} & \best{9.07}}
  {\second{15.41} & \best{2.28} & \best{9.55}}
  {2.18 & 1.68 & \second{4.57}}
  {18.78 & 1.24 & \best{3.07}}
  {2.43 & \second{2.16} & \best{1.44}}
  {\second{6.89}}
  {0.94}

\DataRow
  {\textbf{Ours (RGB)}}
  {15.49 & 2.40 & 9.39}
  {16.64 & 2.56 & 9.87}
  {2.27 & 1.78 & 4.61}
  {19.01 & 1.25 & \second{3.17}}
  {2.30 & 2.13 & 1.35}
  {7.09}
  {\second{0.92}}

\StateRow{Listening}

\DataRow
  {\MethodRef{UniLS}{chu2026unils}}
  {18.61 & 3.60 & 13.07}
  {19.74 & 3.73 & 13.53}
  {2.50 & 2.00 & 5.45}
  {\best{17.34} & \best{1.65} & 3.25}
  {1.99 & 1.65 & 0.95}
  {7.08}
  {\best{1.10}}

\DataRow
  {\MethodRef{DualTalk}{peng2025dualtalk}}
  {17.64 & 5.07 & 11.06}
  {17.78 & 5.11 & 11.14}
  {\best{1.82} & 1.95 & \best{3.88}}
  {24.94 & 2.01 & 3.65}
  {0.33 & 0.54 & 0.28}
  {10.31}
  {1.99}

\DataRow
  {\textbf{Ours (audio)}}
  {\best{15.09} & \second{3.08} & \second{8.79}}
  {\best{16.30} & \second{3.23} & \second{9.31}}
  {\second{2.17} & \second{1.88} & 4.25}
  {\second{19.79} & \second{1.71} & 3.05}
  {\best{2.15} & 1.86 & \best{1.39}}
  {\best{6.73}}
  {1.26}

\DataRow
  {\textbf{Ours (FLAME)}}
  {\second{15.57} & \best{2.94} & \best{8.46}}
  {\second{16.73} & \best{3.10} & \best{8.95}}
  {2.19 & \best{1.84} & \second{4.08}}
  {23.39 & 1.79 & \best{2.81}}
  {\second{2.10} & \best{1.89} & \best{1.39}}
  {\second{6.77}}
  {\second{1.23}}

\DataRow
  {\textbf{Ours (RGB)}}
  {17.01 & 3.17 & 8.99}
  {18.13 & 3.33 & 9.46}
  {2.30 & 1.95 & 4.18}
  {22.45 & 1.78 & \second{2.88}}
  {1.95 & \second{1.87} & \second{1.29}}
  {6.97}
  {\second{1.23}}

\DatasetBand{Out-of-Distribution Test Set (OOD)}

\StateRow{Speaking}

\DataRow
  {\MethodRef{UniLS}{chu2026unils}}
  {15.41 & 2.14 & 13.58}
  {16.37 & 2.21 & 14.00}
  {2.15 & \best{1.06} & 5.68}
  {\second{14.70} & \best{1.10} & 3.69}
  {\best{2.69} & \best{2.15} & 1.02}
  {7.00}
  {\best{0.65}}

\DataRow
  {\MethodRef{DualTalk}{peng2025dualtalk}}
  {15.80 & 3.23 & 12.86}
  {15.91 & 3.26 & 12.94}
  {\best{1.61} & \second{1.20} & \best{4.41}}
  {19.16 & 1.42 & 4.21}
  {0.22 & 0.43 & 0.17}
  {11.10}
  {1.64}

\DataRow
  {\textbf{Ours (audio)}}
  {\best{12.97} & \second{2.02} & 9.94}
  {\best{14.13} & \second{2.12} & 10.43}
  {\second{1.99} & 1.39 & 4.84}
  {\best{14.29} & 1.26 & 3.39}
  {\second{2.52} & \second{2.13} & \best{1.38}}
  {\second{6.66}}
  {\second{0.69}}

\DataRow
  {\textbf{Ours (FLAME)}}
  {\second{13.46} & \best{1.96} & \second{9.88}}
  {\second{14.61} & \best{2.06} & \second{10.37}}
  {2.02 & 1.36 & 4.77}
  {16.39 & \second{1.24} & \best{3.15}}
  {2.45 & \second{2.13} & \second{1.35}}
  {\best{6.65}}
  {0.72}

\DataRow
  {\textbf{Ours (RGB)}}
  {14.10 & 2.24 & \best{9.46}}
  {15.21 & 2.35 & \best{9.94}}
  {2.05 & 1.45 & \second{4.59}}
  {15.32 & \second{1.24} & \second{3.17}}
  {2.33 & 2.06 & \second{1.35}}
  {6.81}
  {0.73}

\StateRow{Listening}

\DataRow
  {\MethodRef{UniLS}{chu2026unils}}
  {17.53 & 2.97 & 15.60}
  {18.51 & 3.07 & 16.05}
  {2.26 & 1.54 & 6.17}
  {\best{16.63} & \best{1.72} & 3.69}
  {1.89 & 1.61 & 0.81}
  {6.93}
  {\best{0.87}}

\DataRow
  {\MethodRef{DualTalk}{peng2025dualtalk}}
  {16.04 & 3.86 & 12.97}
  {16.13 & 3.88 & 13.03}
  {\best{1.64} & \second{1.43} & \best{4.45}}
  {22.69 & 2.13 & 3.82}
  {0.24 & 0.44 & 0.19}
  {10.12}
  {1.62}

\DataRow
  {\textbf{Ours (audio)}}
  {\best{13.65} & \second{2.18} & 10.53}
  {\best{14.81} & \second{2.30} & 11.12}
  {\second{1.95} & \second{1.43} & 4.81}
  {\second{17.48} & \second{1.91} & \second{3.35}}
  {\best{2.26} & \best{1.89} & \best{1.31}}
  {\second{6.53}}
  {\best{0.87}}

\DataRow
  {\textbf{Ours (FLAME)}}
  {\second{14.01} & \best{2.17} & \second{10.33}}
  {\second{15.14} & \best{2.29} & \second{10.89}}
  {1.96 & \best{1.41} & 4.68}
  {20.44 & 1.95 & \best{3.04}}
  {\second{2.18} & \second{1.86} & \second{1.30}}
  {\best{6.50}}
  {\second{0.89}}

\DataRow
  {\textbf{Ours (RGB)}}
  {14.61 & 2.37 & \best{9.90}}
  {15.70 & 2.49 & \best{10.45}}
  {1.99 & 1.47 & \second{4.52}}
  {18.92 & 1.96 & \best{3.04}}
  {2.05 & 1.83 & 1.29}
  {6.61}
  {\second{0.89}}

\DatasetBand{Hard Out-of-Distribution Test Set (OOD-Hard)}

\StateRow{Speaking}

\DataRow
  {\MethodRef{UniLS}{chu2026unils}}
  {20.87 & 3.84 & 26.15}
  {21.64 & 3.91 & 26.38}
  {2.56 & 1.88 & 9.33}
  {19.01 & 2.28 & 6.04}
  {1.83 & 1.58 & 0.55}
  {4.01}
  {0.81}

\DataRow
  {\MethodRef{DualTalk}{peng2025dualtalk}}
  {13.83 & 3.17 & 8.77}
  {13.89 & 3.18 & 8.81}
  {\best{1.39} & \best{1.10} & \best{2.97}}
  {16.73 & 1.49 & 6.46}
  {0.08 & 0.21 & 0.13}
  {5.70}
  {1.48}

\DataRow
  {\textbf{Ours (audio)}}
  {13.87 & \second{2.23} & 9.08}
  {14.70 & \second{2.31} & 9.35}
  {1.89 & 1.37 & 3.89}
  {15.91 & \best{1.30} & 5.48}
  {\second{2.29} & \best{2.04} & 1.15}
  {3.74}
  {\second{0.66}}

\DataRow
  {\textbf{Ours (FLAME)}}
  {\second{12.87} & \best{2.20} & \best{7.55}}
  {\second{13.65} & \best{2.27} & \best{7.79}}
  {1.77 & \second{1.33} & \second{3.27}}
  {\second{15.65} & \second{1.32} & \best{4.99}}
  {\best{2.36} & \second{1.99} & \best{1.19}}
  {\second{3.46}}
  {\best{0.65}}

\DataRow
  {\textbf{Ours (RGB)}}
  {\best{12.72} & 2.47 & \second{8.14}}
  {\best{13.50} & 2.54 & \second{8.40}}
  {\second{1.75} & 1.39 & 3.46}
  {\best{15.44} & 1.34 & \second{5.45}}
  {\best{2.36} & 1.93 & \second{1.17}}
  {\best{3.38}}
  {0.68}

\StateRow{Listening}

\DataRow
  {\MethodRef{UniLS}{chu2026unils}}
  {22.76 & 3.92 & 22.61}
  {23.70 & 4.03 & 22.88}
  {2.70 & 1.84 & 8.10}
  {22.93 & 2.86 & 5.86}
  {1.47 & 1.29 & 0.58}
  {4.63}
  {1.05}

\DataRow
  {\MethodRef{DualTalk}{peng2025dualtalk}}
  {13.50 & 3.25 & 8.61}
  {\best{13.57} & 3.28 & 8.65}
  {\best{1.36} & \best{1.14} & \best{2.92}}
  {18.77 & 2.44 & 6.02}
  {0.13 & 0.22 & 0.13}
  {4.81}
  {1.30}

\DataRow
  {\textbf{Ours (audio)}}
  {14.41 & \second{2.43} & 9.56}
  {15.36 & \second{2.54} & 9.83}
  {1.89 & 1.35 & 3.88}
  {17.29 & \best{1.81} & 5.36}
  {\second{2.11} & \best{1.81} & \second{1.13}}
  {4.49}
  {0.89}

\DataRow
  {\textbf{Ours (FLAME)}}
  {\best{13.34} & \best{2.29} & \best{7.61}}
  {\second{14.20} & \best{2.39} & \best{7.83}}
  {1.76 & \second{1.28} & \second{3.13}}
  {\second{16.93} & \second{1.82} & \best{4.71}}
  {\best{2.14} & \best{1.81} & \best{1.17}}
  {\second{3.95}}
  {\second{0.86}}

\DataRow
  {\textbf{Ours (RGB)}}
  {\second{13.39} & 2.54 & \second{8.25}}
  {14.25 & 2.63 & \second{8.48}}
  {\second{1.75} & 1.31 & 3.34}
  {\best{16.82} & 1.84 & \second{4.98}}
  {2.09 & \second{1.69} & 1.11}
  {\best{3.85}}
  {\best{0.85}}

\bottomrule
\end{tabular}%
}
\end{table*}

\begin{table*}[t]
\centering

\caption{\textbf{Context signals in mesh space.} Speaking/listening results with
LVE/MHD in millimeters. Best and second best per split and state use green
bold and blue underlining. Lower is better.}
\label{tab:context-mesh}

\setlength{\tabcolsep}{1.8pt}
\renewcommand{\arraystretch}{1.08}
\footnotesize

\resizebox{\linewidth}{!}{%
\begin{tabular}{
Y
!{\color{TableSeparator}\vrule width 0.45pt}
*{6}{c}
!{\color{TableSeparator}\vrule width 0.45pt}
*{6}{c}
!{\color{TableSeparator}\vrule width 0.45pt}
*{6}{c}
}
\toprule
& \multicolumn{6}{c!{\color{TableSeparator}\vrule width 0.45pt}}{\textbf{ID}}
& \multicolumn{6}{c!{\color{TableSeparator}\vrule width 0.45pt}}{\textbf{OOD}}
& \multicolumn{6}{c}{\textbf{OOD-Hard}} \\
\cmidrule(l{1.5pt}r{1.5pt}){2-7}
\cmidrule(l{1.5pt}r{1.5pt}){8-13}
\cmidrule(l{1.5pt}r{1.5pt}){14-19}
\multicolumn{1}{c!{\color{TableSeparator}\vrule width 0.45pt}}{\textbf{Method}}
& \multicolumn{3}{c}{\textbf{Speaking}}
& \multicolumn{3}{c!{\color{TableSeparator}\vrule width 0.45pt}}{\textbf{Listening}}
& \multicolumn{3}{c}{\textbf{Speaking}}
& \multicolumn{3}{c!{\color{TableSeparator}\vrule width 0.45pt}}{\textbf{Listening}}
& \multicolumn{3}{c}{\textbf{Speaking}}
& \multicolumn{3}{c}{\textbf{Listening}} \\
\cmidrule(l{1.5pt}r{1.5pt}){2-4}
\cmidrule(l{1.5pt}r{1.5pt}){5-7}
\cmidrule(l{1.5pt}r{1.5pt}){8-10}
\cmidrule(l{1.5pt}r{1.5pt}){11-13}
\cmidrule(l{1.5pt}r{1.5pt}){14-16}
\cmidrule(l{1.5pt}r{1.5pt}){17-19}
& \textbf{FDD} $\downarrow$ & \textbf{LVE} $\downarrow$ & \textbf{MHD} $\downarrow$
& \textbf{FDD} $\downarrow$ & \textbf{LVE} $\downarrow$ & \textbf{MHD} $\downarrow$
& \textbf{FDD} $\downarrow$ & \textbf{LVE} $\downarrow$ & \textbf{MHD} $\downarrow$
& \textbf{FDD} $\downarrow$ & \textbf{LVE} $\downarrow$ & \textbf{MHD} $\downarrow$
& \textbf{FDD} $\downarrow$ & \textbf{LVE} $\downarrow$ & \textbf{MHD} $\downarrow$
& \textbf{FDD} $\downarrow$ & \textbf{LVE} $\downarrow$ & \textbf{MHD} $\downarrow$ \\
\midrule

\MethodRef{UniLS}{chu2026unils}
& \best{29.21} & \best{7.62} & \best{1.79} & \best{31.11} & 9.92 & 2.21
& \best{27.37} & \best{7.29} & \second{1.71} & \best{28.19} & 8.91 & 1.97
& 21.84 & 10.28 & 2.27 & 21.78 & 10.82 & 2.37
\\

\MethodRef{DualTalk}{peng2025dualtalk}
& 68.51 & \second{8.12} & \second{1.84} & 61.57 & \best{9.26} & \best{2.03}
& 64.35 & \best{7.29} & \best{1.65} & 55.63 & \best{8.03} & \best{1.75}
& 42.83 & \best{7.65} & \best{1.67} & 38.31 & \best{7.64} & \best{1.67}
\\

\textbf{Ours (audio)}
& \second{36.64} & 9.15 & 2.05 & \second{35.49} & 9.91 & \second{2.17}
& \second{32.36} & 8.45 & 1.89 & \second{28.49} & 8.83 & 1.93
& \second{18.90} & 9.06 & 1.97 & 18.35 & 9.24 & 2.00
\\

\textbf{Ours (FLAME)}
& 36.96 & 9.16 & 2.06 & \second{35.49} & \second{9.90} & 2.18
& 33.09 & \second{8.34} & 1.87 & 29.22 & \second{8.73} & \second{1.91}
& \best{18.74} & \second{8.73} & \second{1.90} & \second{18.14} & 8.77 & 1.90
\\

\textbf{Ours (RGB)}
& 37.63 & 9.35 & 2.10 & 36.23 & 10.08 & 2.22
& 33.96 & 8.55 & 1.90 & 29.68 & 8.84 & 1.93
& 19.09 & 8.83 & 1.92 & \best{18.13} & \second{8.76} & \second{1.89}
\\

\bottomrule
\end{tabular}%
}
\end{table*}

\paragraph{Context choices follow the task definition}
Following Section~\ref{sec:preliminaries}, we compare dyadic audio alone,
audio with user RGB frames, and audio with pre-extracted user FLAME.
Tables~\ref{tab:context-parameter} and~\ref{tab:context-mesh} compare speaking
and listening with UniLS and DualTalk on the same splits. Each setting retains
our adaptive generator design. The models are trained separately, with RGB
from an earlier generator revision.

\paragraph{Motion statistics improve across context interfaces}
With comparable inputs, our audio model improves FD and P-FD over UniLS
across motion components, splits, and states, while our FLAME model improves
FD over DualTalk. All three settings lower FDD relative to DualTalk, which
retains lower expression/neck MSE, LVE, and MHD. These complementary results
support the adaptive backbone across context representations. The
same-checkpoint comparisons in Table~\ref{tab:ablation-ttt} separately show
that continued TTT updates and retained context improve expression
statistics on ID.

\paragraph{Explicit geometry provides useful motion structure}
User FLAME lowers jaw FD and P-FD relative to both RGB and audio alone in
every split and state. Its neck FD/P-FD advantage over RGB holds on ID and
OOD-Hard, while RGB remains stronger on OOD. FLAME also generally lowers mesh
FDD, LVE, and MHD relative to RGB, which is slightly better on OOD-Hard
listening.

\paragraph{Visual evidence complements dyadic audio}
Audio alone achieves lower expression FD and P-FD than RGB on ID and OOD. RGB
achieves lower neck FD/P-FD on OOD and expression and neck FD/P-FD on OOD-Hard
in both states. It also lowers LVE and MHD on OOD-Hard in both states. These
gains beyond jaw articulation motivate better use of visible conversational
behavior.

\paragraph{RGB establishes a baseline from observable inputs}
We retain RGB as the primary setting to establish an end-to-end inference
baseline from causally arrived face frames and both audio streams to avatar
motion. It avoids the precomputed user FLAME input used by DualTalk, even
when explicit geometry yields better scores. Our FLAME comparison evaluates
pre-extracted inputs rather than an online reconstruction pipeline.
Together, these settings provide a starting point for learning how visual
observations can improve responsive head motion during an unfolding
conversation.

\clearpage

\section{Method implementation and streaming algorithm}
\label{app:method-details}

This appendix specifies the default model in Section~\ref{sec:evolving-avatar}.
Motion is grouped into 200\,ms intervals, giving five frames at 25\,fps. Audio
features use 10\,ms hops aligned with video frames. Training statistics remain
fixed after preprocessing. The codec is trained first and frozen, supplying
posterior-mean targets during generator training and decoding latents at
inference without motion targets.

\subsection{Causal motion codec}
\label{app:codec-details}

\paragraph{Architecture} Table~\ref{tab:method-architecture} lists the default
module dimensions. The encoder standardizes FLAME parameters, adds learned
within-interval phase embeddings, and mixes the five frames into one token
before causal convolutions. Regional decoder paths share convolution weights
but keep separate bounded histories. Phase embeddings and linear heads expand
each token to five motion frames, followed by inverse normalization. The
encoder samples its posterior only during codec training.

\begin{table}[h]
  \centering
  \small
  \caption{\textbf{Default architecture.} Dimensions and context sizes for the
  causal codec and generator.}
  \label{tab:method-architecture}
  \begin{tabular}{p{0.23\linewidth}p{0.70\linewidth}}
    \toprule
    Component & Configuration \\
    \midrule
    FLAME codec & Input 106, latent $[\mathrm{Exp}8,\mathrm{Neck}2,\mathrm{Jaw}2,\mathrm{Coord}4]$ \\
    Causal codec stack & Width 1024, kernel 3, residual dilations $(1,3,9)$ \\
    Posterior and decoder & Log-variance clamp $[-30,20]$, regional inputs $(12,6,6)$, outputs $(100,3,3)$ \\
    Audio features & 16\,kHz, 80 Mel bins, window/FFT 400, hop 160, range 20--7,600\,Hz \\
    Shared Emformer & 4 layers, width 256, 4 heads, FFN 1024, output 384 \\
    Emformer context & Acoustic segment 4, left/right context 8/0, memory 16 vectors, dropout 0 \\
    Frozen visual model & DINOv2 ViT-B/14 with 4 registers, $224\times224$ RGB, backbone width 768 \\
    Visual pooling & 1 class-token feature and 3 learned local queries, 4 heads, regional width 128 \\
    Causal visual adapter & Input 512, width 384, kernel 3, dilations $(1,2)$, dropout 0.1, output 256 \\
    Interval tokens & Width 512, 4 visual queries, 1 audio query per role \\
    Activity predictor & Two width-256 tanh branches, product fusion, 2 classes \\
    Main flow & 8 blocks, width 512, 8 heads, FFN 2048, output 16 \\
    Jaw flow & 4 blocks, width 256, 4 heads, FFN 1024, output 6 \\
    Flow conditioning & 1,000 time buckets, attention/FFN dropout 0, 4 left-Euler steps \\
    \bottomrule
  \end{tabular}
\end{table}

\paragraph{Reconstruction and rate} Reconstruction uses normalized motion
residuals, scaling their first and second differences by channelwise training
RMS. Within each 16-sequence loss group $\mathcal G$, squared errors are
averaged over valid frames and channels within each region, then equally over
the three regions and valid temporal orders to give $e_{\mathcal G}$. For
$N_{\mathcal G}$ valid frames, the reconstruction term is
\begin{equation}
  \begin{aligned}
    \log\sigma_{\mathcal G}
      &=-6+\operatorname{softplus}\!\left(\tfrac12\log e_{\mathcal G}+6\right),\\
    \mathcal L_{\mathcal G}
      &=106N_{\mathcal G}\!\left[
        \frac{e_{\mathcal G}}{2\sigma_{\mathcal G}^2}
        +\log\sigma_{\mathcal G}+\tfrac12\log(2\pi)\right].
  \end{aligned}
  \label{eq:codec-gaussian-loss}
\end{equation}
A positive floor stabilizes the logarithm. Reconstruction losses average over
loss groups, while unweighted KL sums each sequence's intervals and latent
dimensions before averaging over sequences. The rate constraint adds
$\mu\bar\kappa$ with target $\rho=69.3$ nats per interval, where $\bar\kappa$
is the global KL per valid interval. The multiplier starts at zero and updates
after step 5,000 as follows:
\begin{equation}
  \mu_{k+1}=\operatorname{clip}_{[0,91]}
  \left(\mu_k+\frac{0.013}{1+k/10000}
  \left(\frac{\bar\kappa}{69.3}-1\right)\right),
\end{equation}
Here $k$ counts dual updates, with initial step size $0.013=10^{-3}\times13$
for a 13-interval codec clip.

\subsection{Audiovisual features and motion backbone}
\label{app:generator-architecture}

\paragraph{Features and tokenization} The shared Emformer keeps separate caches
for the two audio roles. DINOv2 uses RGB means $(0.485,0.456,0.406)$ and
standard deviations $(0.229,0.224,0.225)$. Local pooling adds signed and
absolute patch changes through tanh gates initialized at zero. Regional outputs
are concatenated before the temporal adapter, whose input projection also
receives signed and absolute frame differences. Phase-aware queries pool five
frames into four visual and two audio tokens. Learned slots identify visual
queries and audio roles, followed by audio LayerNorm.

\paragraph{Activity and conditioning} Two separate tanh branches project
normalized user audio/video and avatar audio. Their product enters LayerNorm
and a binary head. Speaking combines avatar-only speech and overlap, and
listening combines user-only speech and silence. Detached posteriors enter
state embeddings and a phase-aware mixer. LayerNorm, a zero-initialized linear
map, and $1+\tanh(\cdot)$ produce the gate. Motion gradients train the state
embedding, mixer, and gate.

\paragraph{Main backbone} Visual tokens alternate audio cross-attention and
visual self-attention. The post-attention context $\mB_g^\ell$ is
RMS-normalized to $\mH_g^\ell$ before DCP projections. The context residual
uses $\mB_g^\ell+\mD_g^\ell$, whereas motion attends to
$\mH_g^\ell+\bm\gamma_g\odot\mD_g^\ell$ without condition-value or output
biases. Both streams share a GELU feed-forward block. Context uses a zero-time
AdaLN anchor, and motion uses its sampled flow time. Only the motion token
enters the $512\rightarrow512\rightarrow16$ ReLU velocity decoder.

\subsection{Fast weights and transient jaw adaptation}
\label{app:adaptation-details}

\paragraph{Persistent update} Each main block uses eight
$64\rightarrow128\rightarrow64$ GELU MLPs with biases. The deltas contain
1,060,864 FP32 scalars per conversation. DCP averages squared errors over four
context tokens and 512 channels. Keys, values, queries, and update rates share
the normalized context:
\begin{equation}
  \eta_{g,a}^\ell=\operatorname{sigmoid}\!\left(
    \alpha^\ell+\vw_{\eta,a}^{\ell\top}
    \frac14\sum_{r=1}^4\mH_g^\ell[r]\right).
  \label{eq:contextual-update-rate}
\end{equation}
The $8\times512$ rate projection is zero-initialized, giving initial rates of
0.1 bounded by 1. Analytic updates retain outer gradients. Adapted and initial
outputs are separately RMS-normalized, differenced, and projected without bias
in FP32. The outer graph is detached every 40 intervals while numerical deltas
persist. DCP never reads interval indices, activity predictions, or motion.

\paragraph{Jaw adaptation and decoding} Six detached audio/visual tokens are
projected to width 256 with learned role slots and LayerNorm. A four-head fast
MLP uses the same head geometry and rate settings, with DCP averaged over six
tokens and 256 channels. It updates from zero deltas and adds its adaptation
residual without an activity gate. Temporary deltas are discarded, and all four
solver steps reuse the adapted context. Routed decoding requires the full main
endpoint and private six-dimensional jaw endpoint. One decoder call processes
all three routes with separate histories.

\subsection{Generator training and optimization}
\label{app:generator-training}

\paragraph{Clip sampling} Training clips contain 130--600 frames in complete
intervals, retaining all available intervals for shorter records. A prefix $A$
builds context for a supervised suffix $B$. With probability 0.2, $A$ is empty.
Otherwise, its length is uniform between 20 intervals and the largest feasible
prefix, reducing the lower bound for short clips. We reserve at least two $B$
intervals when available. Both parts advance causal frontends and DCP from cold
state, with motion losses restricted to $B$.

\paragraph{Loss settings} Each interval independently samples
$u_g\sim\operatorname{Beta}(1.5,1)$ and $\tau_g=0.999(1-u_g)$, with independent
Gaussian noise. The jaw branch shares the time and the noise's jaw and
coordination slice. Endpoint alignment averages half the squared error in
adjacent clean-latent differences over valid $B$--$B$ scalar elements, covering
16 main dimensions or two private jaw dimensions. It is zero without a valid
pair. The decoded-jaw loss compares normalized jaw parameters through the
frozen decoder, initialized from empty history. Activity cross-entropy covers
valid clip frames. Each loss has an independent distributed sum and count of
valid targets, adding $10^{-6}$ to the denominator.

\paragraph{Optimization} We follow Table~\ref{tab:method-training}. Flow losses
have unit weight. Endpoint alignment, decoded jaw supervision, and activity
cross-entropy each have weight 0.1. Detached jaw inputs block gradients to
shared frontends and the main branch, with separate gradient clipping for main
and jaw parameters. Training uses BF16 with FP32 codec targets, losses, and
adaptation. The codec and DINOv2 stay frozen. The cosine horizon spans one
complete epoch. Development does not select checkpoints.

\begin{table}[t]
  \centering
  \small
  \caption{\textbf{Optimization settings.} Both stages train on eight GPUs
  using the listed global batch sizes.}
  \label{tab:method-training}
  \begin{tabular}{lll}
    \toprule
    Setting & Codec & Generator \\
    \midrule
    Optimizer & AdamW & AdamW \\
    Peak learning rate & $2\times10^{-4}$ & $10^{-4}$ \\
    Adam moments & $(0.9,0.99)$ & $(0.9,0.99)$ \\
    Weight decay & $0$ & $10^{-6}$ \\
    Global batch size & 512 & 64 \\
    Training duration & 200,000 steps & One epoch \\
    Warmup & 250 steps & 50 steps \\
    Schedule & $\times0.05$ at step 190,000 & Cosine decay \\
    Clip length & 65 frames & 130--600 frames \\
    Gradient clipping & --- & Norm 1 per parameter group \\
    Random seed & 42 & 42 \\
    \bottomrule
  \end{tabular}
\end{table}

\subsection{Streaming execution}
\label{app:streaming-details}

\paragraph{Scheduling and boundaries} Algorithm~\ref{alg:streaming-inference}
consumes one frame and its aligned audio samples per call. A completed interval
emits five frames after the fifth arrival, giving a four-frame buffering delay
of 160\,ms before computation. Both flows use four left-Euler steps at times
$0,1/4,1/2,3/4$. Only the first main evaluation updates fast weights. Later
evaluations read the updated weights without advancing frontend or decoder
histories. A partial final interval uses masked pooling without either
fast-weight update and emits only its valid frames. All state resets between
conversations.

\begin{algorithm}[t]
  \caption{Causal streaming generation with persistent context and transient jaw adaptation}
  \label{alg:streaming-inference}
  \small
  \begin{algorithmic}[1]
    \Require Trained generator, frozen decoder, conversation seed
    \State Initialize empty frontend/decoder caches and pending buffer $P$
    \State Set main fast weights $W^\ell\gets W_0^\ell$ and interval index $g=0$
    \For{each arrived user frame and aligned audio increments}
      \State Advance frontends once and append features and activity posteriors to $P$
      \If{$|P|=5$}
        \State \Call{GenerateInterval}{$P$, $\mathrm{true}$}, then clear $P$
      \EndIf
    \EndFor
    \If{$|P|>0$}
      \State \Call{GenerateInterval}{$P$, $\mathrm{false}$}
    \EndIf
    \State Discard conversation-local state
    \Procedure{GenerateInterval}{$P$, $w$}
      \State Set $r=|P|$ and pool valid frames into $\mV_g$, $\mA_g$, and activity token
      \State Draw $\bm\epsilon_g$ from the conversation seed and counter $g$
      \State $z\gets\bm\epsilon_g$, $z^j\gets\Pi_{\mathrm{jaw}}(\bm\epsilon_g)$
      \For{$n=0,1,2,3$}
        \State Evaluate main velocity $v(z,n/4)$, updating then reading $W^\ell$ iff $w$ and $n=0$
        \State $z\gets z+v/4$ \Comment{All other evaluations read without writing}
      \EndFor
      \State Form jaw context $J_g$ from $[\mA_g;\mV_g]$ and set temporary deltas to zero
      \State $C^j\gets J_g+\operatorname{TTTResidual}(J_g)$ if $w$, else $C^j\gets J_g$
      \State Discard temporary deltas \Comment{One update then read before integration}
      \For{$n=0,1,2,3$}
        \State $z^j\gets z^j+v_{\theta_j}(z^j,n/4,C^j)/4$
      \EndFor
      \State Decode $(z,z^j)$ once with Equation~\ref{eq:codec-defined-routing}, emitting $r$ frames
      \State $g\gets g+1$ \Comment{Main fast weights carry to the next interval}
    \EndProcedure
  \end{algorithmic}
\end{algorithm}

\paragraph{Noise and memory controls} Noise depends deterministically on the
conversation seed and interval index. Matched cross-checkpoint comparisons use
an explicit shared seed and stable conversation key. No-write disables both
adaptation routes. Freeze-one retains the first main update, and current-only
clears main deltas after each interval. Both retain transient jaw updates. Each
control retains frontend and decoder histories, frame counters, inputs, solver
steps, and matched initial noise.

\section{Dataset construction details}
\label{app:dataset-construction}

\paragraph{Preserving a shared timeline}
The offline framework converts both formats into paired participant streams.
Dual-view recordings provide audio and video per person, whereas single-view
videos contain two faces and mixed speech. We resample both formats to 25\,fps
video and mono 16\,kHz audio while preserving their temporal origin. Dual-view
inputs must already be synchronized. The resulting crops, speech, FLAME
motion, activity labels, and timestamped utterances share the retained
timeline. This alignment preserves when one participant acts and the other
responds.

\begin{table}[t]
  \centering
  \small
  \caption{\textbf{Extraction settings.} Models and parameters for
  extracting aligned dyadic supervision.}
  \label{tab:dataset-construction-settings}
  \begin{tabular}{p{0.23\linewidth}p{0.70\linewidth}}
    \toprule
    Component & Model and settings \\
    \midrule
    Face tracking & MediaPipe, full-range for dual-view and short-range for single-view, detection/landmark thresholds 0.5/0.2 \\
    Participant crops & $512\times512$, box scale 2, box interpolation limit 5 frames \\
    Crop smoothing & Bidirectional One-Euro average, minimum cutoff 0.5\,Hz, $\beta=0.007$, derivative cutoff 1\,Hz \\
    Retained segments & Both tracks and landmarks valid, boundary trim 2 frames, inward whole-second alignment, minimum 2\,s \\
    Single-view speech & DialogueSidon with SyncNet assignment, 100 separation steps, 120\,s chunks, offset search $\pm15$ frames \\
    Speech activity & UniTalk-finetuned TalkNCE, 5-frame smoothing, on/off thresholds 0.55/0.45, minimum run 3 frames \\
    Transcription & Whisper turbo, English, sole-speaker intervals, beam size 5, temperature 0 \\
    FLAME fitting & EMICA-CVT, FLAME 2020, $224\times224$ input, crop scale 1.25, 100 expression and 300 fixed shape coefficients \\
    Temporal filtering & Gaussian expression kernel 5 with $\sigma=0.5$, pose kernel 7 with $\sigma=2$ \\
    RealTalk identity checks & MICA/ArcFace, within-stream medoid similarity $\geq0.50$, cross-stream similarity $\leq0.55$ \\
    \bottomrule
  \end{tabular}
\end{table}

\paragraph{Selecting valid paired observations}
MediaPipe~\citep{lugaresi2019mediapipe} detects faces and estimates landmarks,
using one track per dual-view stream or two tracks in a single-view video. We
interpolate only short gaps in bounding boxes and average forward and backward
One-Euro filtering to reduce crop jitter and directional lag. Retained
intervals require valid landmarks for both participants. Unreliable boundaries
are trimmed and short intervals discarded, so an intact target face is not
paired with a missing conversational partner. Each audio crop uses these same
interval boundaries.

\paragraph{Recovering participant speech}
Separating voices does not identify their visible speakers.
DialogueSidon~\citep{nakata2026dialoguesidon} separates single-view audio into
two speech channels. SyncNet~\citep{chung2016out} then compares both
channel-to-face assignments and selects the pairing with the lower total
audiovisual distance. This resolves speaker correspondence without imposing
alternating turns. Dual-view recordings retain their supplied participant
audio. The framework also supports Dolphin~\citep{li2026efficient}, which
directly conditions separation on mouth motion. All outputs share the
timeline, so assigning a voice to a face retains overlap and silence.

\paragraph{Separating facial identity from motion}
We reconstruct FLAME parameters with EMICA-CVT, combining DECA-style pose
estimation~\citep{feng2021learning}, MICA identity
shape~\citep{zielonka2022towards}, and expression features trained with
emotion and visual-speech
supervision~\citep{danvevcek2022emoca,filntisis2022visual}. Shape is estimated
from the first frame and held fixed within each participant segment, while
expression, neck pose, and jaw pose vary over time. This prevents framewise
shape estimates from changing the identity underlying the motion. Separate
temporal filters act on expression and pose within valid intervals to suppress
reconstruction jitter. The resulting annotations use the same 106-D motion
representation for single-view and dual-view videos.

\paragraph{Annotating interaction before speech content}
TalkNCE~\citep{jung2024talknce} estimates speaking activity from participant
audio and both face streams. Smoothing, hysteresis, and short-run suppression
stabilize activity estimates before combining them into four states: either
participant speaking alone, overlapping speech, and mutual silence. Keeping
overlap and silence explicit preserves events that forced speaker alternation
would discard. Whisper~\citep{radford2022robust} then transcribes each
participant's sole-speaker intervals, and its utterance timestamps are mapped
to video frames. Overlap remains labeled without transcripts. The resulting
annotations associate speech content with the participant and interval while
retaining the conversational context.

\paragraph{Validating correspondence}
Plausible faces and speech can still pair the wrong participants. We validate
matching durations and required outputs, then check single-view samples for
incorrect speech assignment, suppressed or duplicated speech, leakage, and
identity inconsistency. RealTalk identity checks require each track to remain
internally consistent and the two tracks to represent different people. Failed
or inconclusive samples are excluded. These checks target preprocessing errors
that could resemble conversational behavior.
Table~\ref{tab:dataset-construction-settings} summarizes the extraction
settings.

\paragraph{Splits}
Seamless uses session-level splits. OOD excludes training participant IDs and
source interaction IDs. Other held-out sessions form ID, and Dev comes from
the training pool. RealTalk supplies OOD-Hard from official single-view
intervals, each containing a fixed participant pair.

\paragraph{Counting}
Durations count physical segments before expansion into two directions. The
4,365 metadata IDs do not establish globally unique people, as RealTalk has no
cross-video person IDs. Turns follow sole-speaker changes across silence or
overlap, so pauses do not create turns. Shared bins use pooled 20th, 40th,
60th, and 80th duration percentiles at 6, 14, 36, and 96 seconds.

\section{Baseline training and comparison protocol}
\label{app:baseline-comparison}

All four baselines use the benchmark training partition, method-specific
objectives and schedules, and 106-D FLAME outputs at 25 fps. We retain
their generation mechanisms and conditioning inputs
(Table~\ref{tab:baseline-contexts}). DiffPoseTalk and ARTalk use fixed
training-set styles and offline audio. UniLS uses dyadic audio, and
DualTalk uses offline windows with pre-extracted User FLAME. Held-out
Avatar motion never conditions generation. We evaluate every method on the
same held-out records with shared frame selections, speaking/listening
masks, FLAME decoding, and equal-record aggregation.

\section{Perceptual evaluation details}
\label{app:perceptual-evaluation}

\begin{figure}[t]
    \centering
    \includegraphics[width=1\linewidth]{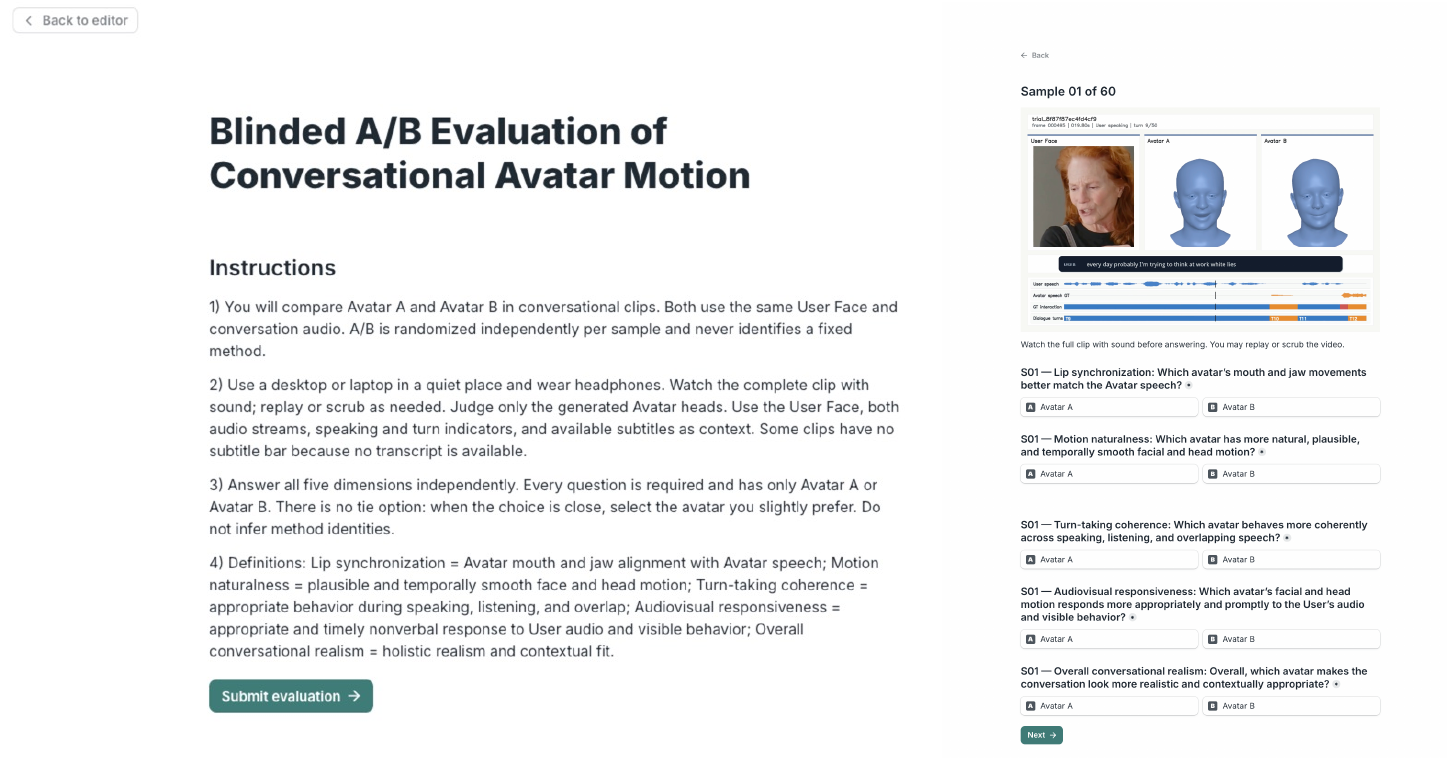}
    \caption{\textbf{User study interface.} Left: instructions for the blind A/B comparison.
    Right: a trial showing the User video and two anonymous avatar outputs with shared
    conversation audio and activity cues. Annotators choose Avatar A or Avatar B separately
    for lip synchronization, motion naturalness, turn-taking coherence, audiovisual
    responsiveness, and overall conversational realism.}
    \label{fig:a-b-test-ui}
\end{figure}

We randomly sample 60 comparison clips, with 20 each from ID, OOD, and OOD-Hard.
Ten annotators evaluate the clips using the instructions and blind A/B interface shown
in Figure~\ref{fig:a-b-test-ui}.

\end{document}